\documentclass[10pt,twocolumn,letterpaper]{article}

\usepackage{cvpr}
\usepackage{times}
\usepackage{epsfig}
\usepackage{graphicx}
\usepackage{amsmath}
\usepackage{amssymb}
\usepackage{subfigure}
\usepackage{array}
\usepackage{multirow}
\usepackage{diagbox}
\usepackage{graphicx}

\usepackage[breaklinks=true,bookmarks=false]{hyperref}

\cvprfinalcopy % *** Uncomment this line for the final submission

\def\cvprPaperID{68} % *** Enter the CVPR Paper ID here

\begin{document}

%%%%%%%%% TITLE
\title{Deep 3D Detector for RGB-D Images}
\title{Deep Sliding Shapes: A 3D ConvNets Object Detector for RGB-D Images}
\title{Deep Sliding Shapes for 3D Object Detection in RGB-D Images}
\title{Deep Sliding Shapes for Amodal 3D Object Detection in RGB-D Images}

\author{Shuran Song \quad Jianxiong Xiao\\
Princeton University\\
\href{http://dss.cs.princeton.edu}{http://dss.cs.princeton.edu}}

\maketitle
%\thispagestyle{empty}

%%%%%%%%% ABSTRACT
\begin{abstract}

We focus on the task of amodal 3D object detection in RGB-D images,
which aims to produce a 3D bounding box of an object in metric form at its full extent.
We introduce Deep Sliding Shapes, 
a 3D ConvNet formulation that takes a 3D volumetric scene from a RGB-D image as input and outputs 3D object bounding boxes. 
In our approach, we propose the first 3D Region Proposal Network (RPN)
to learn objectness from geometric shapes and the first joint Object Recognition Network (ORN) to extract geometric features in 3D and color features in 2D.
In particular, we handle objects of various sizes by training an amodal RPN at two different scales and an ORN to regress 3D bounding boxes.
Experiments show that our algorithm outperforms the state-of-the-art by 13.8 in mAP
and is $200\times$ faster than the original Sliding Shapes. 
Source code and pre-trained models are available.

%We focus on the task of amodal 3D object detection in RGB-D images,
%which aims to produce  an object's 3D bounding box that gives real-world dimensions at the object's full extent, regardless of truncation or occlusion.
%a bounding box of an object in metric form at its full extent.
%which uses a 3D ConvNet to output 3D object bounding boxes from an RGB-D image.
%a 3D ConvNet formulation that takes a 3D volumetric scene from a depth map as input and outputs 3D object bounding boxes.
%This is also the first network to regress 3D bounding boxes for objects directly from 3D proposals.
%To bypass computation limit because of 3D, we runs RPN at a low-res and ORN at a high-res.

\end{abstract}

%%%%%%%%% BODY TEXT
\vspace{-5mm}
\section{Introduction}

Typical object detection predicts the category of an object along with a 2D bounding box on the image plane for the visible part of the object.
While this type of result is useful for some tasks, such as object retrieval, 
it is rather unsatisfying for doing any further reasoning grounded in the real 3D world.
In this paper, we 
%focus on the task of amodal 3D object detection,
%which aims to produce a 3D bounding box of an object in metric form at its full extent.
%We 
focus on the task of amodal 3D object detection in RGB-D images,
which aims to produce  an object's 3D bounding box that gives real-world dimensions at the object's full extent, regardless of truncation or occlusion.
This kind of recognition is much more useful, for instance, in the perception-manipulation loop for robotics applications.
But adding a new dimension for prediction significantly enlarges the search space, %increases the number of possibilities, 
and makes the task much more challenging.

\begin{figure}[t]

\vspace{-2mm}

\includegraphics[width=\linewidth]{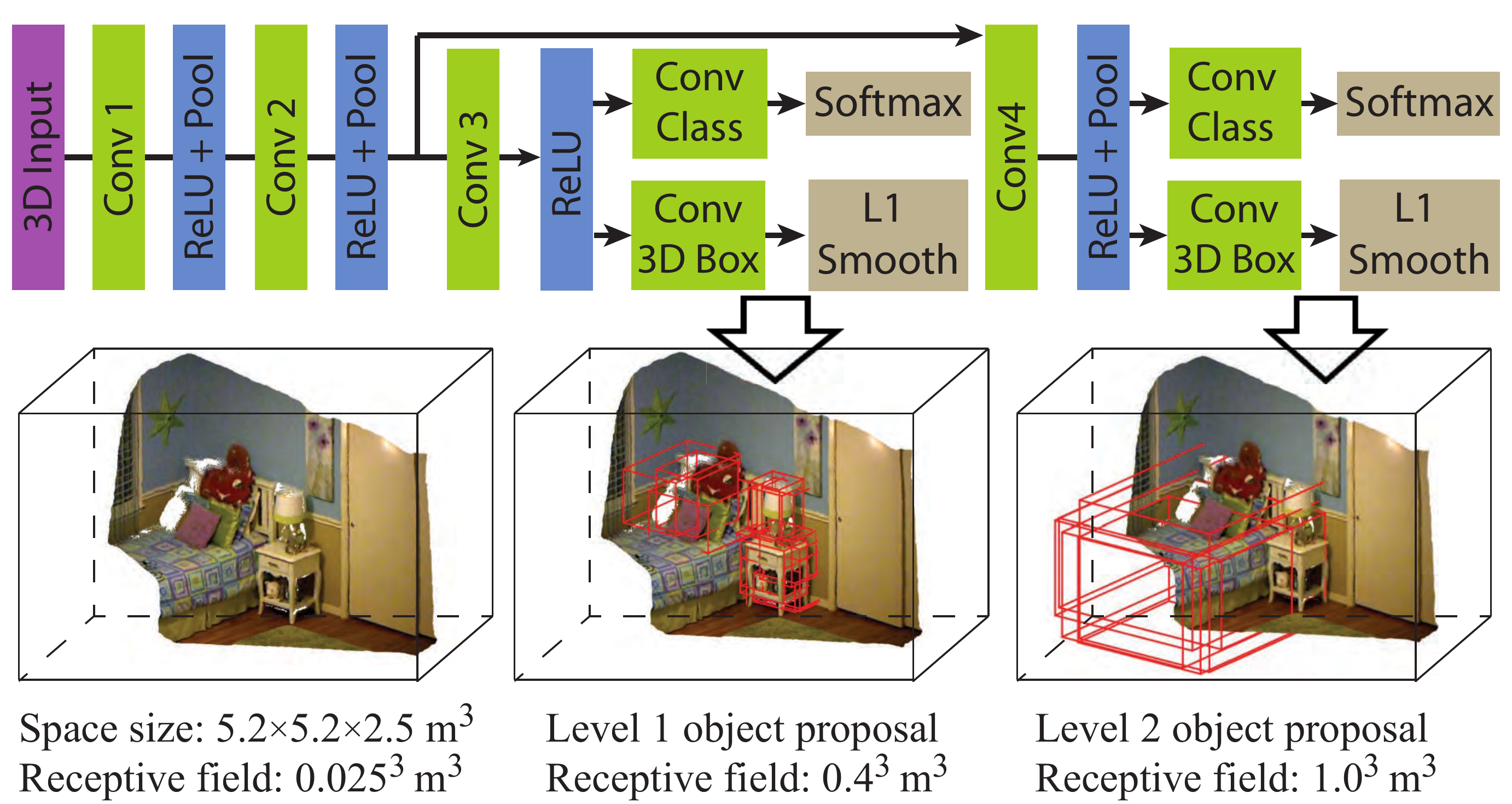}

\vspace{-0.5mm}
\caption{{\bf 3D Amodal Region Proposal Network}: Taking a 3D volume from depth as input,
our fully convolutional 3D network extracts 3D proposals at two scales with different receptive fields.}
\label{fig:RegionProposeNet}
\end{figure}

\begin{figure}[t]
\vspace{-3mm}
\includegraphics[width=\linewidth]{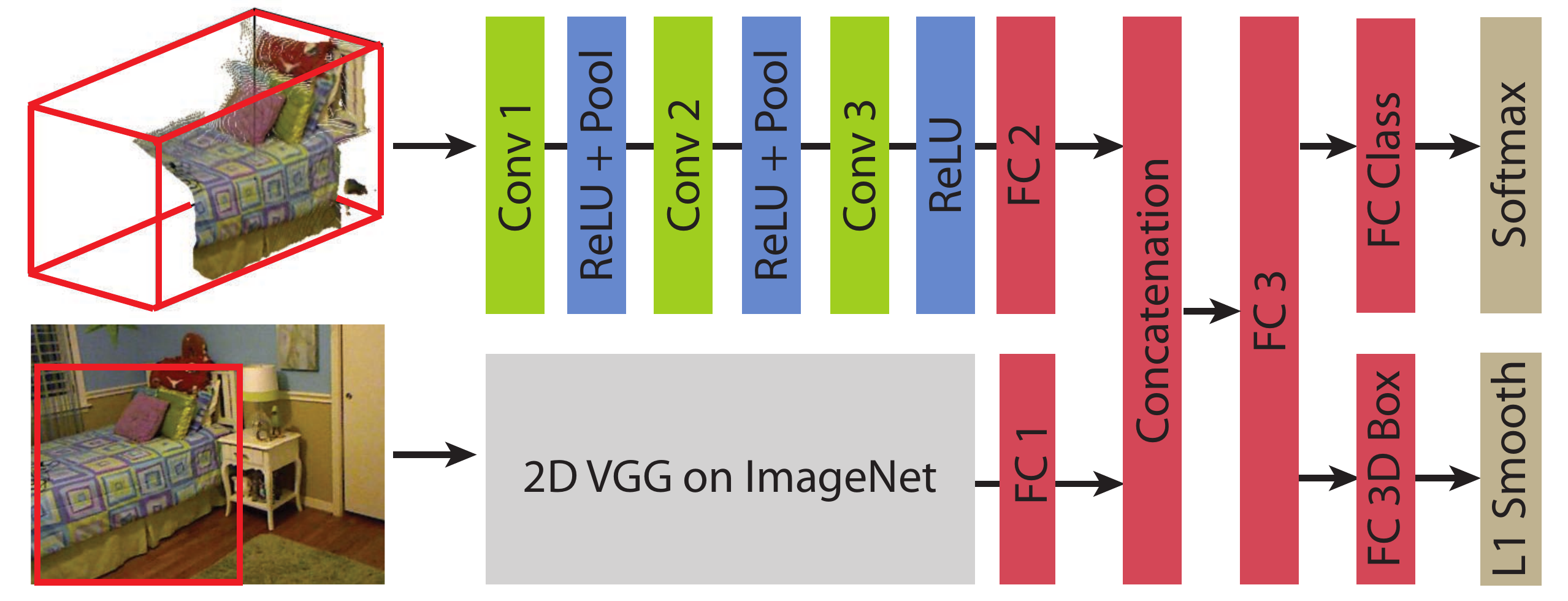}

\vspace{-0.5mm}
\caption{{\bf Joint Object Recognition Network:} For each 3D proposal,
we feed the 3D volume from depth to a 3D ConvNet,
and feed the 2D color patch (2D projection of the 3D proposal) to a 2D ConvNet,
to jointly learn object category and 3D box regression.
}
\label{fig:ObjectRecogNet}
\vspace{-3mm}
\end{figure}

The arrival of reliable and affordable RGB-D sensors (\eg, Microsoft Kinect)
has given us an opportunity to revisit this critical task.
However na\"{\i}vely converting 2D detection result to 3D does not work well (see Table \ref{fig:3ddetection} and \cite{guptaCVPR15}).
To make good use of the depth information, 
Sliding Shapes \cite{SlidingShapes} was proposed to slide a 3D detection window in 3D space.
While it is limited by the use of hand-crafted features, this approach naturally formulates the task in 3D. 
%On the other hand, Depth RCNN \cite{depthRCNN,guptaCVPR15} took a 2D approach to first detect objects in the 2D image plane, by treating depth as extra channels of an color image,
%and then fit a 3D model to the points inside the 2D detected window, by ICP alignment.
Alternatively, 
Depth RCNN \cite{guptaCVPR15} takes a 2D approach: detect objects in the 2D image plane by treating depth as extra channels of a color image, then fit a 3D model to the points inside the 2D detected window by using ICP alignment.
Given existing 2D and 3D approaches to the problem, 
it is natural to ask: {\bf which representation is better for 3D amodal object detection, 2D or 3D?}
Currently, the 2D-centric Depth RCNN outperforms the 3D-centric Sliding Shapes.
%But rather than 2D \vs 3D, 
%it may be mostly because Depth RCNN uses a well-designed 2D deep network pre-trained with ImageNet.
But perhaps Depth RCNN's strength comes from using a well-designed deep network pre-trained with ImageNet, rather than its 2D representation. 
%it may be mostly because Depth RCNN uses a well-designed 2D deep network pre-trained with ImageNet.
Is it possible to obtain an elegant but even more powerful 3D formulation by also leveraging deep learning in 3D?

In this paper, we introduce Deep Sliding Shapes, 
a complete 3D formulation to learn object proposals and classifiers using 3D convolutional neural networks (ConvNets).
We propose the first 3D Region Proposal Network (RPN) that takes a 3D volumetric scene as input and outputs 3D object proposals (Figure \ref{fig:RegionProposeNet}).
It is designed to generate amodal proposals for whole objects at two different scales for objects with different sizes.
We also propose the first joint Object Recognition Network (PRN) to use a 2D ConvNet to extract image features from color, 
and a 3D ConvNet to extract geometric features from depth  (Figure \ref{fig:ObjectRecogNet}).
This network is also the first to regress 3D bounding boxes for objects directly from 3D proposals.
Extensive experiments show that our 3D ConvNets can learn a more powerful representation for encoding geometric shapes 
(Table \ref{fig:3ddetection}),
than 2D representations (\eg HHA in Depth-RCNN). 
%when compared with 2D representations that encode depth maps as extra channels in a 2D ConvNet (\eg HHA from Depth-RCNN \cite{depthRCNN}).
%Compared to the original Sliding Shapes \cite{SlidingShapes} approach,
%while also completely formulated in 3D,
Our algorithm is also much faster than Depth-RCNN and the the original Sliding Shapes, 
as it only requires a single forward pass of the ConvNets in GPU at test time.

Our design fully exploits the advantage of 3D. % and design everything from ground up in 3D. 
Therefore, our algorithm naturally benefits from the following five aspects:
First, 
we can predict 3D bounding boxes without the extra step of fitting a model from extra CAD data.
This elegantly simplifies the pipeline, accelerates the speed, and boosts the performance because the network can directly optimize for the final goal.
%we enable an elegant solution to predict 3D bounding boxes without the extra step of model fitting using extra CAD data.
Second, 
amodal proposal generation and recognition is very difficult in 2D,
because of occlusion, limited field of view, and large size variation due to projection.
%introduced by photography bias and projection, such as .
But in 3D, 
because objects from the same category typically have similar physical sizes and the distraction from occluders falls outside the window,
our 3D sliding-window proposal generation can support amodal detection naturally.
%If it is in 2D,  has much less invariance 
Third, 
by representing shapes in 3D, our ConvNet can have a chance to learn meaningful 3D shape features in a better aligned space. % that eases feature learning.
Fourth, in the RPN, 
the receptive field is naturally represented in real world dimensions, 
% in metric form with physical meaning, 
which guides our architecture design.
Finally, we can exploit simple 3D context priors by using the Manhattan world assumption to define bounding box orientations.

While the opportunity is encouraging, there are also several unique challenges for 3D object detection. % in RGB-D images.
First, a 3D volumetric representation requires much more memory and computation. %and we cannot just extend the 2D architecture \cite{FasterRCNN} to 3D.
To address this issue, 
we propose to separate the 3D Region Proposal Network with a low-res whole scene as input,
and the Object Recognition Network with high-res input for each object.
Second, 
3D physical object bounding boxes vary more in size than 2D pixel-based bounding boxes (due to photography and dataset bias) \cite{RCNNmR}.
%(``R-CNN minus R'' \cite{RCNNmR} shows small variances of 2D boxes. ) % and even a constant set of proposals work ).
%different from 2D object detection where 2D bounding boxes have smaller variance because of photography bias \cite{RCNNmR}, . 
To address this issue, 
we propose a multi-scale Region Proposal Network 
that predicts proposals with different sizes using different receptive fields. 
Third, although the geometric shapes from depth are very useful, 
%the signal is usually in a quite low frequency, compared to the texture from color images. 
%their signal is usually in low frequency, compared to the texture from color images. 
%their signal is usually lower frequency than the frequency in texture from color images.
their signal is usually lower in frequency than the texture signal in color images.
To address this issue, 
we propose a simple but principled way to jointly incorporate color information from the 2D image patch
derived by projecting the 3D region proposal.

\subsection{Related works}

Deep ConvNets have revolutionized 2D image-based object detection. 
RCNN \cite{RCNN}, Fast RCNN  \cite{FastRCNN}, and Faster RCNN \cite{FasterRCNN}
are three iterations of the most successful state-of-the-art. 
Beyond predicting only the visible part of an object,
\cite{amodalKarTCM15} further extended RCNN to estimate the amodal box for the whole object.
But their result is in 2D and only the height of the object is estimated, while we desire an amodal box in 3D.
Inspired by the success from 2D, 
this paper proposes an integrated 3D detection pipeline
to exploit 3D geometric cues 
using 3D ConvNets for RGB-D images.

\vspace{-4mm}\paragraph{2D Object Detector in RGB-D Images}
2D object detection approaches for RGB-D images 
treat depth as extra channel(s) appended to the color images,
using hand-crafted features \cite{guptaCVPR13}, sparse coding \cite{bo2013unsupervised,bo2014learning}, or recursive neural networks \cite{Socher}.
Depth-RCNN \cite{depthRCNN,guptaCVPR15} is the first object detector using deep ConvNets on RGB-D images.
They extend the RCNN framework \cite{RCNN} for color-based object detection 
by encoding the depth map as three extra channels (with Geocentric Encoding: Disparity, Height, and Angle) appended to the color images. 
\cite{guptaCVPR15} extended Depth-RCNN to produce 3D bounding boxes by aligning 3D CAD models to the recognition results.
\cite{gupta2015cross} further improved the result by cross model supervision transfer.
For 3D CAD model classification, 
\cite{su15mvcnn} and \cite{shi2015deeppano} took a view-based deep learning approach by rendering 3D shapes as 2D image(s).

\vspace{-4mm}\paragraph{3D Object Detector}
Sliding Shapes \cite{SlidingShapes} is a 3D object detector that runs sliding windows in 3D to directly classify each 3D window.
However, because the feature dimension is different per classifier and there are many exemplar classifiers,
the algorithm is very slow.
Furthermore, the features are hand-crafted and it cannot combine with color features.
\cite{Zhile2016COG} proposed the Clouds of Oriented Gradients feature on RGB-D images.
In this paper we hope to improve these hand-crafted feature representation with 3D ConvNets that can jointly learn powerful 3D and color features from the data.

\vspace{-4mm}\paragraph{3D Feature Learning}
%Several seminal works pioneer in 3D feature learning from data \cite{bo2014learning,lai2014unsupervised}.
%Most related to this paper, 
HMP3D \cite{lai2014unsupervised} introduced
a hierarchical sparse coding technique for unsupervised learning features from RGB-D images and 3D point cloud data. 
The feature is trained on a synthetic CAD dataset, and test on scene labeling task in RGB-D video. 
In contrast, we desire a supervised way to learn 3D features using the deep learning techniques that are proved to be more effective for image-based feature learning.
%We take the inspiration from this work and further extend it to supervised feature learning 
%There are seminal works in 
%In the work \cite{lai2014unsupervised}, they introduced HMP3D, 

\begin{figure*}[t]

\vspace{-3mm}

%\centering
{\footnotesize
~~~TSDF for a scene used in Region Proposal Network~~~~~~~~~~~~~~~~~~~~~~~~~~~~~~~TSDF for six objects used in the Object Recognition Network
}

\includegraphics[width=1\linewidth]{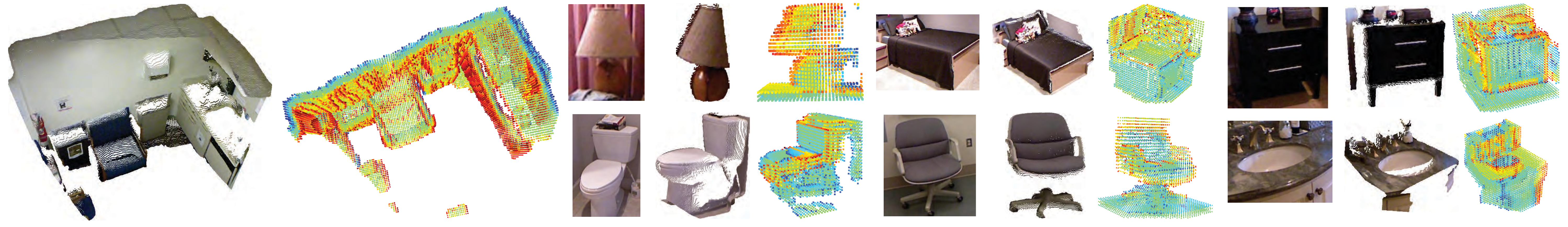}

\caption{{\bf Visualization of TSDF Encoding.} We only visualize the TSDF values when close to the surface. 
Red indicates the voxel is in front of surfaces; and blue indicates the voxel is behind the surface. 
The resolution is 208$\times$208$\times$100 for the Region Proposal Network, and 30$\times$30$\times$30 for the Object Recognition Network.}
\label{fig:tsdf}
\vspace{-3mm}
\end{figure*}

\vspace{-4mm}\paragraph{3D Deep Learning}
3D ShapeNets \cite{3DShapeNets} introduced 3D deep learning for modeling 3D shapes,
and demonstrated that powerful 3D features can be learned from a large amount of 3D data.
Several recent works \cite{VoxNet,fang20153d,xie2015deepshape,Huang:2015:deeplearningsurfaces}
also extract deep learning features for retrieval and classification of CAD models.
While these works are inspiring,
none of them focuses on 3D object detection in RGB-D images.

\vspace{-4mm}\paragraph{Region Proposal}
%Region proposal generation is a critical step in an object detection pipeline. 
For 2D object proposals, previous approaches \cite{SelectiveSearch,arbelaez2014multiscale,depthRCNN} mostly base on merging segmentation results. 
%most approaches greedily merge superpixels based on engineered low-level features \cite{SelectiveSearch,arbelaez2014multiscale,depthRCNN}. 
Recently, Faster RCNN \cite{FasterRCNN} introduces a more efficient and effective ConvNet-based formulation, % to learn objectness. 
which inspires us to learn 3D objectness using ConvNets.
%Inspired by this, our RPN learns objectness in 3D 
%Our RPN is inspired by 
For 3D object proposals, \cite{XiaozhiChen} introduces a MRF formulation with hand-crafted features for a few object categories in street scenes.
We desire to learn 3D features for general scenes from the data using ConvNets.
%an enegry minimization formulation to encode object size priors, ground plane
%  to formulate the problem as minimizing an energy function encoding object size priors, ground plane as well as several depth informed features. However they use manually defined features and energy function and only focus on a few object categories for autonomous driving.

\section{Encoding 3D Representation}
\label{sec:Representation}

\vspace{-1mm}

The first question that we need to answer for 3D deep learning is:
how to encode a 3D space to present to the ConvNets?
For color images, naturally the input is a 2D array of pixel color.
For depth maps, Depth RCNN \cite{guptaCVPR15,depthRCNN} proposed to encode depth as a 2D color image with three channels.
%depth map is not a good one because it doesn't put the object their rightful 3D location
Although it has the advantage to reuse the pre-trained ConvNets for color images \cite{gupta2015cross},
%this 2D representation does not encode the geometric meaning spatially in 3D.
we desire a way to encode the geometric shapes naturally in 3D, preserving spatial locality.
%On the other hand, 
Furthermore,
compared to methods using hand-crafted 3D features \cite{fang20153d,xie2015deepshape},
we desire a representation that encodes the 3D geometry as raw as possible, 
and let ConvNets learn the most discriminative features from the raw data.
%we want the network to learn features, not hand craft
%A good encoding should preserve the raw 3D information as much as possible. %, and also provide enough resolution for reconstructing the shape.

To encode a 3D space for recognition, we propose to adopt a directional Truncated Signed Distance Function (TSDF). %The  which is typically used for 3D modeling \cite{kinectfusion}. 
Given a 3D space, we divide it into an equally spaced 3D voxel grid.
%In the standard TSDF setup, t
The value in each voxel is defined to be the shortest distance between the voxel center and the surface from the input depth map. 
Figure \ref{fig:tsdf} shows a few examples.
To encode the direction of the surface point, 
instead of a single distance value, 
we propose a directional TSDF to store a three-dimensional vector $[dx,dy,dz]$ in each voxel to record the distance in three directions to the closest surface point. 
The value is clipped by $2\delta$, where $\delta$ is the grid size in each dimension. 
The sign of the value indicates whether the cell is in front of or behind the surface. 
The conversion from a depth map to a 3D TSDF voxel grid is implemented on a GPU.

\begin{figure}[t]
\vspace{-5mm}
\includegraphics[width=\linewidth]{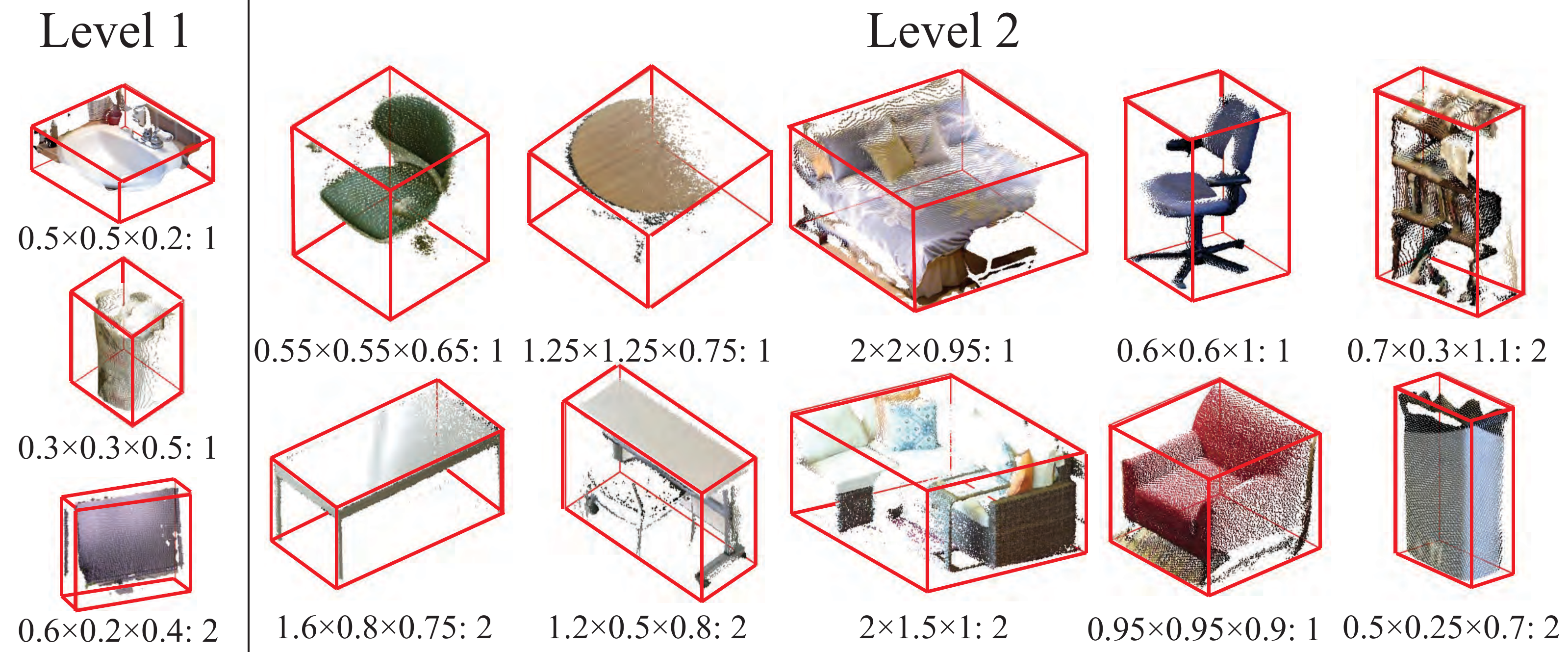}

\caption{{\bf List of All Anchors Types. }
%We also  
%We show and typical objects at those dimensions.
%Physical dimensions of all our
The subscripts show the width $\times$ depth $\times$ height in meters, followed by the number of orientations for this anchor after the colon. }
\label{fig:AllAnchor}
\vspace{-3mm}
\end{figure}

To further speed up the TSDF computation, 
as an approximation, 
we can also use projective TSDF instead of accurate TSDF where the nearest point is found only on the line of sight from the camera.
The projective TSDF is faster to compute,
but empirically worse in performance compare to the accurate TSDF for recognition (see Table \ref{fig:detection}).
We also experiment with other encodings, 
and we find that the proposed directional TSDF outperforms all the other alternatives (see Table \ref{fig:detection}).
Note that we can also encode colors in this 3D volumetric representation,
by appending RGB values to each voxel \cite{whelan2013robust}.

\section{Multi-scale 3D Region Proposal Network}
\label{sec:RegionProposalNetwork}
%The problem is challenging because:
%1. 3D is hard (more possible location, one more aspect ratios)
%2. amodal is hard (you don't even see it)
%3. object size has very different scale: a bed vs a pillow

% providing object agnostic candidates for recognition
Region proposal generation is a critical step in an object detection pipeline \cite{RCNN,FastRCNN,FasterRCNN}.
%By providing a small set of object agnostic candidates for recognition, it significantly reduces the number of possible candidates.
%Instead of exhaustive search in the original Sliding Shapes \cite{SlidingShapes},
%we desire a way to generalize region proposal to 3D to speed up the computation.
Instead of exhaustive search in the original Sliding Shapes, we desire a region proposal method in 3D to provide a small set of object agnostic candidates and speed up the computation, 
and at the same time still utilize the 3D information .
But there are several unique challenges in 3D.
First, because of an extra dimension, 
the possible locations for an object increases by 30 times \footnote{45 thousand windows per image in 2D \cite{FastRCNN} \vs 1.4 million in 3D.}.
This makes the region proposal step much more important and challenging as it need to be more selective.
%, while at the same time much more challenging (need to be more selective since we cannot afford 30 times more proposals).
Second, we are interested in amodal detection that aims to estimate the full 3D box that covers the object at its full extent. % including all the visible and non-visible parts.
%This is challenging because 
Hence an algorithm needs to infer the full box beyond the visible parts.
Third, different object categories have very different object size in 3D.
In 2D, a picture typically only focuses on the object of interest due to photography bias.
Therefore, the pixel areas of object bounding boxes are all in a very limited range \cite{FasterRCNN,RCNNmR}.
%For example, the physical size of a bed is about 10$\times$ the size of a pillow.
For example, the pixel areas of a bed and a chair can be similar in picture while their 3D physical sizes are very different. %is about 10$\times$ the size of a pillow.
%But in our 3D amodal detection task, we aim to recognize objects based on their physical sizes.

\begin{figure*} [t]
\vspace{-3mm}

{\footnotesize
~~~~~~~~~~~~~~~~Input: Color and Depth~~~~~~~~~~~~~~~~~~~~~~~~~~~~~~~~ Level 1  Proposals~~~~~~~~~~~~~~~~~~~~~~~ Level 2 Proposals~~~~~~~~~~~~~~~~~~~~~~~~Final Recognition Result
}

\centering
\includegraphics[width=0.99\linewidth]{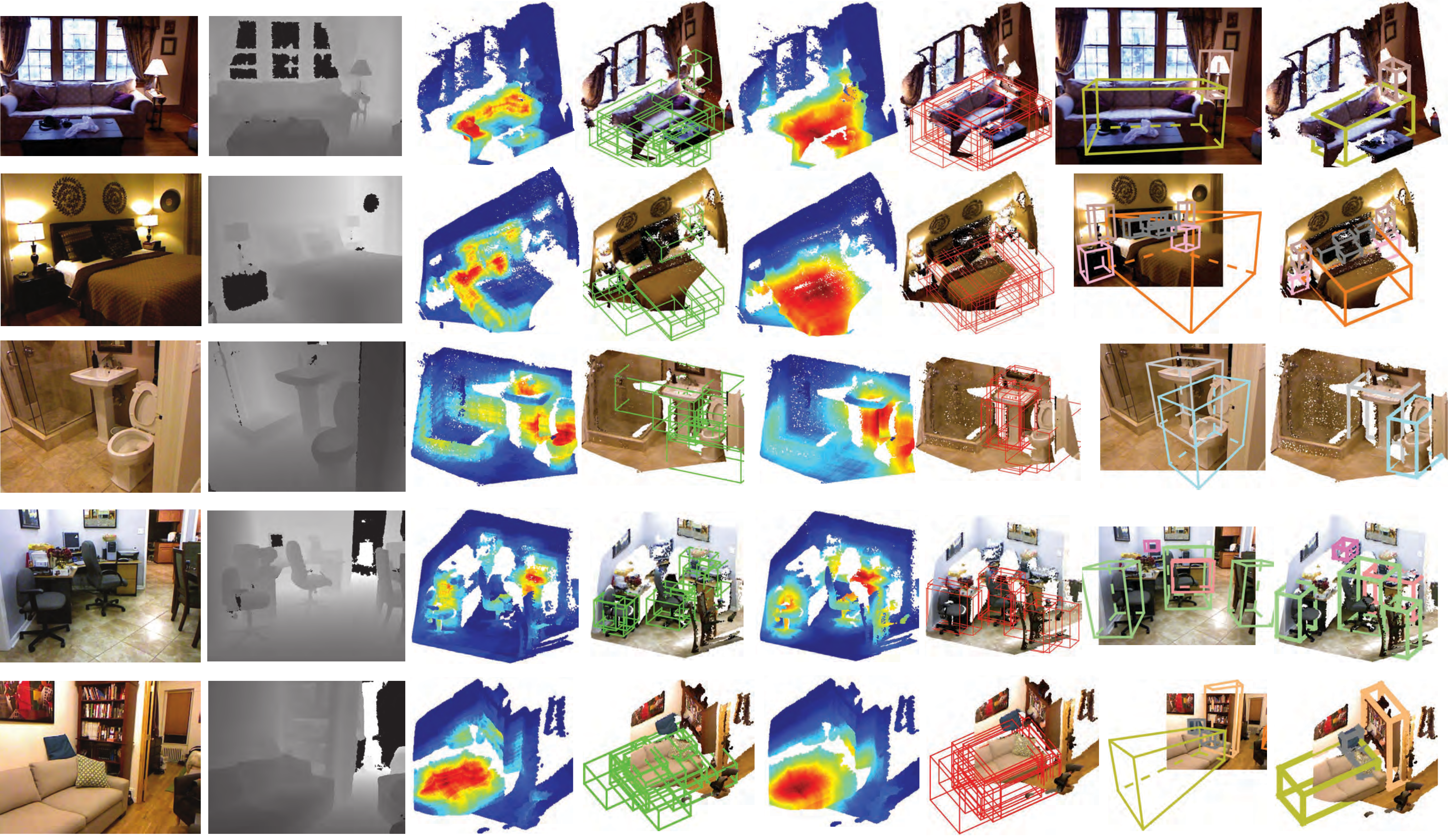}

\includegraphics[width=0.95\linewidth]{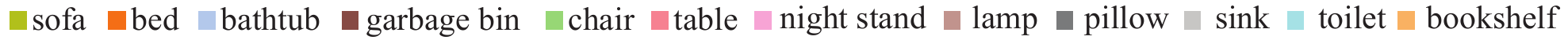}
\caption{{\bf Examples for Detection Results.} 
For the proposal results, we show the heat map for the distribution of the top proposals (red is the area with more concentration),
and a few top boxes after NMS. 
For the recognition results,
our amodal 3D detection can estimate the full extent of 3D both vertically (\eg bottom of a bed)
and horizontally (\eg full size sofa in the last row).
}
\vspace{-3mm}
\end{figure*}

To address these challenges, 
we propose a multi-scale 3D Region Proposal Network (RPN) to learn 3D objectness using back-propagation (Figure \ref{fig:RegionProposeNet}).
Our RPN takes a 3D scene as input and output a set of 3D amodal object bounding boxes with objectness scores.
The network is designed to fully utilize the information from 3D physical world such as object size, physical size of the receptive field, and room orientation.
Instead of a bottom-up segmentation based approach (\eg \cite{SelectiveSearch}) that can only identify the visible part,
our RPN 
%is a fully convolutional neural net that 
looks at all the locations for the whole object, in a style similar to sliding windows, to generate amodal object proposals.
To handle different object sizes, our RPN targets at two scales with two different sizes of receptive fields.

\vspace{-4mm}\paragraph{Range and resolution}
For any given 3D scene, we rotate it to align with gravity direction as our camera coordinate system.
Based on the specs. for most RGB-D cameras,
we target at the effective range of the 3D space $[-2.6,2.6]$ meters horizontally,  $[-1.5,1]$ meters vertically, and $[0.4,5.6]$ meters in depth. 
In this range we encoded the 3D scene by volumetric TSDF with grid size $0.025$ meters,
resulting in a $208\times208\times100$ volume as the input to the 3D RPN.

%Inspired by \cite{FasterRCNN}, we also use a fully convolution network to model the process....

\vspace{-4mm}\paragraph{Orientation}
We desire a small set of proposals to cover all objects with different aspect ratios.
Therefore, as a heuristic, we propose to use the major directions of the room for the orientations of all proposals.
%have a small number of orientations to cover each box.
%To get better box orientations,
%we find major directions of the room.
We use RANSAC plane fitting under the Manhattan world assumption, and use the results as the proposal box orientations. 
This method can give us pretty accurate bounding box orientations for most object categories. 
For objects that do not follow the room orientations, such as chairs, 
their horizontal aspect ratios tend to be a square, and therefore the orientation doesn't matter much in terms of Intersection-Over-Union (IOU).

\vspace{-4mm}\paragraph{Anchor}
For each sliding window (\ie convolution) location, the algorithm will predict $N$ region proposals. 
Each of the proposal corresponds to one of the $N$ anchor boxes with various sizes and aspect ratios.
In our case, based on statistics of object sizes, we define a set of $N=19$ anchors shown in Figure \ref{fig:AllAnchor}.
For the anchors with non-square horizontal aspect ratios, % on the two horizontal directions,
we define another anchor with the same size but rotated 90 degrees.

\vspace{-4mm}\paragraph{Multi-scale RPN}
The physical sizes of anchor boxes vary a lot, from 0.3 meters (\eg trash bin) to 2 meters (\eg bed). 
If we use a single-scale RPN, the network would have to predict all the boxes using the same receptive fields.
This means that the effective feature map will contain many distractions for small object proposals. 
To address this issue,
we propose a multi-scale RPN to output proposals at small and big scales, the big one has a pooling layer to increase receptive field for bigger objects.
We group the list of anchors into two levels based on how close their physical sizes are to the receptive fields of the output layers,
and use different branches of the network to predict them using different receptive fields.

\vspace{-4mm}\paragraph{Fully 3D convolutional architecture}
To implement a 3D sliding window style search, we choose a fully 3D convolutional architecture.
Figure \ref{fig:RegionProposeNet} shows our network architecture. 
The stride for the last convolution layer to predict objectness score and bounding box regression is 1, which is 0.1 meter in 3D.
The filter size is $2\times2\times2$ for Level 1 and $5\times5\times5$ for Level 2,
which corresponds to $0.4~\textrm{m}^3$ receptive field for Level 1 anchors and $1~\textrm{m}^3$ for Level 2 anchors.

\vspace{-4mm}\paragraph{Empty box removal}
Given the range, resolution, and network architecture, %with 19 anchor types,
the total number of anchors for any image is 1,387,646 ($19\times53\times53\times26$).
But on average, $92.2\%$ of these anchor boxes are almost empty, with point density less than 0.005 points per cm$^3$.
To avoid distraction, 
for both training and testing, we automatically remove these anchors.
This is done in constant time, by using 3D integral image. 
After removing these almost empty boxes,  there are on average 107,674 anchors remaining.

\vspace{-4mm}\paragraph{Training sampling}
For the remaining anchors, we label them as positive if their 3D IOU scores with ground truth are larger than 0.35,
and negative if their IOU are smaller than 0.15. 
In our implementation, each mini-batch contains two images.
We randomly sample 256 anchors in each image with positive and negative ratio 1:1. 
If there are fewer than 128 positive samples we pad the mini-batch with negative samples from the same image.
We select them by specifying the weights for each anchor in the final convolution layers.
We also try to use all the positives and negatives with proper weighting, but the training cannot converge.

\vspace{-4mm}\paragraph{3D box regression}
We represent each 3D box by 
its center $[c_x,c_y,c_z]$ and the size of the box $[s_{1},s_{2},s_{3}]$ in three major directions of the box (the anchor orientation for anchors, and the human annotation for ground truth).
To train the 3D box regressor, we will predict the difference of centers and sizes between an anchor box and its ground truth box.
For simplicity, we do not do regression on the orientations.
For each positive anchor and its corresponding ground truth,
we represent the offset of box centers by their difference $[\Delta c_{x},\Delta c_{y},\Delta c_{z}]$ in the camera coordinate system.
For the size difference,
we first find the closest matching of major directions between the two boxes, and then calculate 
the offset of box size $[\Delta s_{1}, \Delta  s_{2}, \Delta s_{3}]$ in each matched direction. 
Similarly to \cite{FasterRCNN}, we normalize the size difference by its anchor size. 
Our target for 3D box regression is a 6-element vector for each positive anchor
$\mathbf{t} = [\Delta c_{x},\Delta c_{y},\Delta c_{z}, \Delta s_{1}, \Delta  s_{2}, \Delta s_{3}]$.

%three orthogonal bases $ [\bf{b_1},\bf{b_2},\bf{b_3}]$\footnote{$\bf{b_3}$ is $[0,0,1]$ since all boxes are aligned with the gravity direction.}, 

\vspace{-4mm}\paragraph{Multi-task loss}
Following the multi-task loss in \cite{FastRCNN,FasterRCNN}, for each anchor, 
our loss function is defined as:
\vspace{-1mm}
\begin{equation}
\vspace{-1mm}
%\small{
L(p,p^*,\mathbf{t},\mathbf{t}^*) =L_{\textrm{cls}}(p,p^*) + \lambda p^* L_{\textrm{reg}}(\mathbf{t},\mathbf{t}^*),
%}
\end{equation}
where the first term is for objectness score, and the second term is for the box regression.
$p$ is the predicted probability of this anchor being an object and $p^*$ is the ground truth (1 if the anchor is positive, and 0 if the anchor is negative).
$L_{\textrm{cls}}$ is log loss over two categories (object \vs non object).
The second term formulates the 3D bounding box regression for the positive anchors (when $p^*=1$). 
$L_{\textrm{reg}}$ is smooth $\mathbf{L}_1$ loss used for 2D box regression by Fast-RCNN \cite{FastRCNN}.

\vspace{-4mm}\paragraph{3D NMS}
The RPN network produces an objectness score for each of the $107,674$ non-empty proposal boxes (anchors offset by regression results).
To remove redundant proposals, we apply 3D Non-Maximum Suppression (NMS) on these boxes with IOU threshold $0.35$ in 3D,
and only pick the top 2000 boxes to input to the object recognition network. %based on their objectness score
These 2000 boxes are only $0.14\%$ of all sliding windows, and it is one of the key factor that makes our algorithm much faster than the original Sliding Shapes \cite{SlidingShapes}.

\begin{figure}[t]

\includegraphics[width=\linewidth]{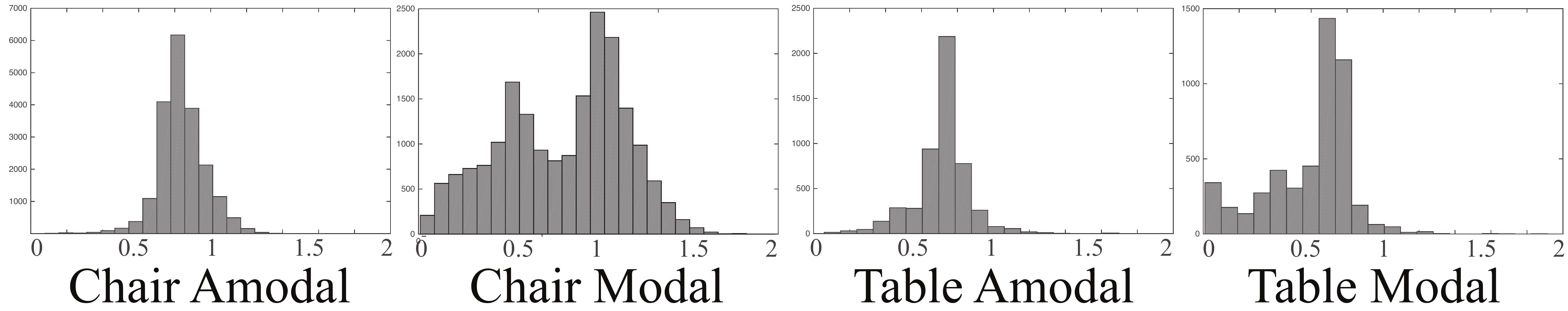}

\caption{{\bf Distributions of Heights for Amodal \vs Modal Boxes}.
The modal bounding boxes for only the visible parts of objects have much larger variance in box sizes, due to occlusion, truncation, or missing depth. 
Representing objects with amodal box naturally enables more invariance for learning.}
\label{fig:amodalHist}
\vspace{-3mm}
\end{figure}

\def \mW {0.095}
\begin{figure*}[t]

\vspace{-3mm}

\includegraphics[width=\mW\linewidth]{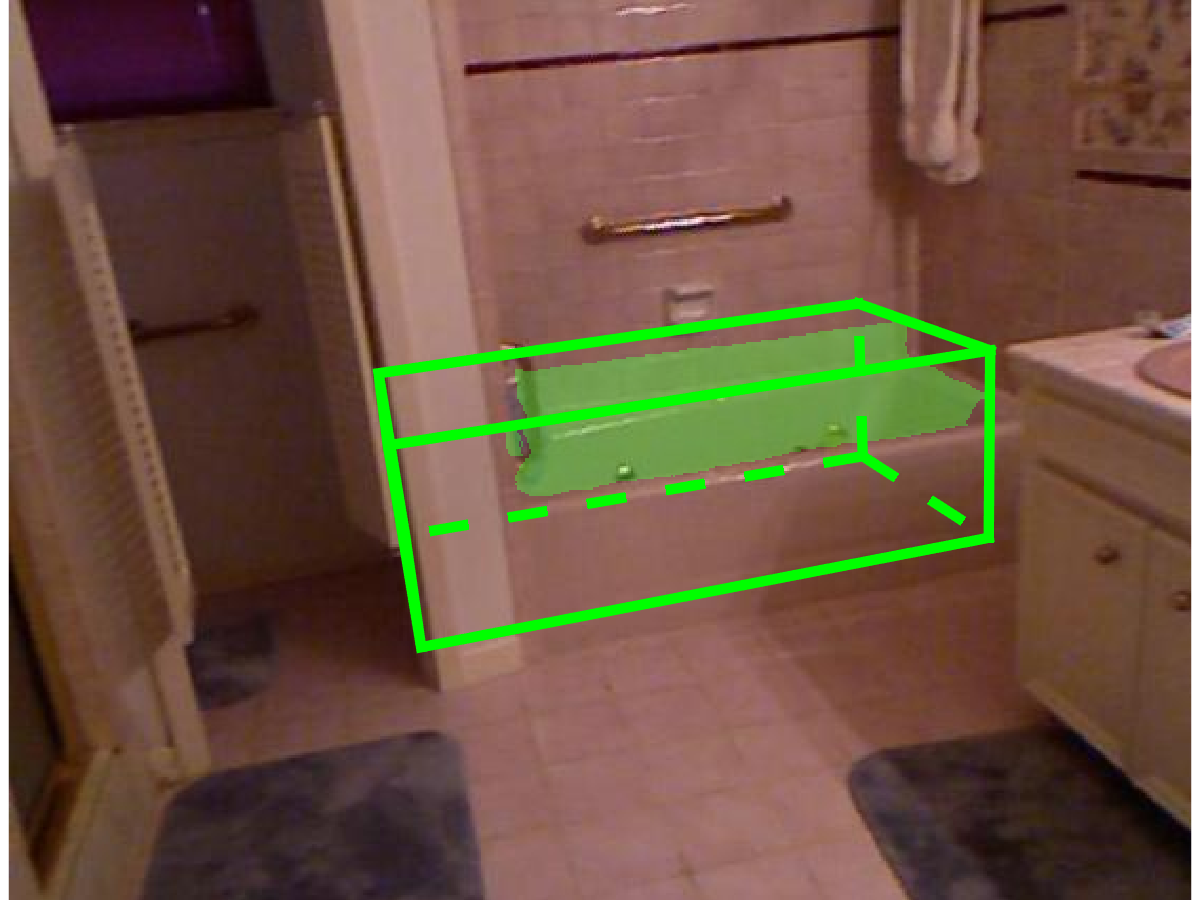}~%
\includegraphics[width=\mW\linewidth]{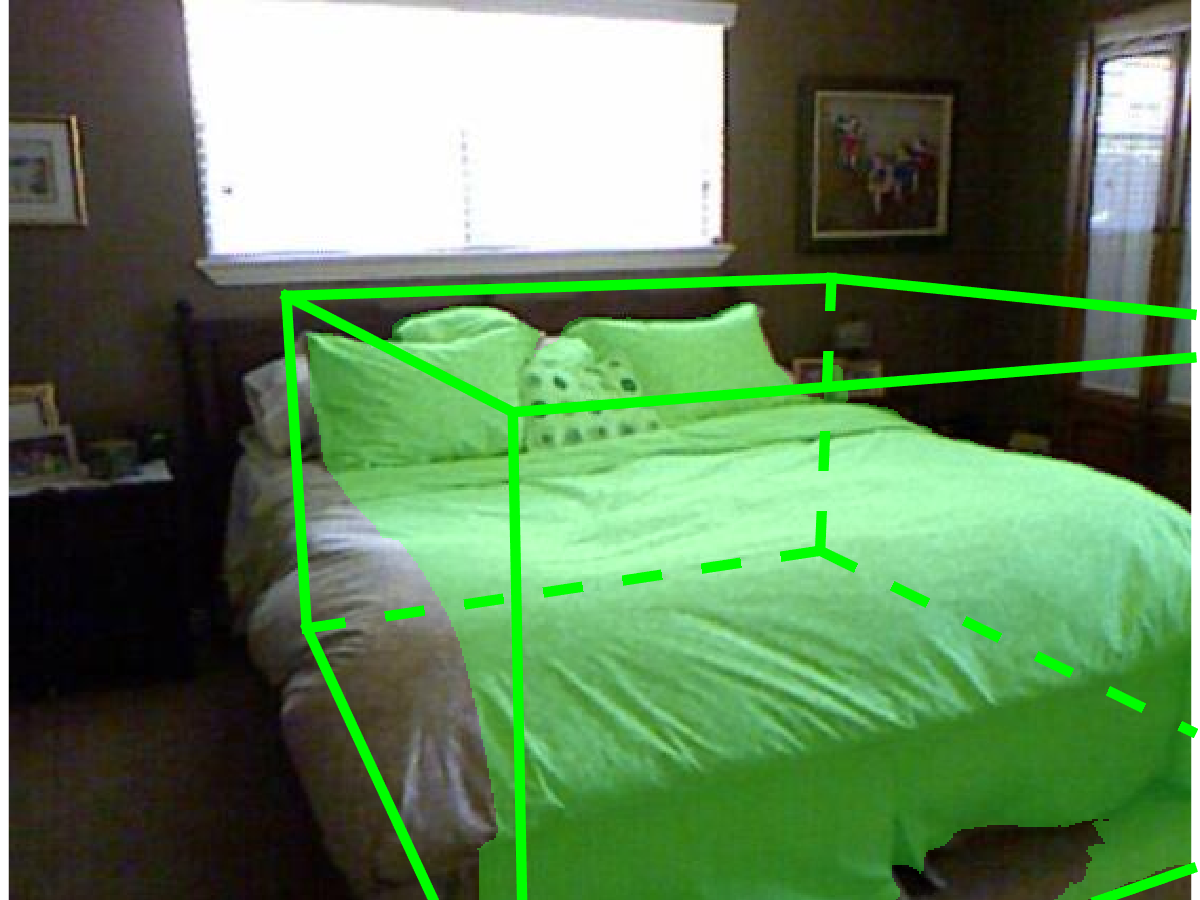}~%
\includegraphics[width=\mW\linewidth]{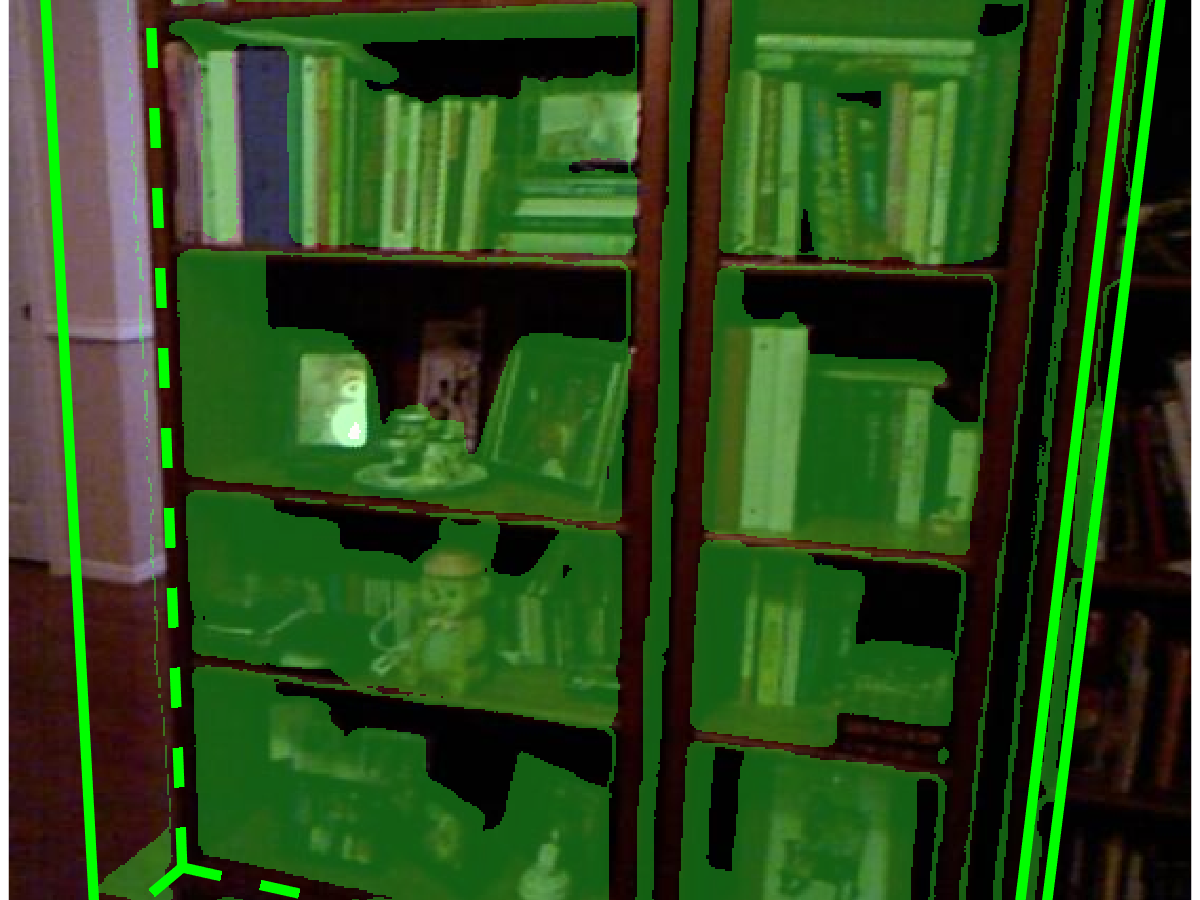}~%
\includegraphics[width=\mW\linewidth]{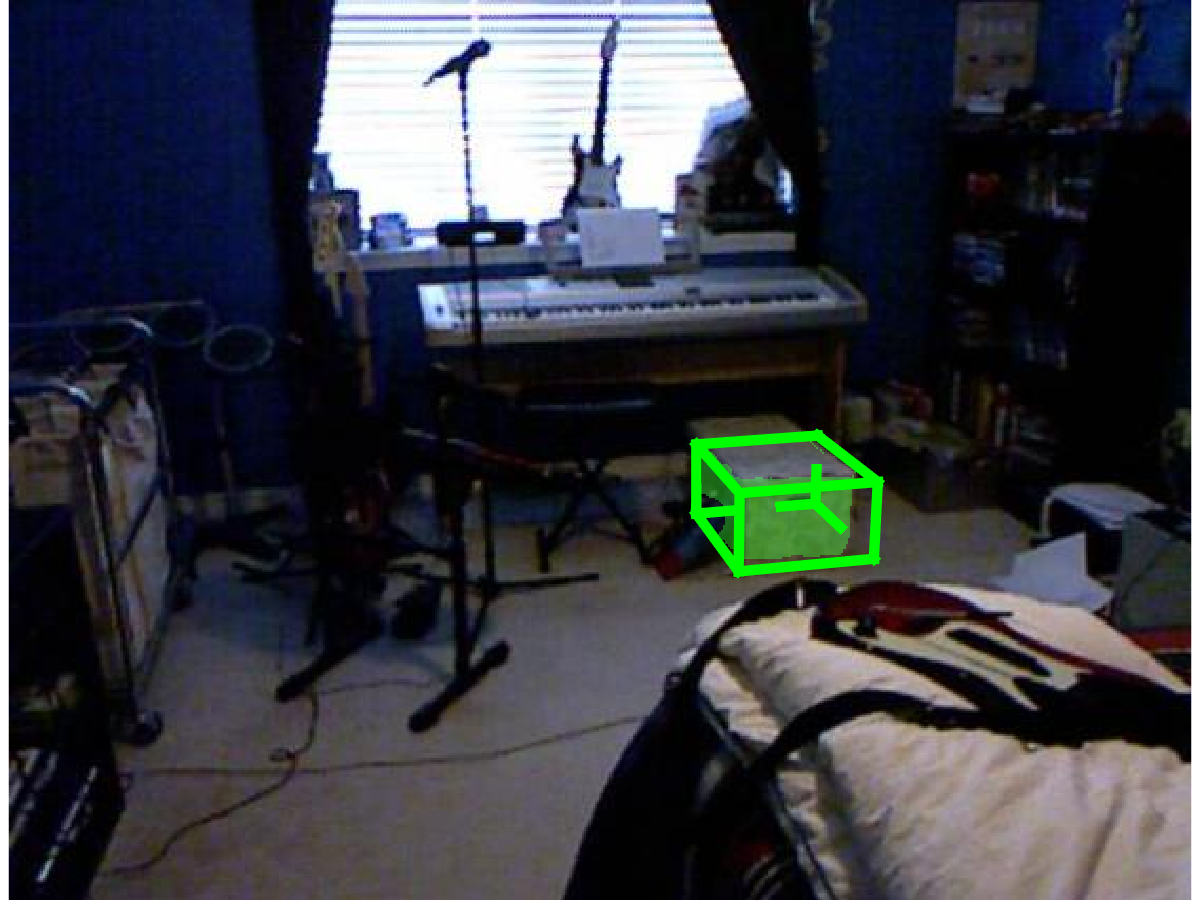}~%
\includegraphics[width=\mW\linewidth]{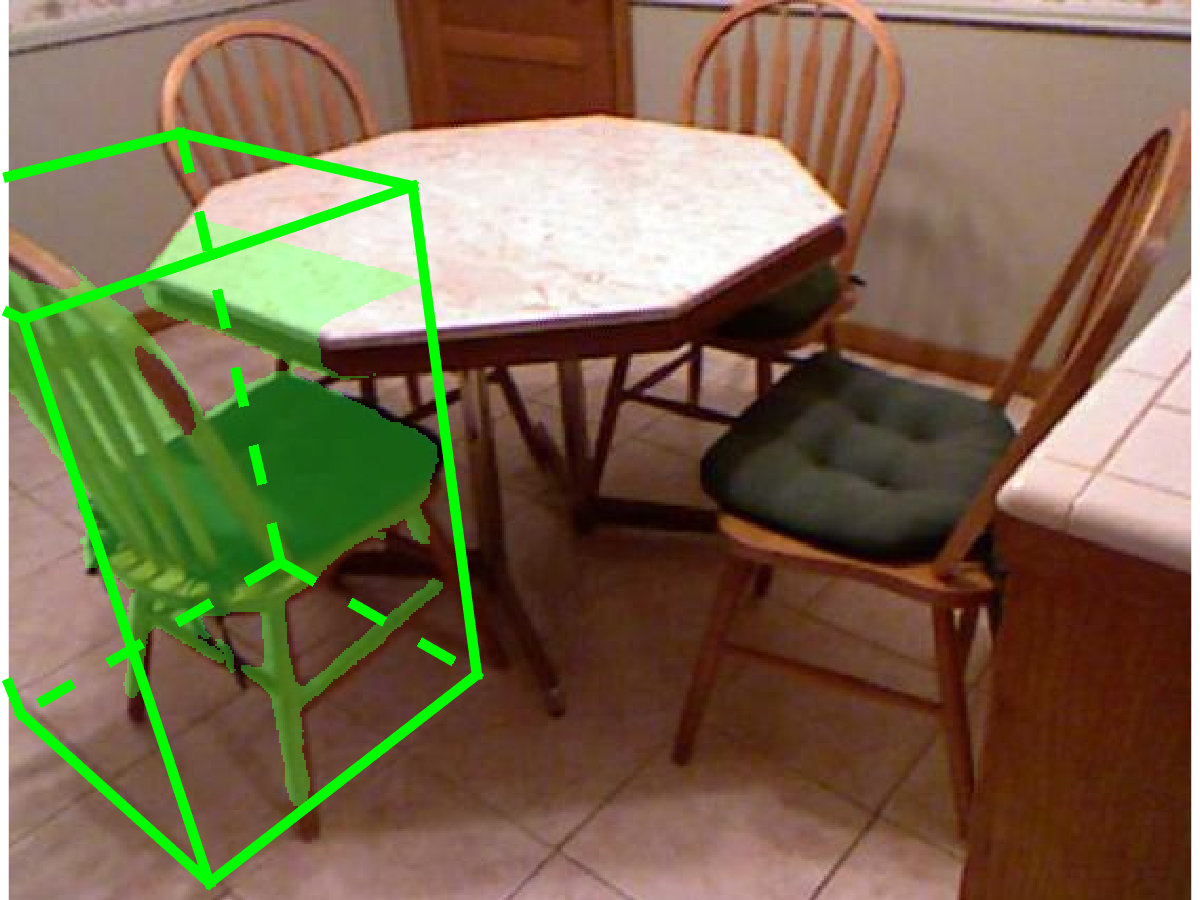}~%
\includegraphics[width=\mW\linewidth]{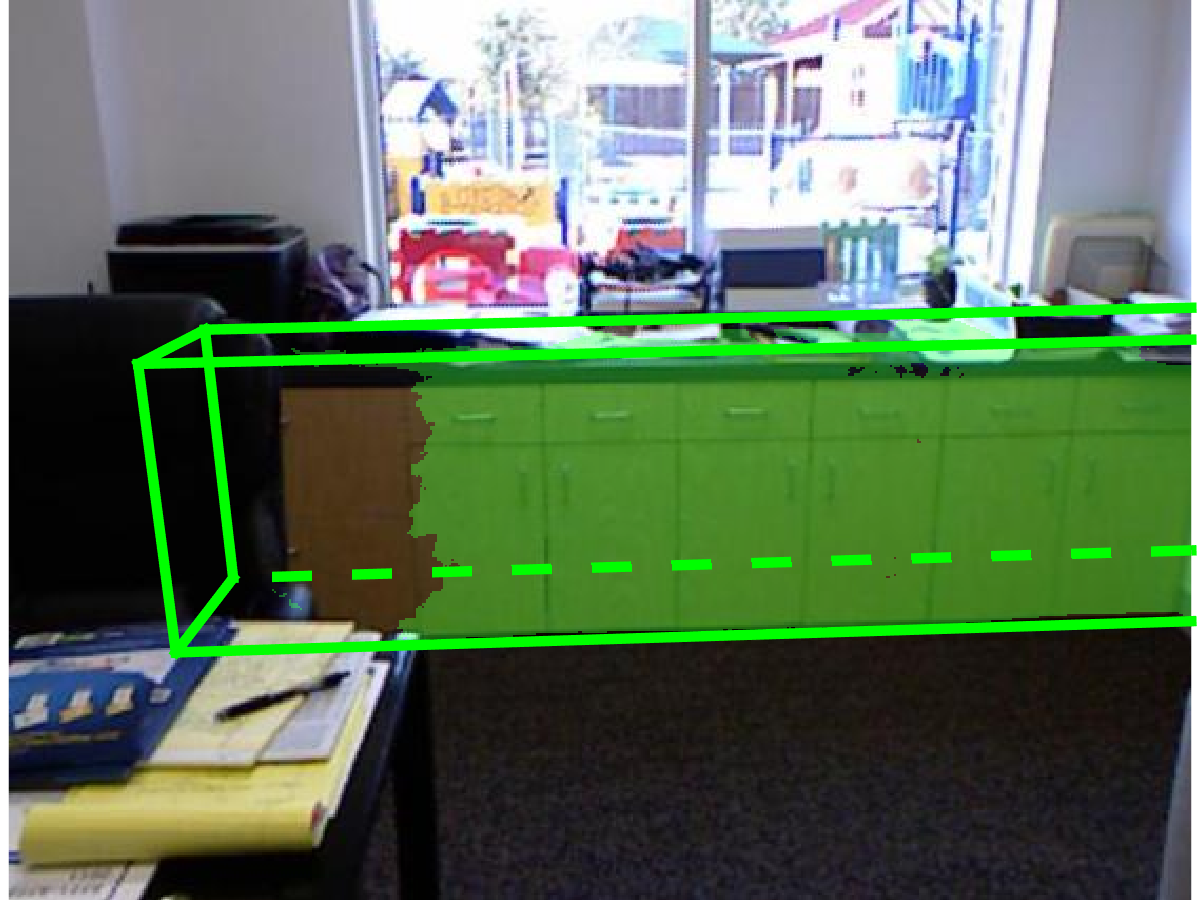}~%
\includegraphics[width=\mW\linewidth]{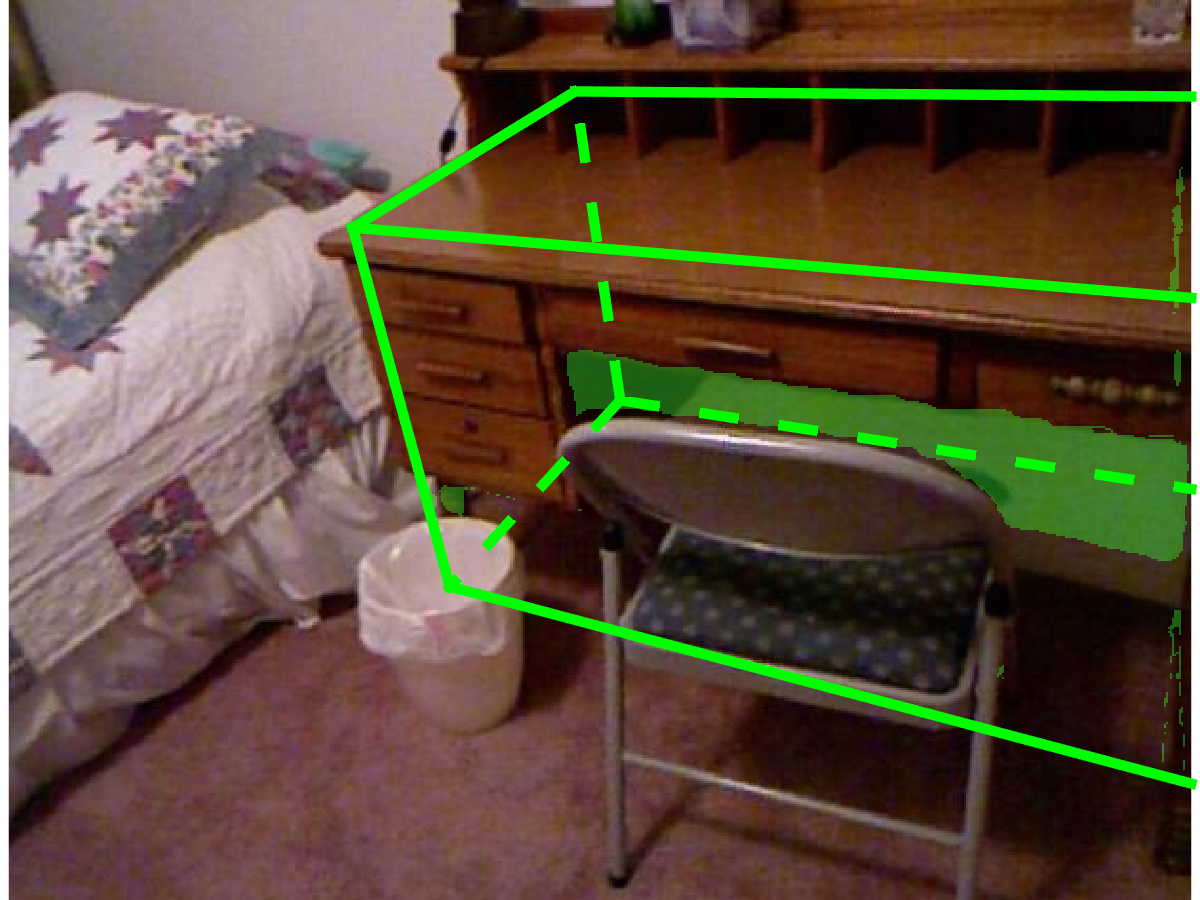}~%
\includegraphics[width=\mW\linewidth]{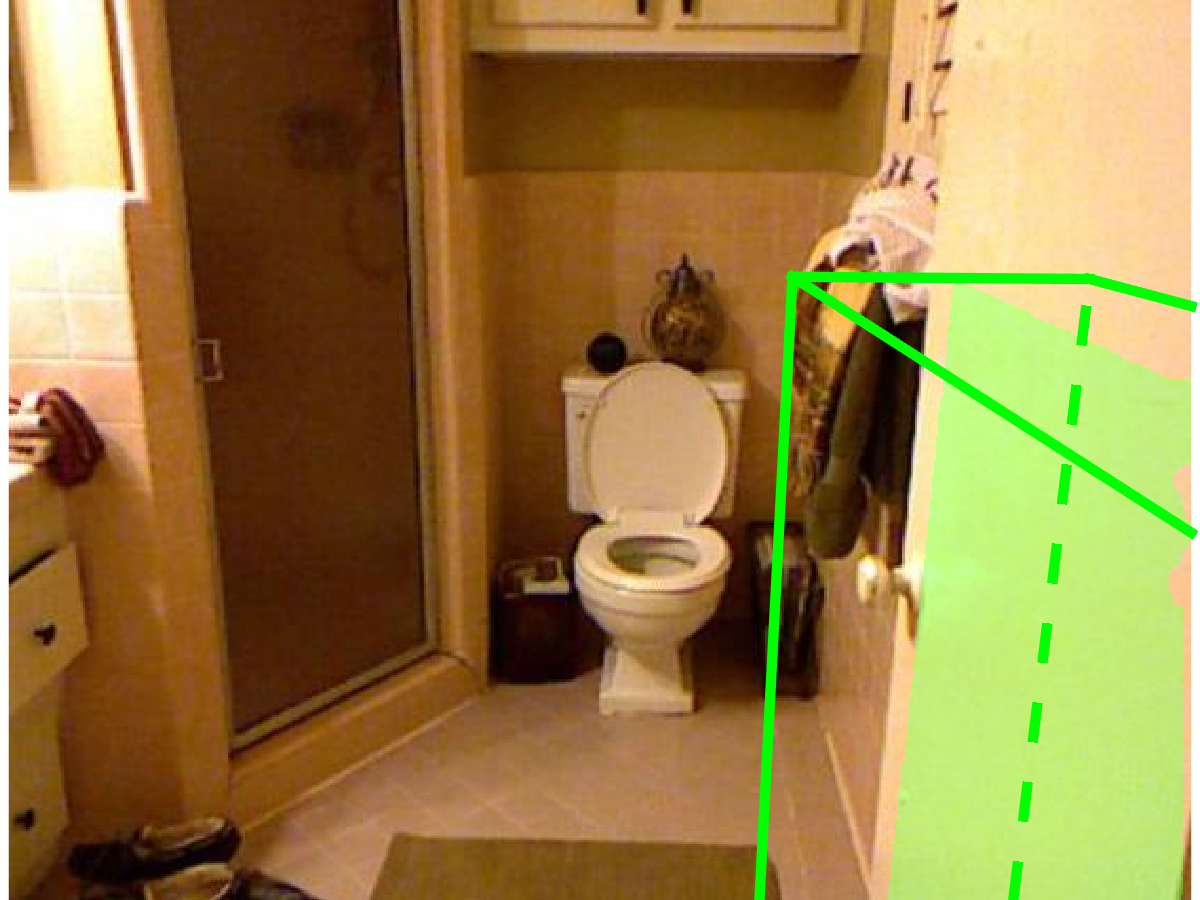}~%
\includegraphics[width=\mW\linewidth]{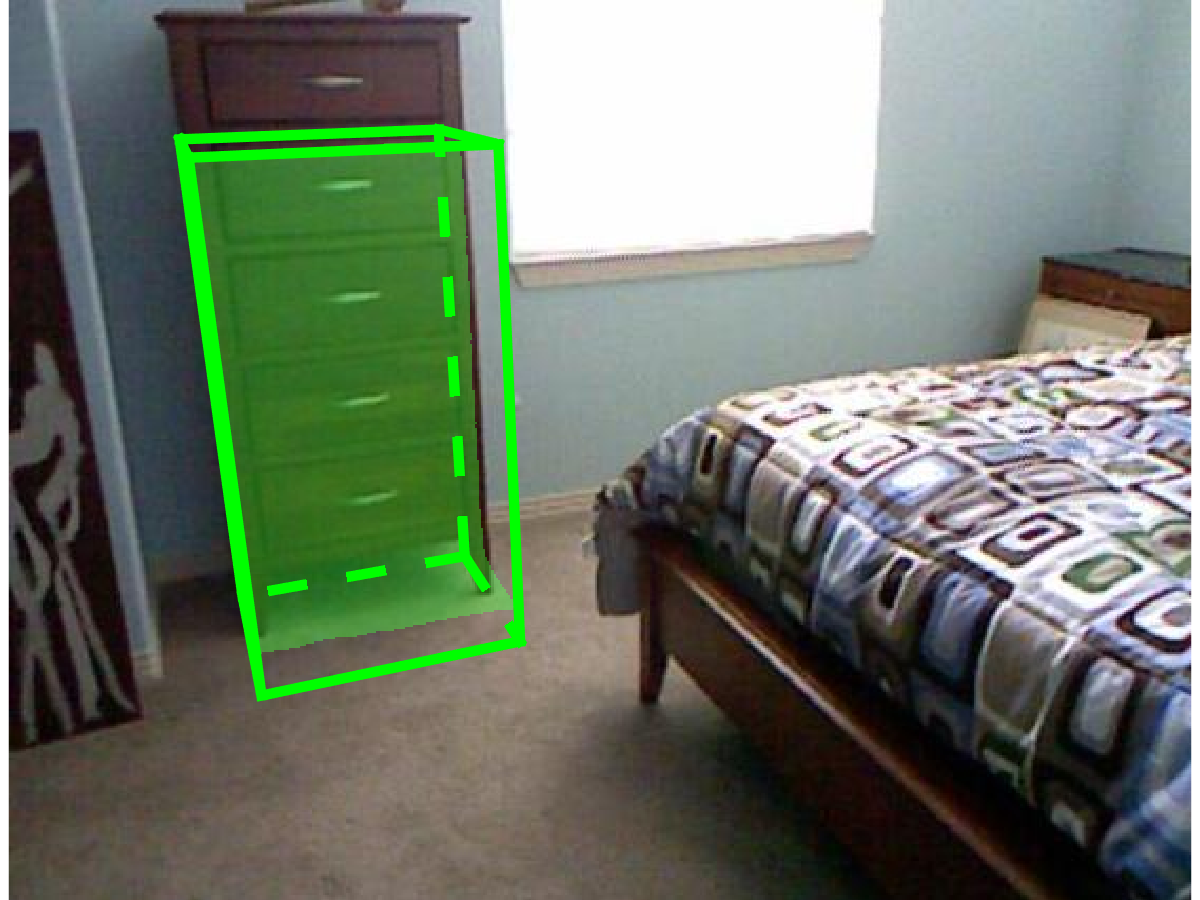}~%
\includegraphics[width=\mW\linewidth]{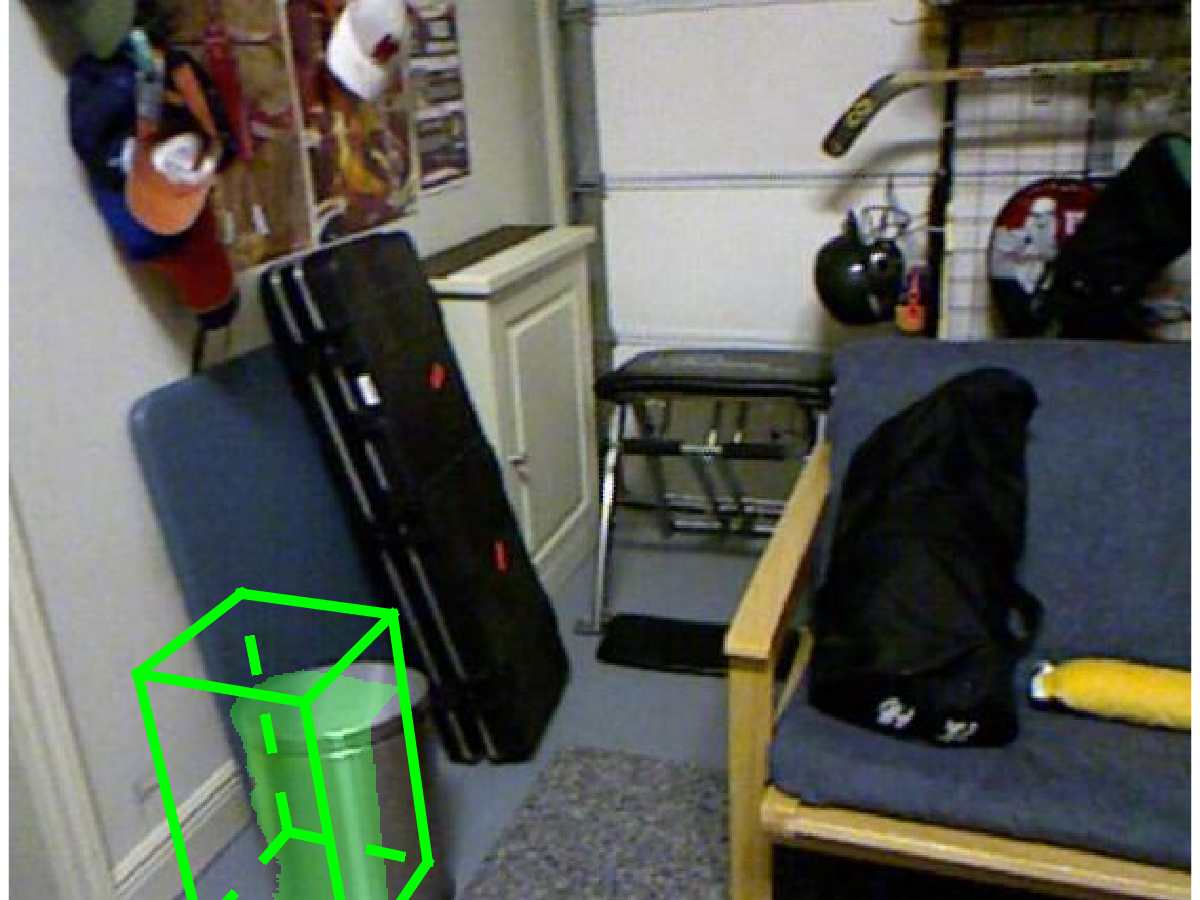}

\includegraphics[width=\mW\linewidth]{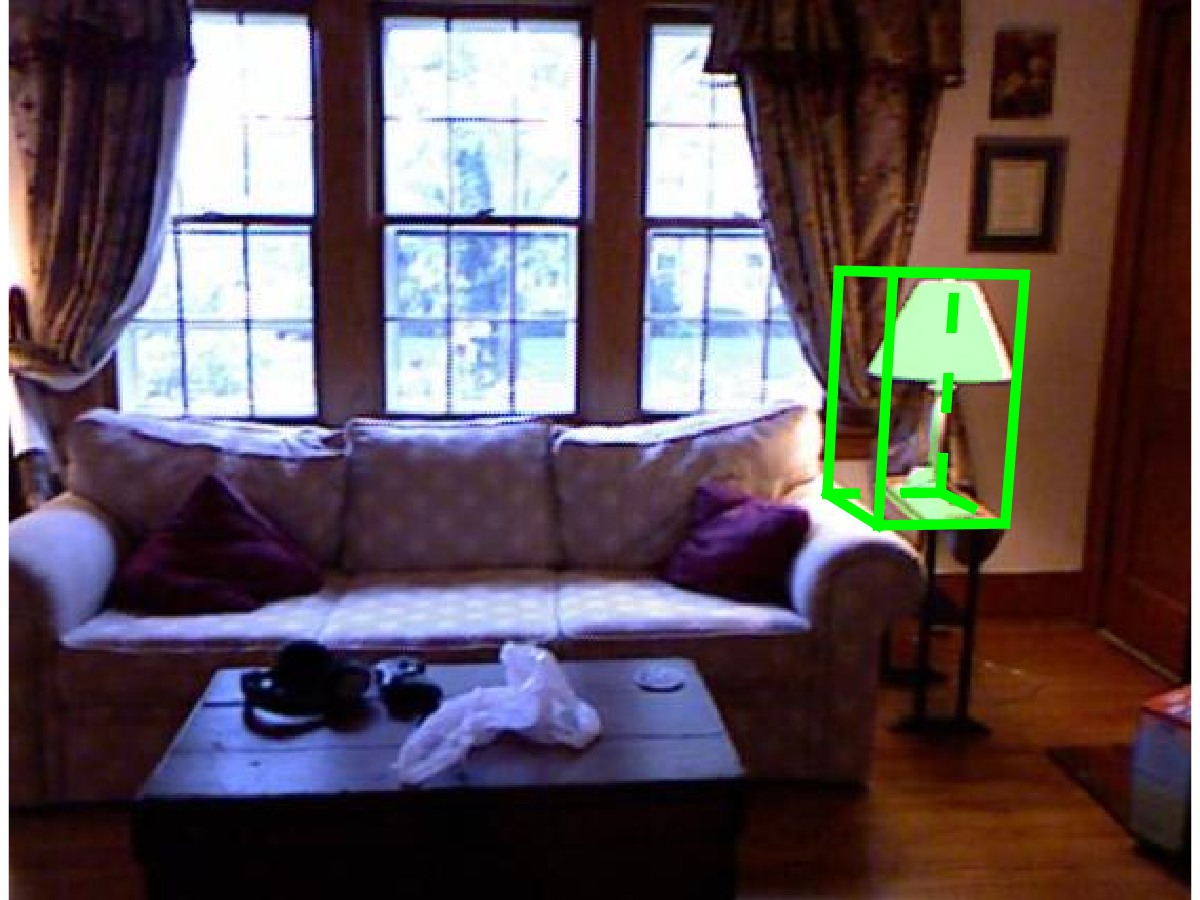}~%
\includegraphics[width=\mW\linewidth]{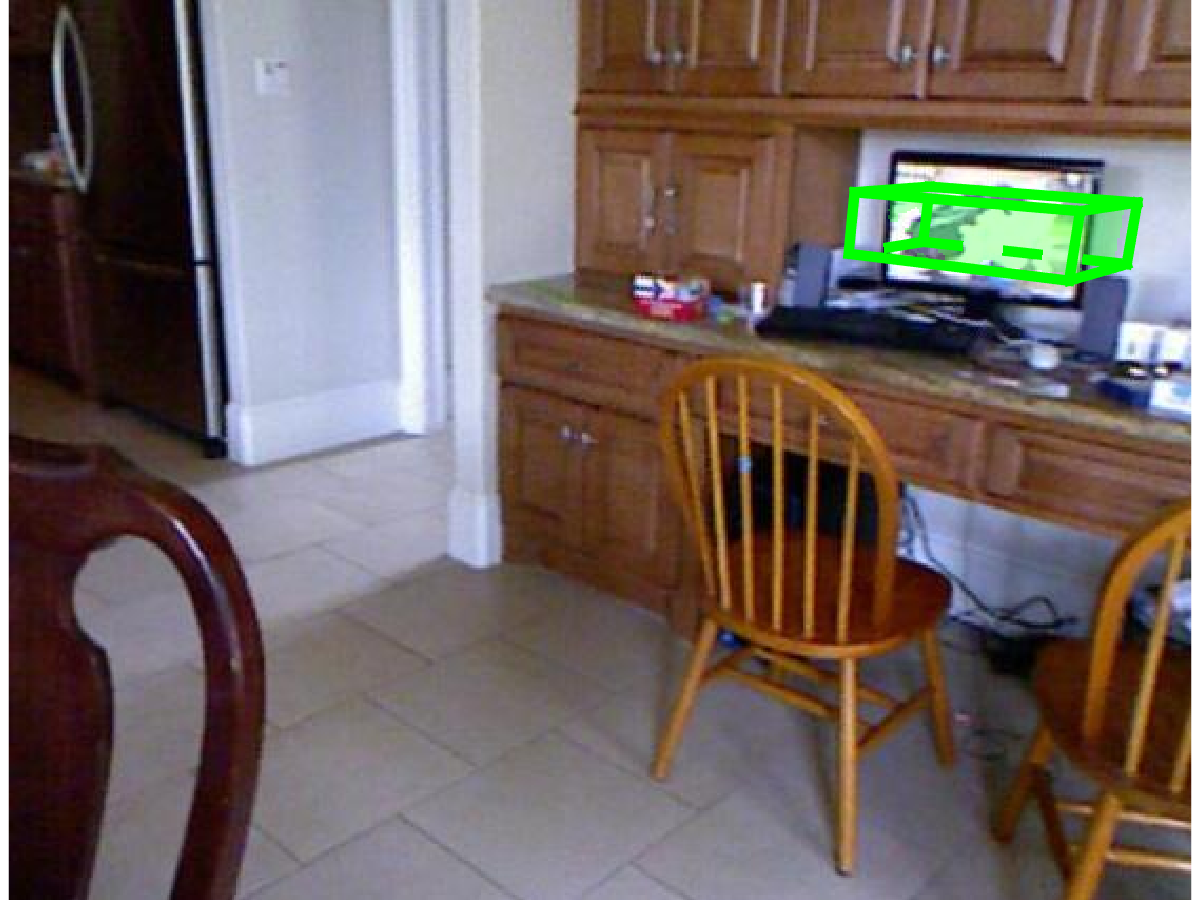}~%
\includegraphics[width=\mW\linewidth]{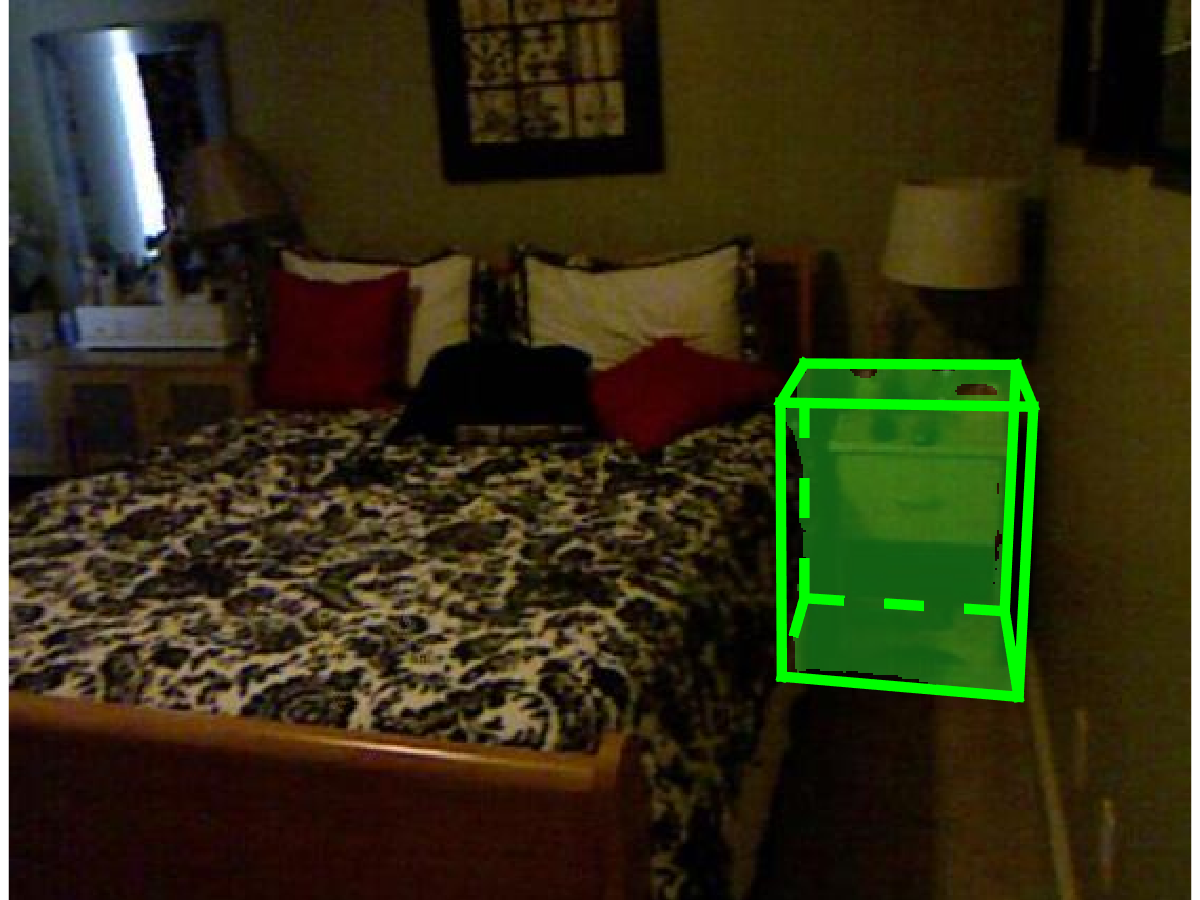}~%
\includegraphics[width=\mW\linewidth]{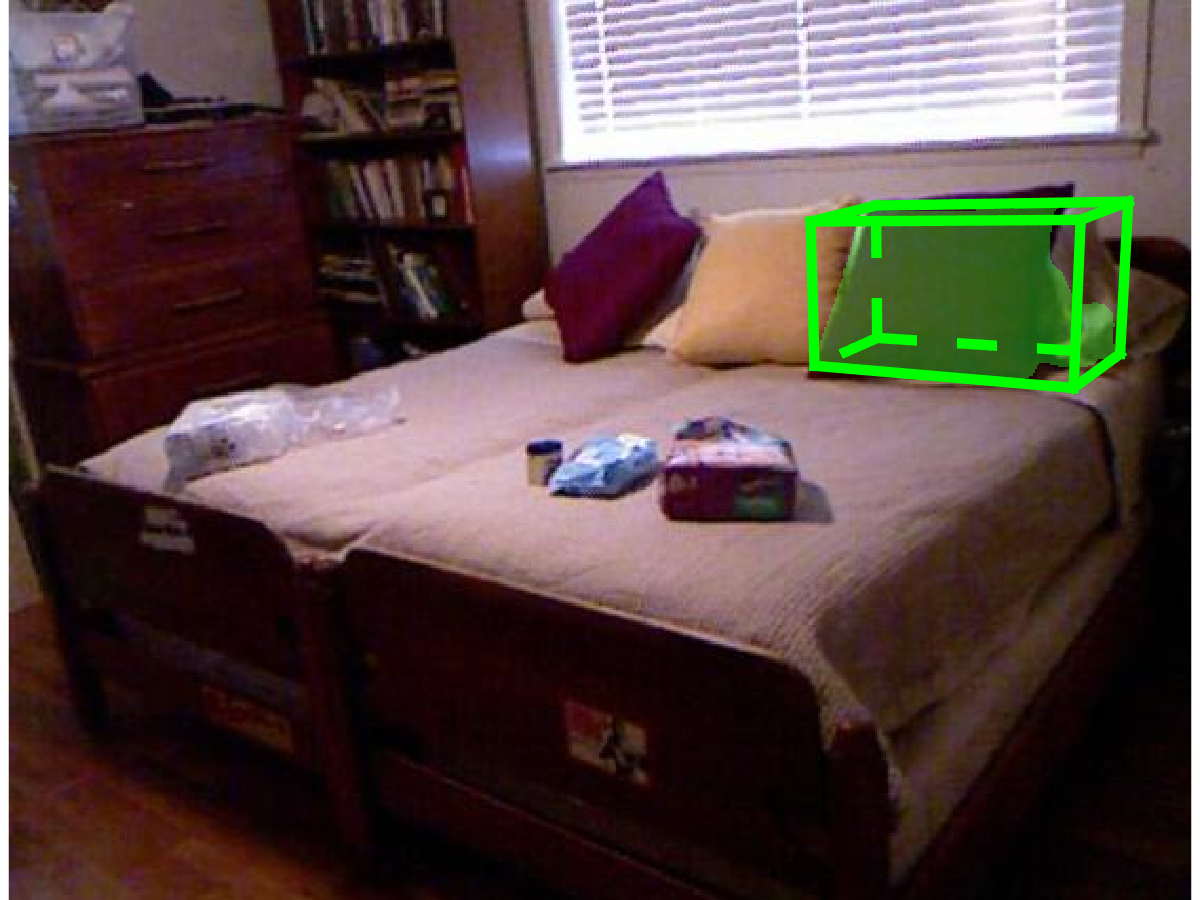}~%
\includegraphics[width=\mW\linewidth]{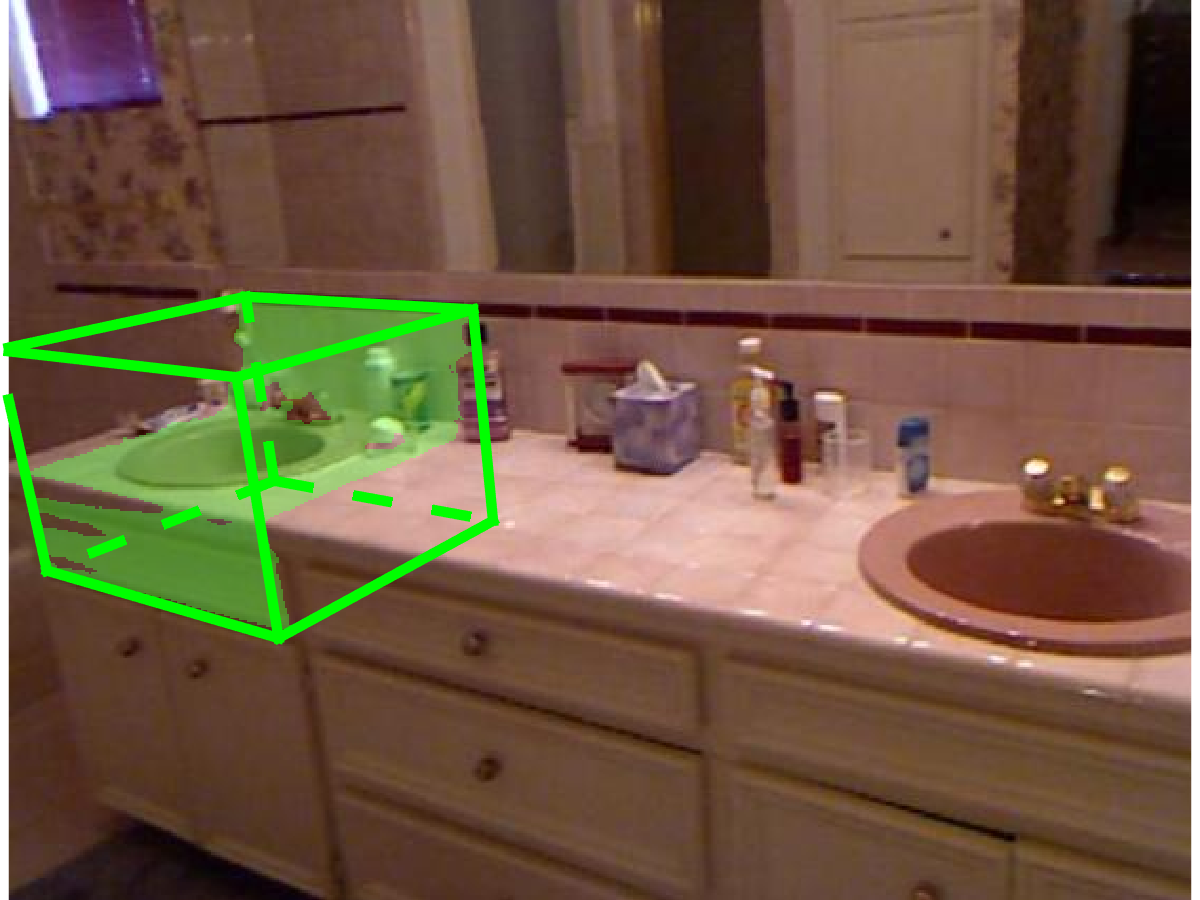}~%
\includegraphics[width=\mW\linewidth]{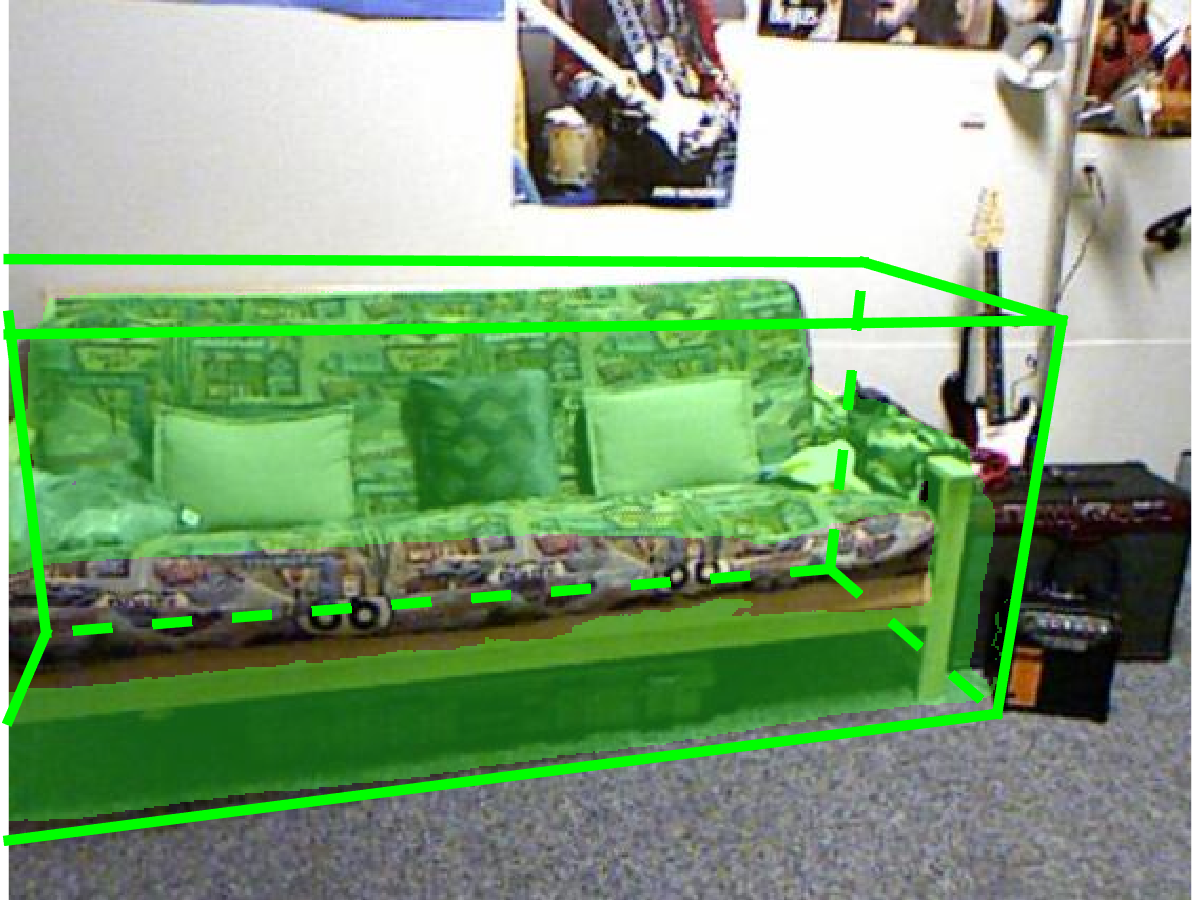}~%
\includegraphics[width=\mW\linewidth]{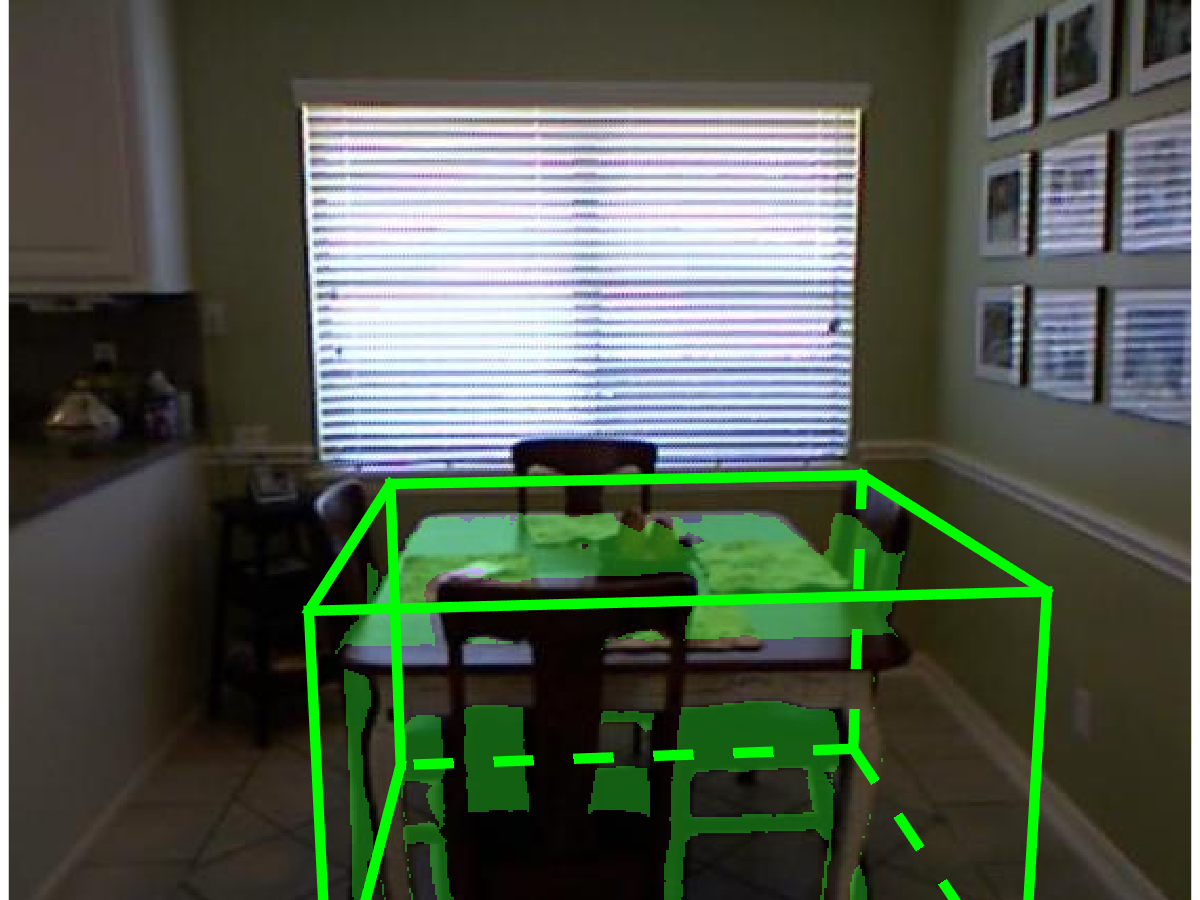}~%
\includegraphics[width=\mW\linewidth]{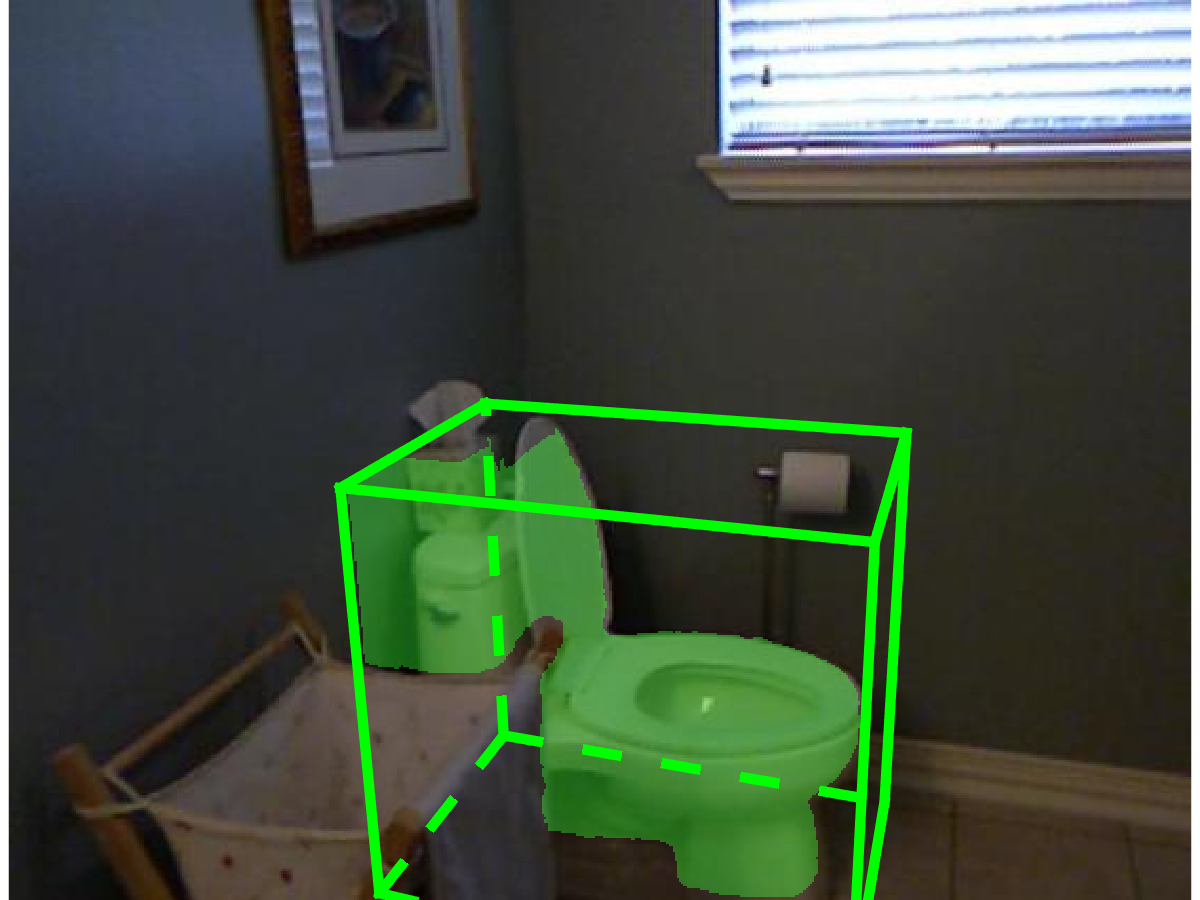}~%
\includegraphics[width=\mW\linewidth]{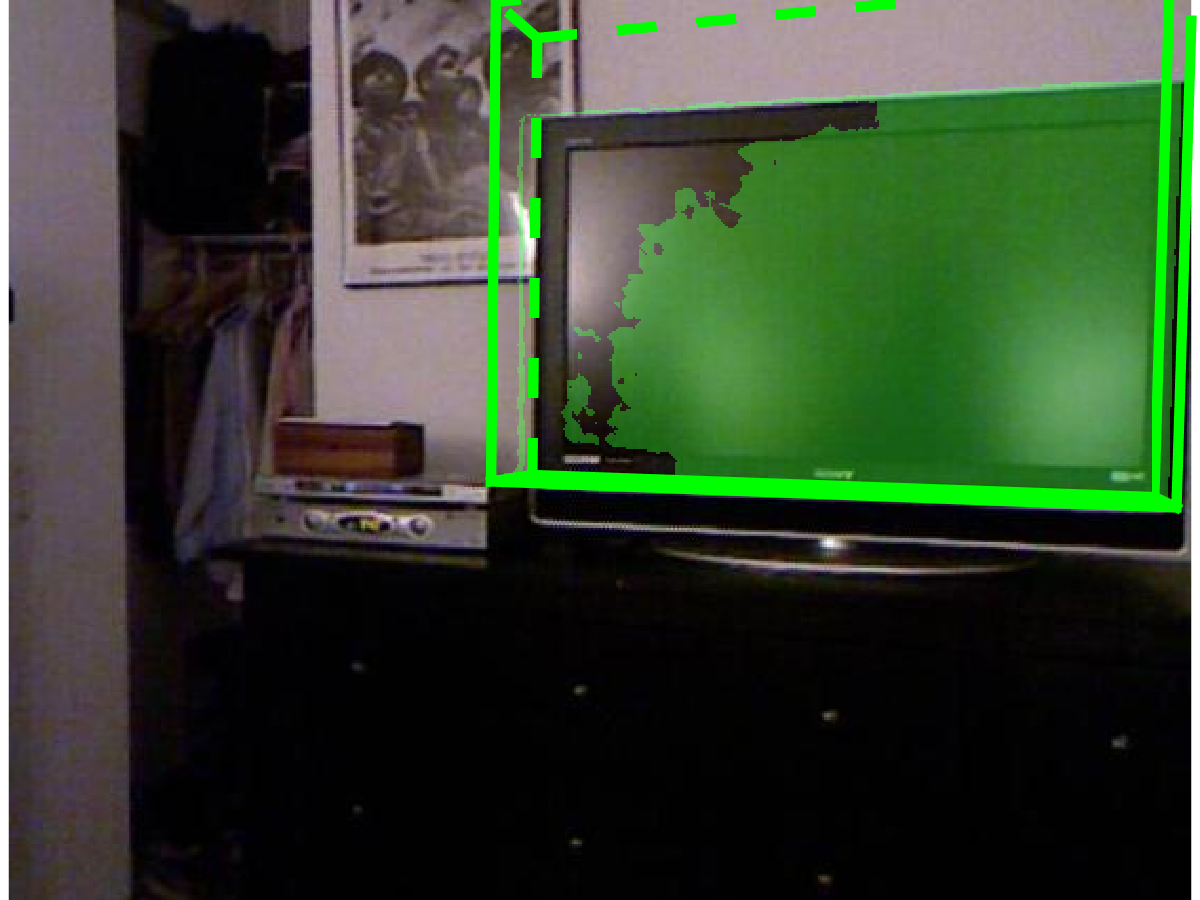}~%
\includegraphics[width=\mW\linewidth]{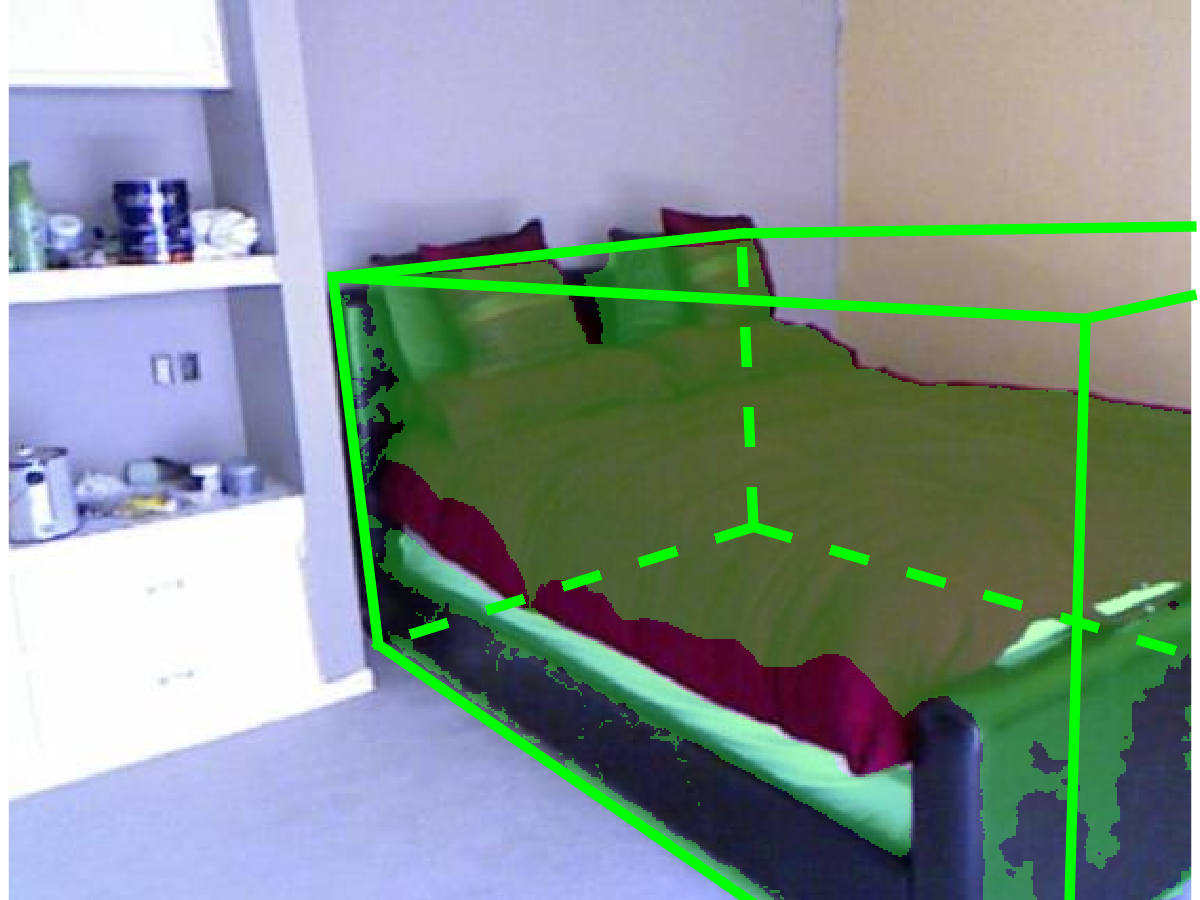}

\vspace{-1mm}

\caption{{\bf Top True Positives.}}
\label{fig:resultTP}

%\end{figure*}
%\begin{figure*}[t]
\vspace{2mm}

%\centering
\includegraphics[width=\mW\linewidth]{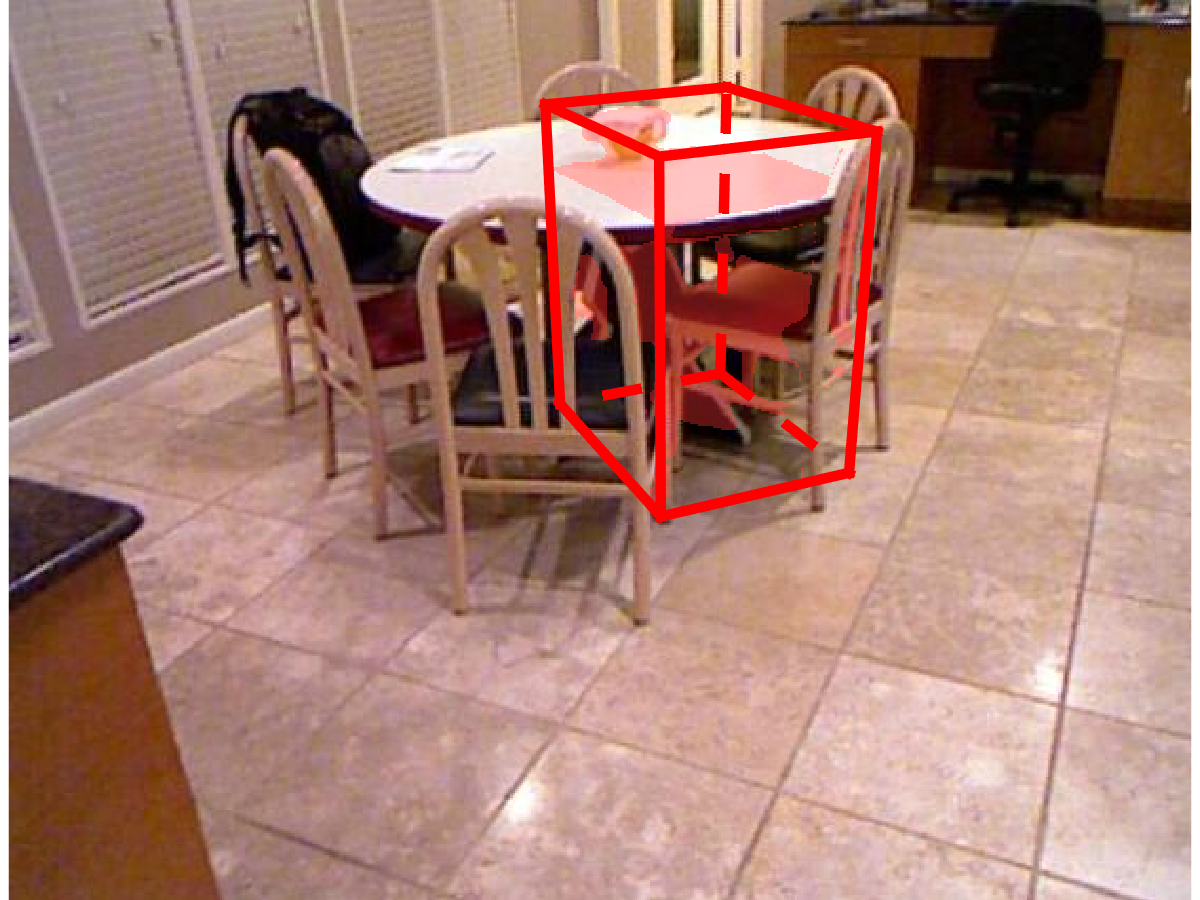}~%
\includegraphics[width=\mW\linewidth]{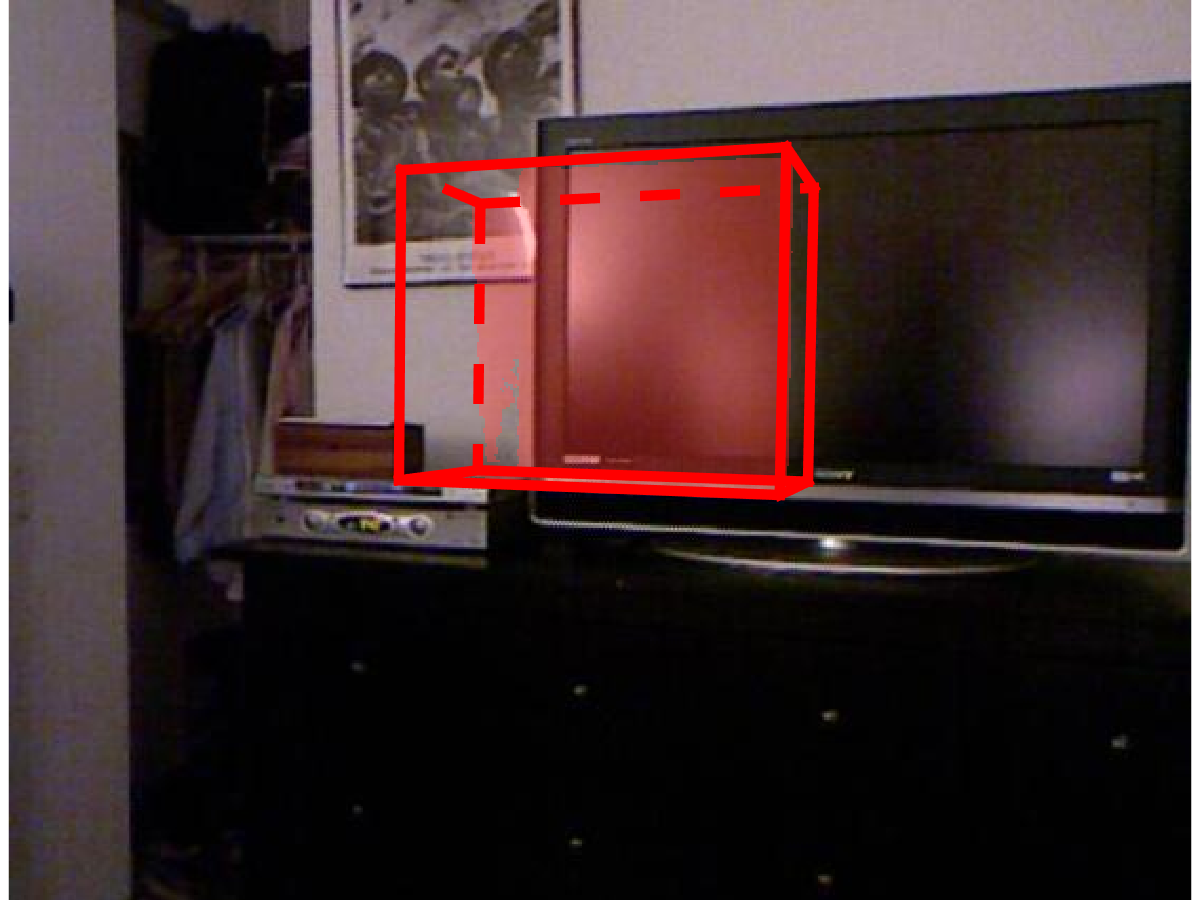}~%
\includegraphics[width=\mW\linewidth]{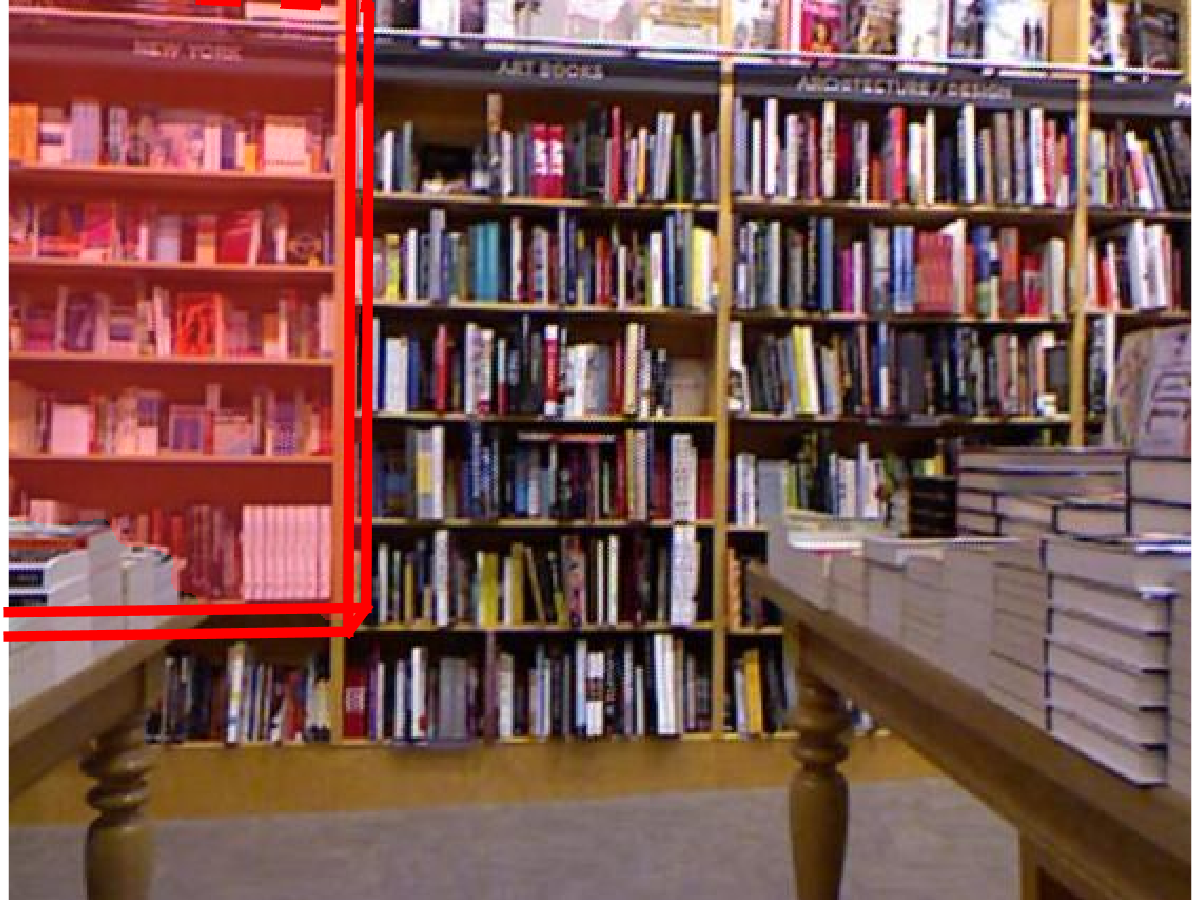}~%
\includegraphics[width=\mW\linewidth]{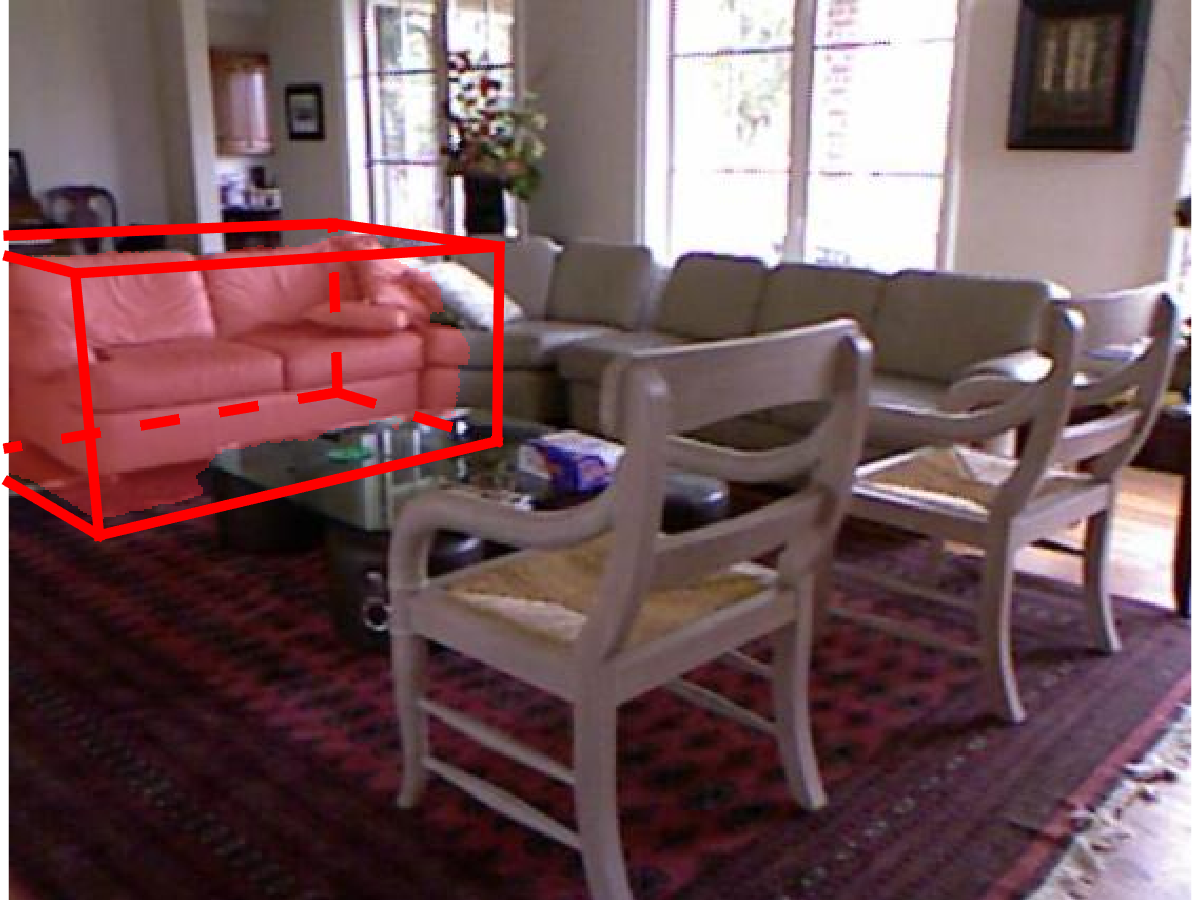}~%
\includegraphics[width=\mW\linewidth]{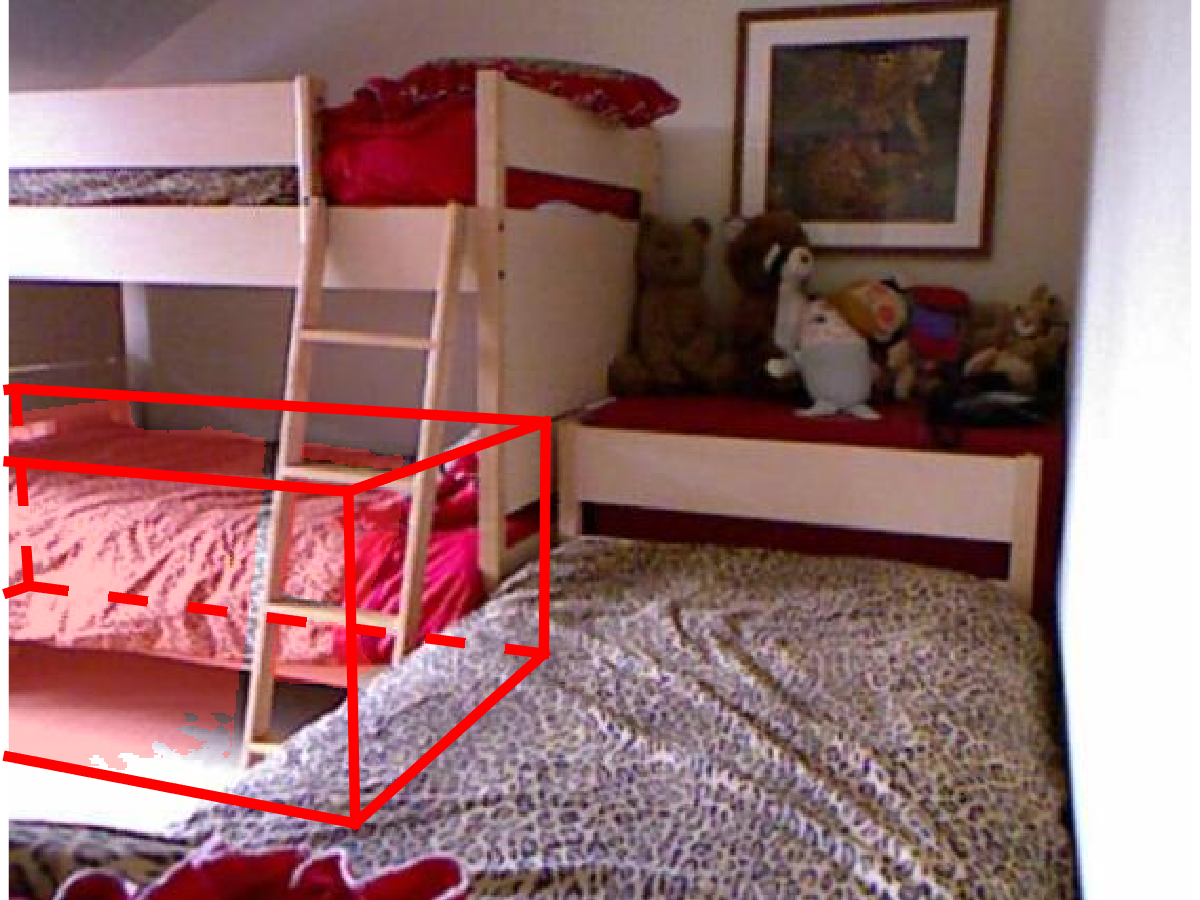}~%
\includegraphics[width=\mW\linewidth]{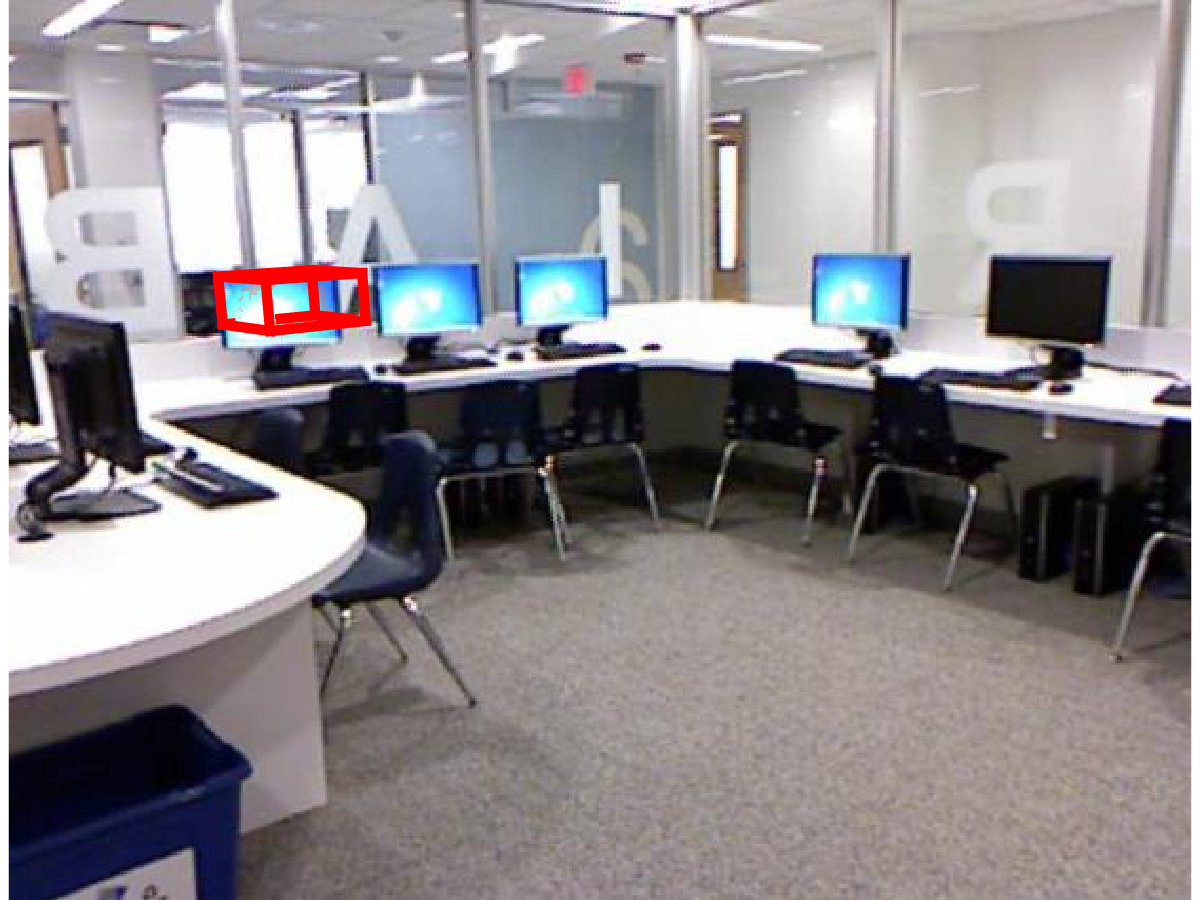}~%
\includegraphics[width=\mW\linewidth]{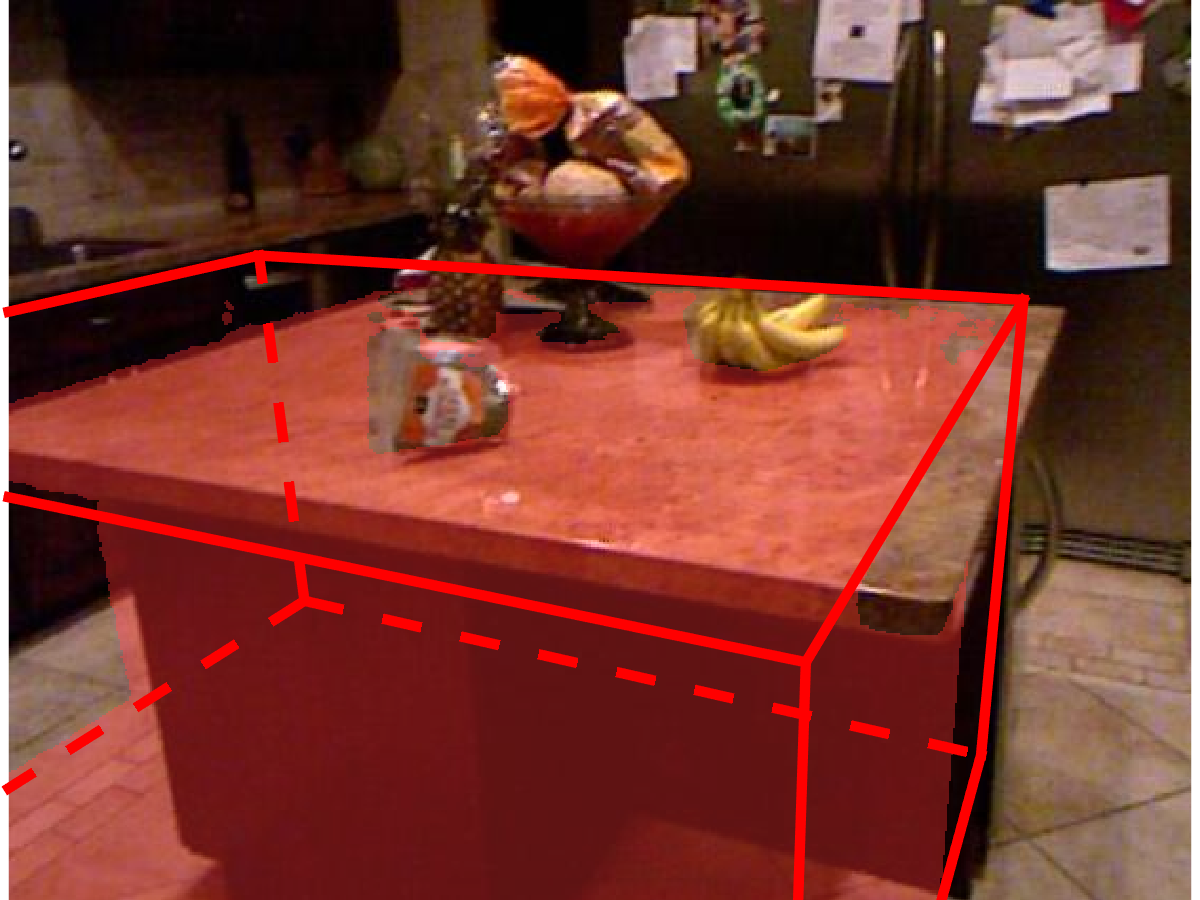}~%
\includegraphics[width=\mW\linewidth]{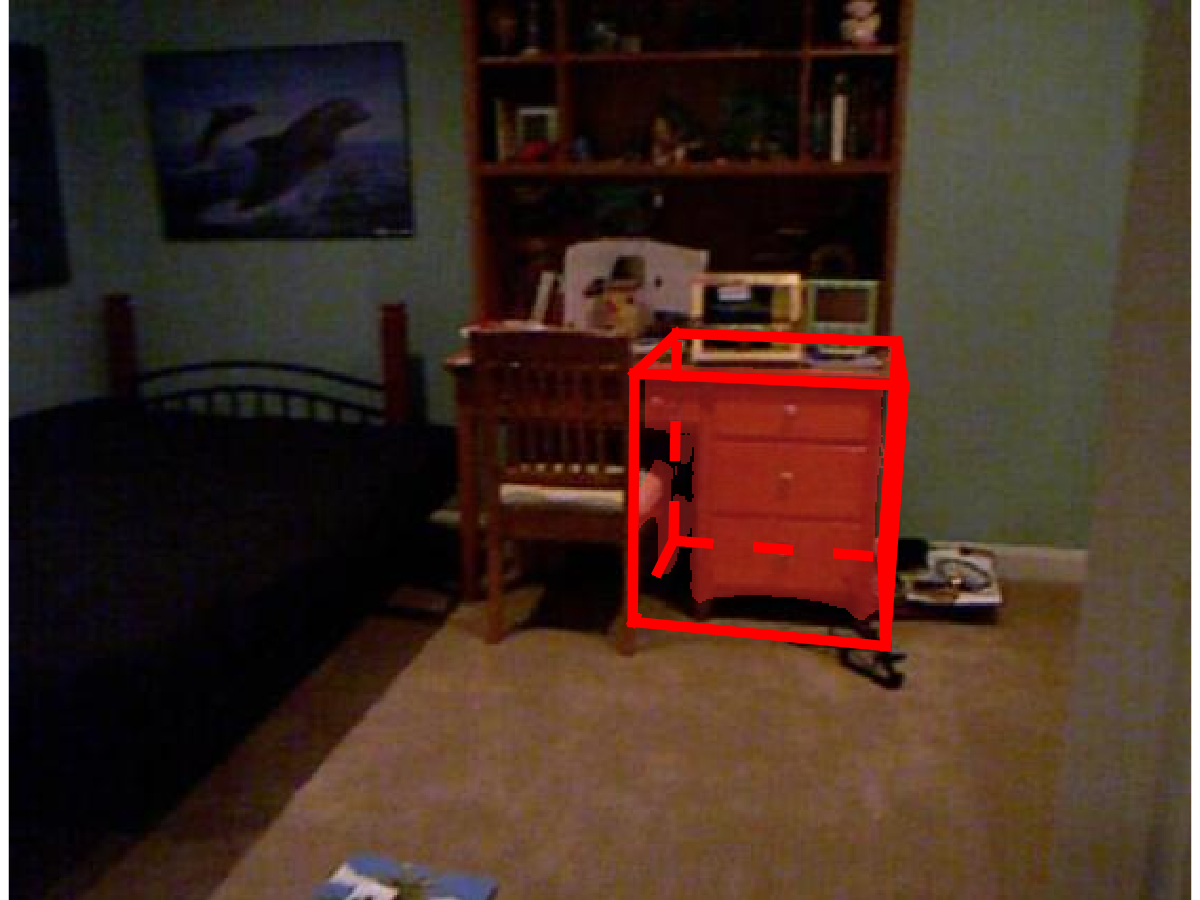}~%
\includegraphics[width=\mW\linewidth]{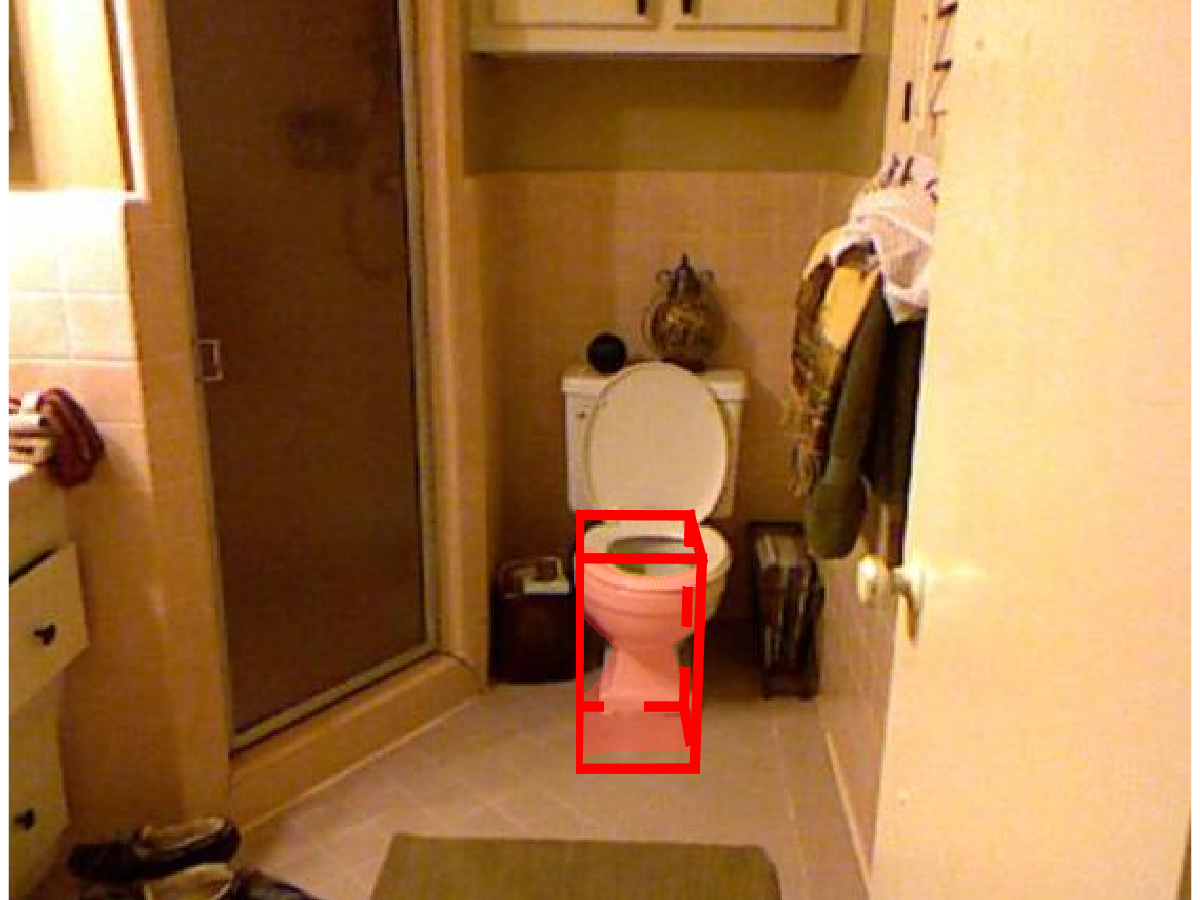}~%
\includegraphics[width=\mW\linewidth]{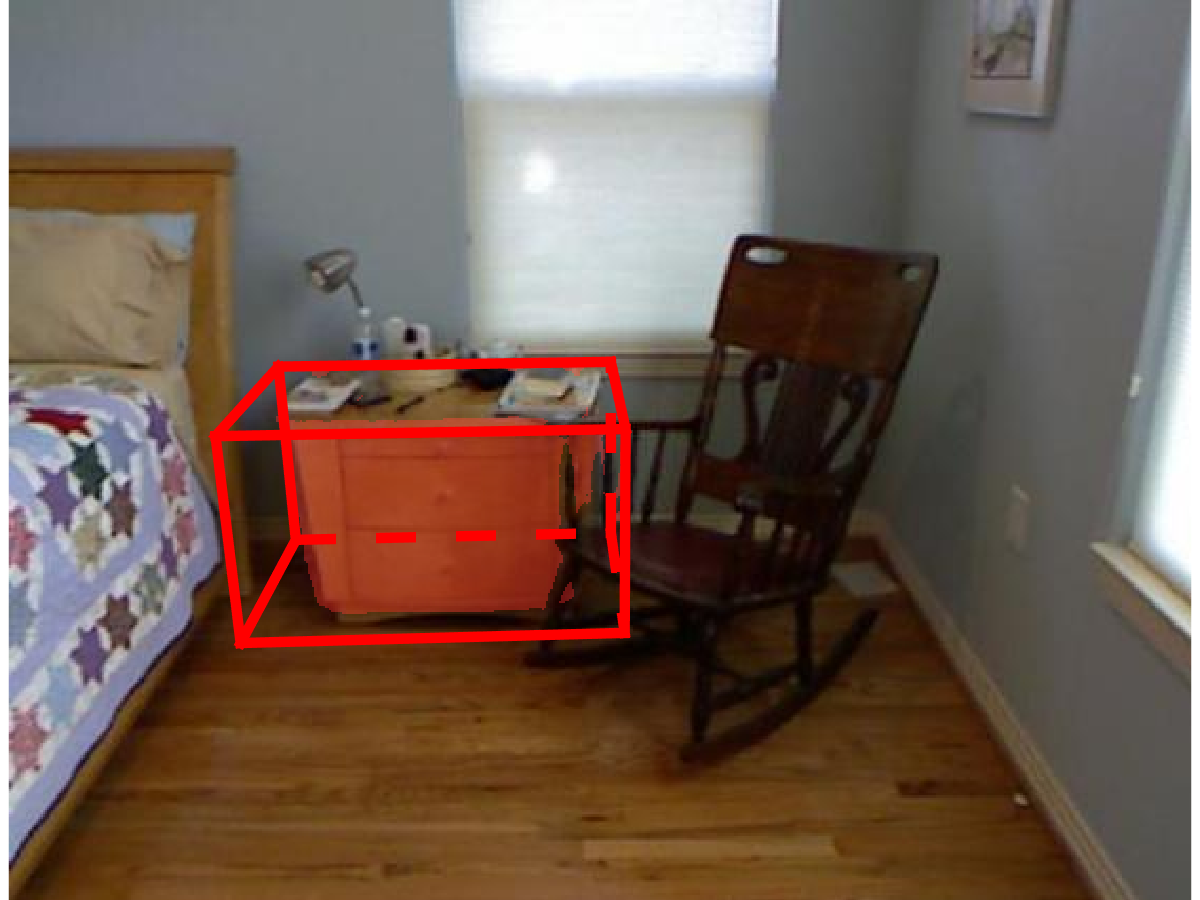}

\vspace{-0.5mm}

{\footnotesize

~~~~(1) chair~~~~~~~~~~~~~~~(2) tv~~~~~~~~~(3) bookshelf~~~~~~~~(4) sofa~~~~~~~~~~~~(5) bed~~~~~~~~~~~(6) monitor~~~~~~~~~~(7) desk~~~~~(8) night stand~~(9) garbage bin~~~~~(10) box

}

%\vspace{-1mm}

\caption{{\bf Top False Positives.}
(1)-(2) show detections with inaccurate locations.
(3)-(6) show detections with wrong box size for the big bookshelf, L-shape sofa, bunk bed, and monitor. 
(7)-(10) show detections with wrong categories.}
\label{fig:resultFP}

%\end{figure*}
%\begin{figure*}[]

\vspace{2mm}

%\vspace{1mm}
%\centering
\includegraphics[width=\mW\linewidth]{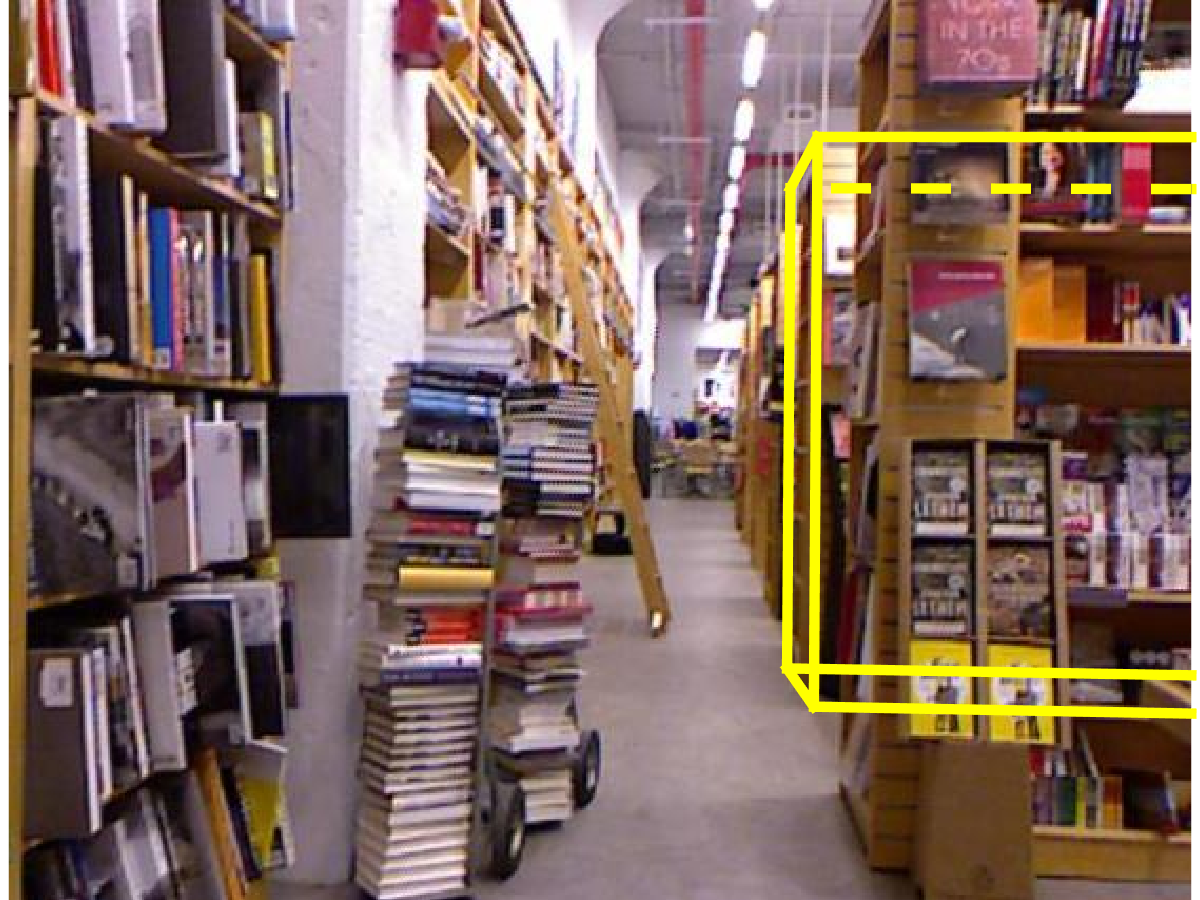}~%
\includegraphics[width=\mW\linewidth]{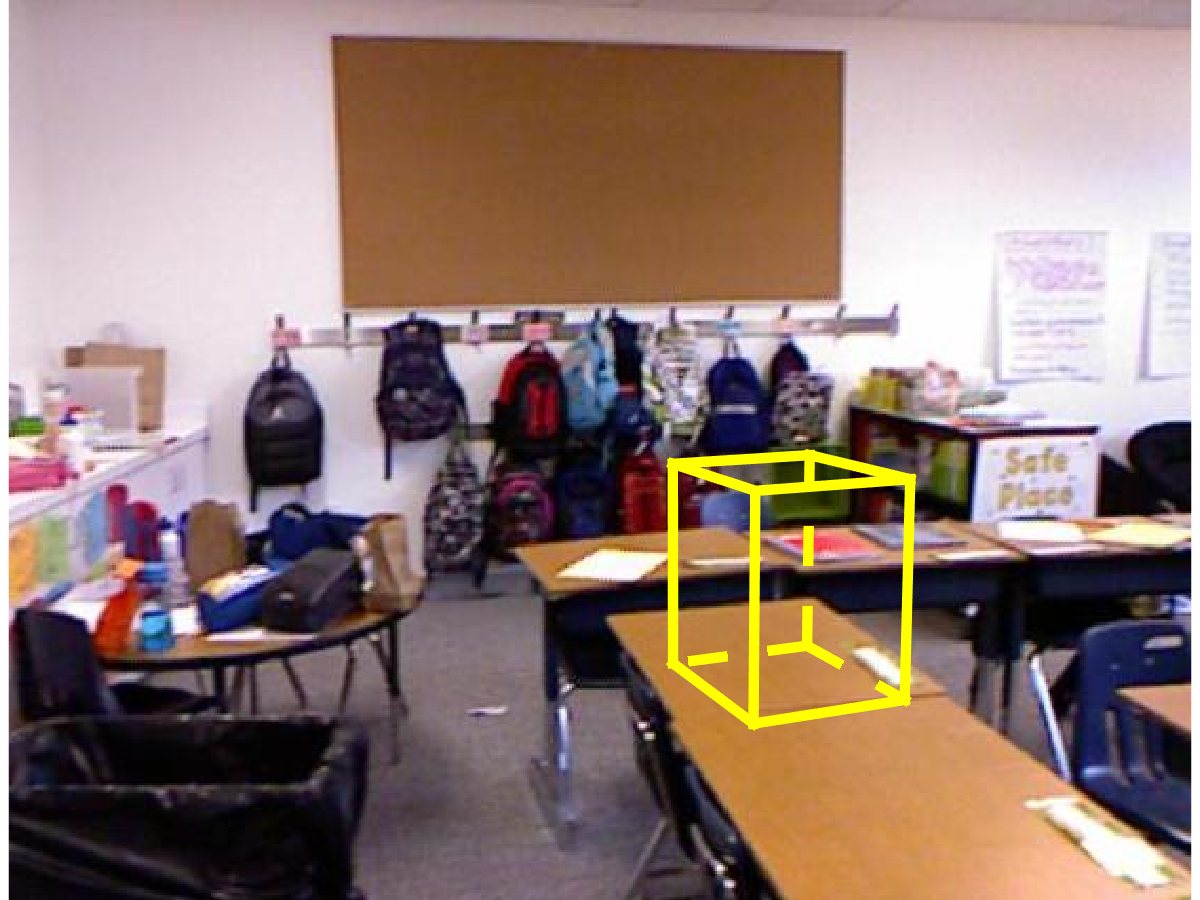}~%
\includegraphics[width=\mW\linewidth]{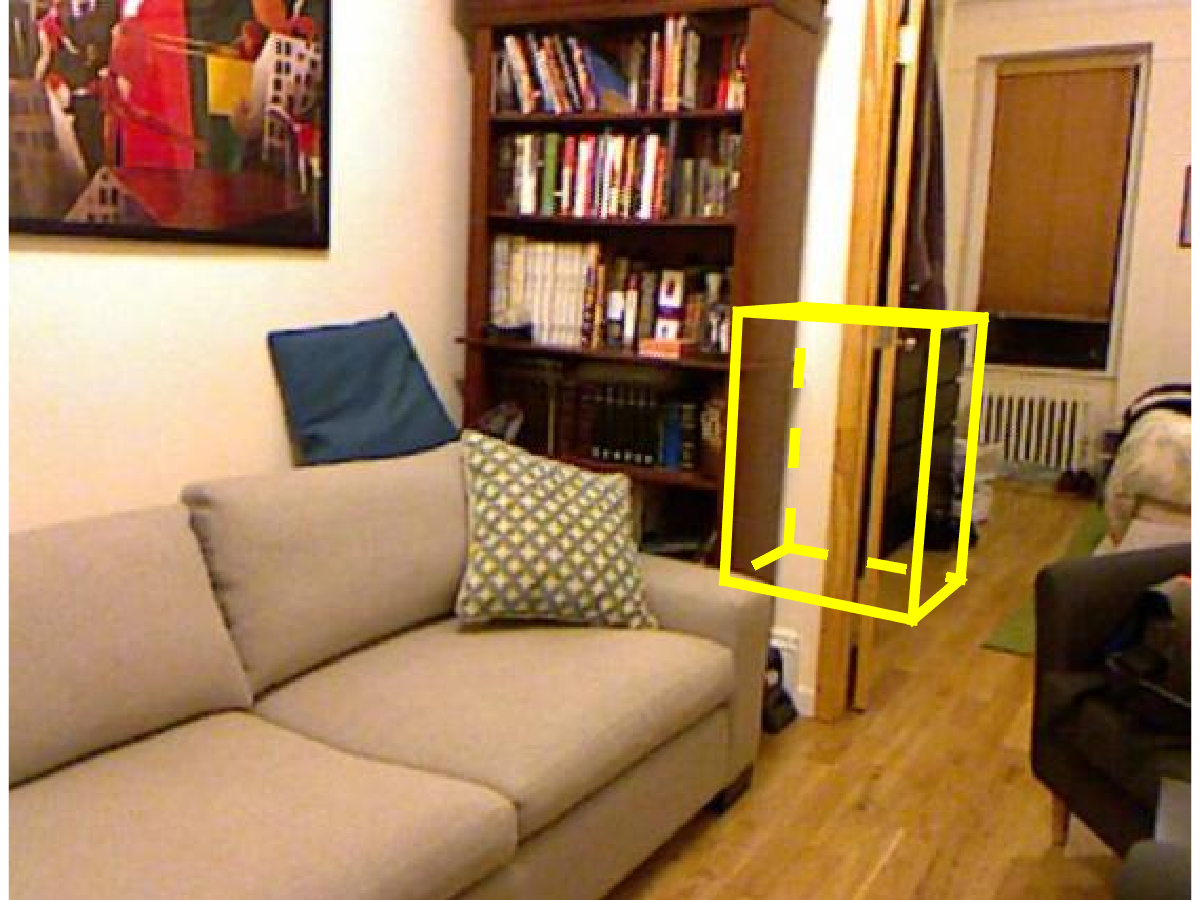}~%
\includegraphics[width=\mW\linewidth]{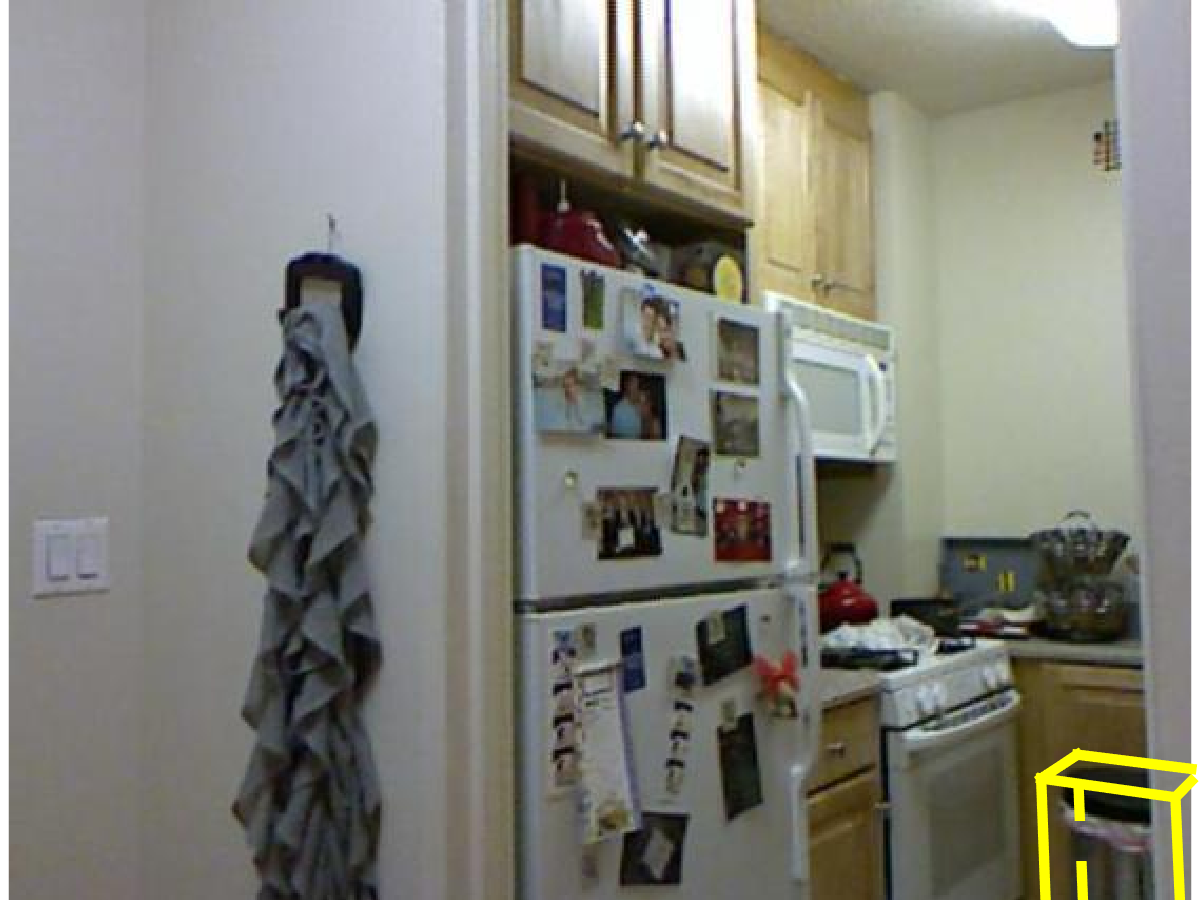}~%
\includegraphics[width=\mW\linewidth]{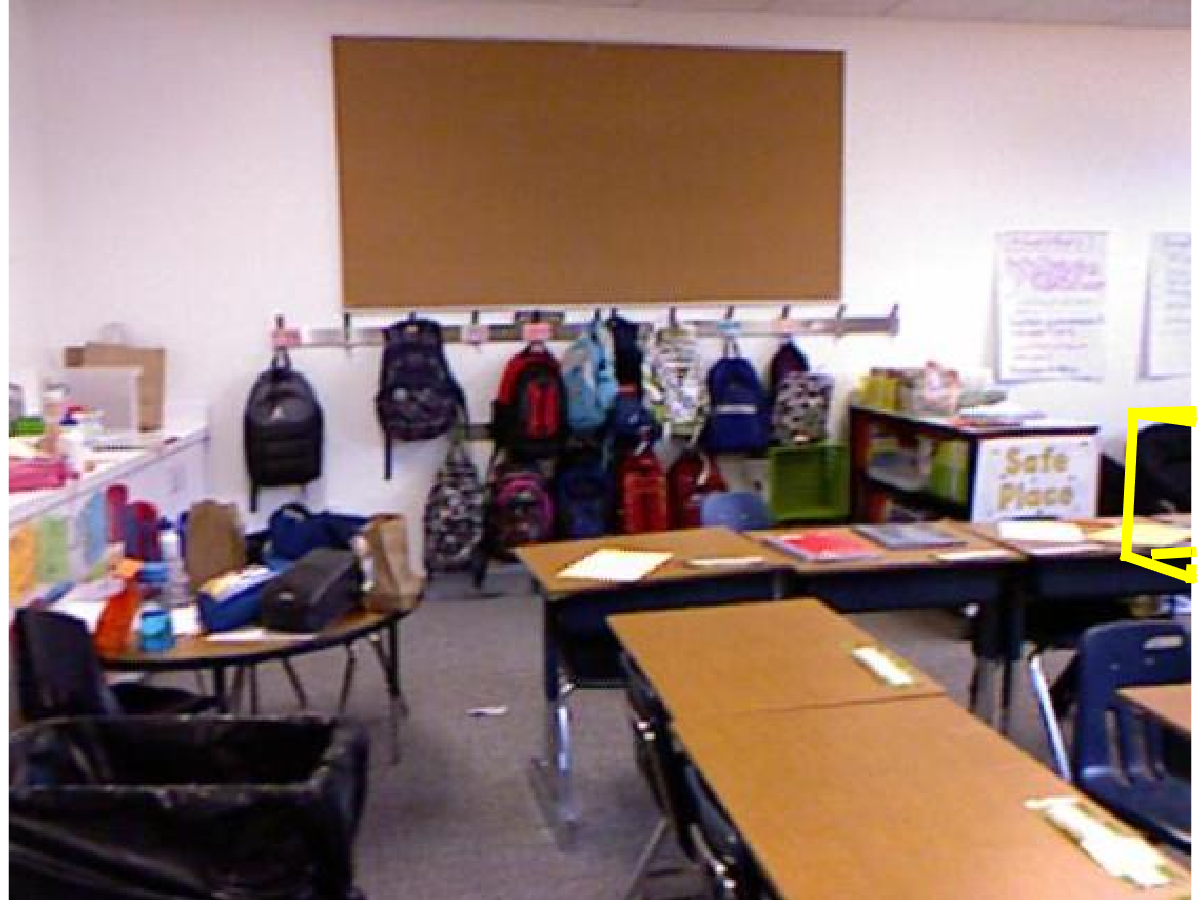}~%
\includegraphics[width=\mW\linewidth]{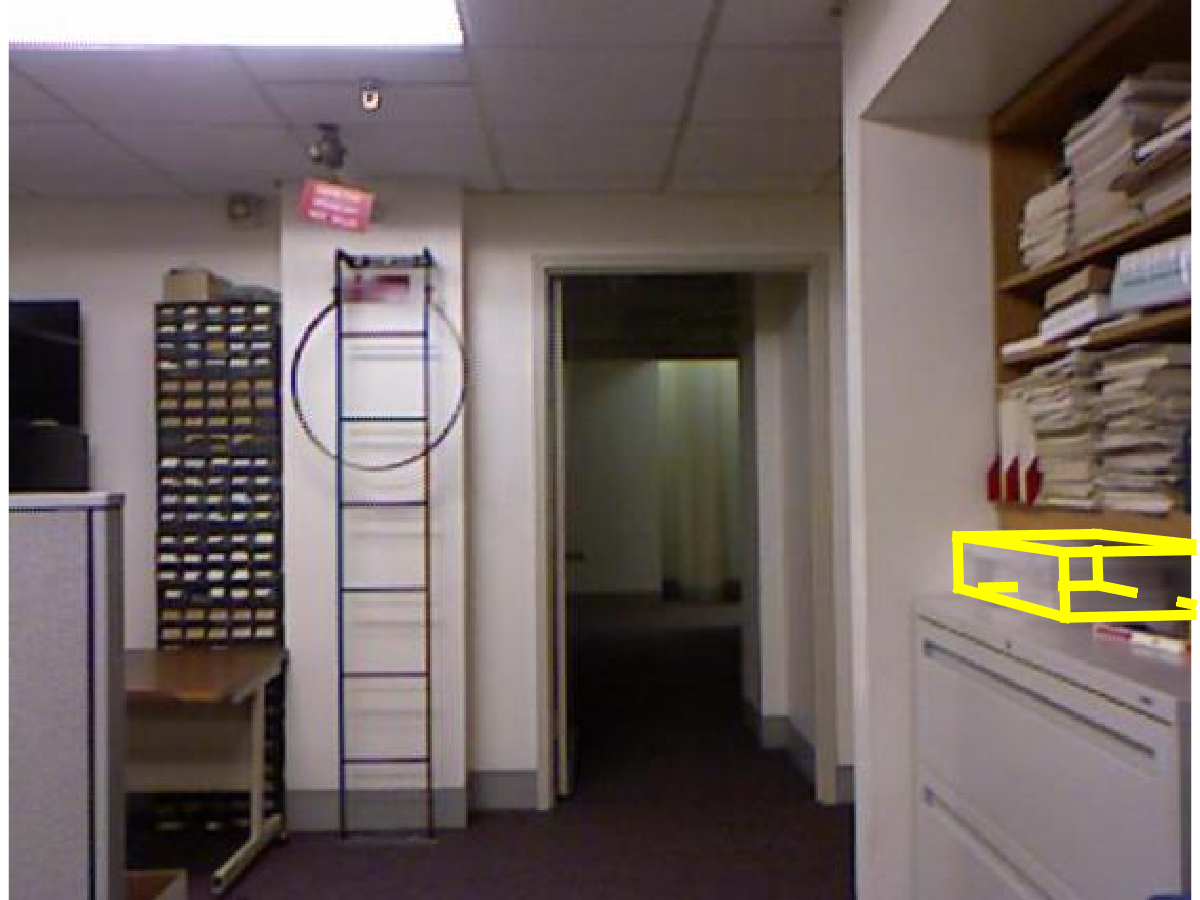}~%
\includegraphics[width=\mW\linewidth]{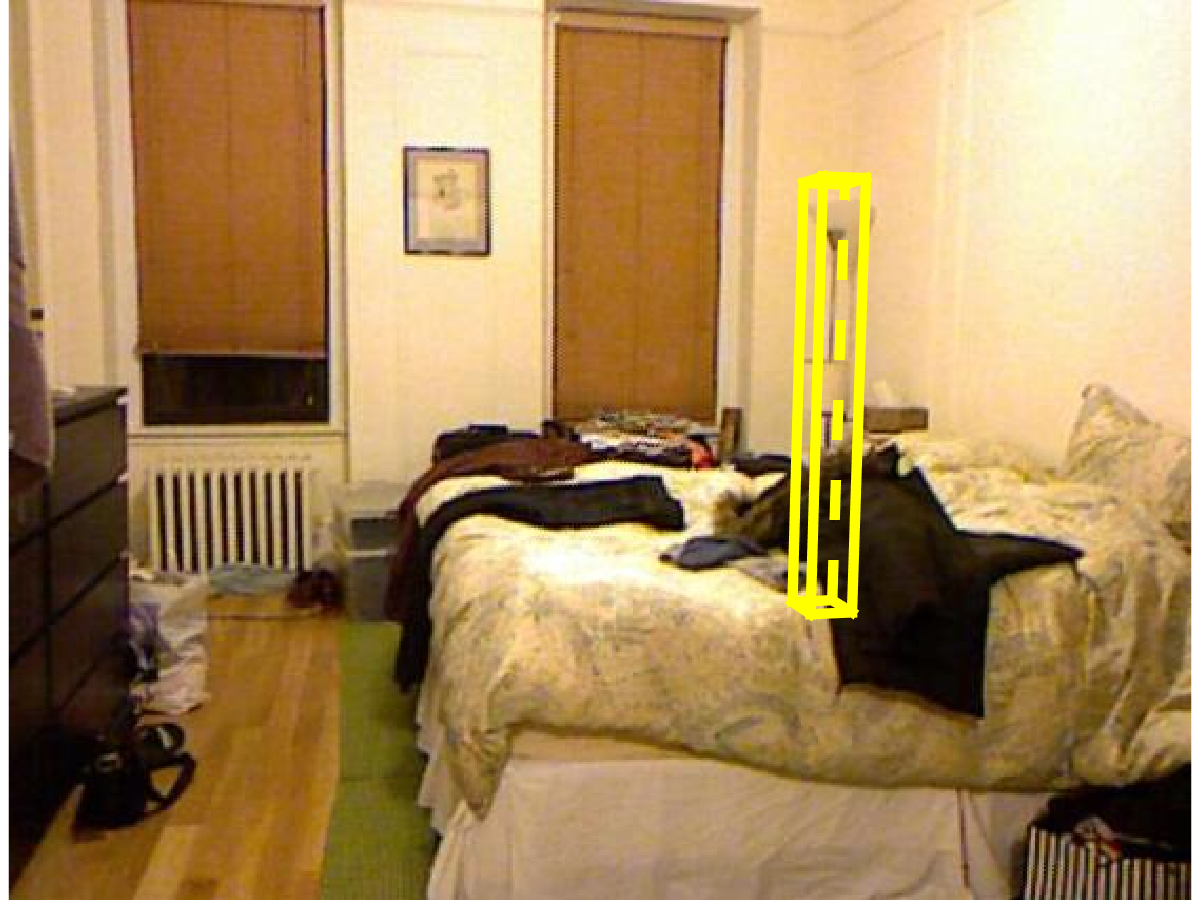}~%
\includegraphics[width=\mW\linewidth]{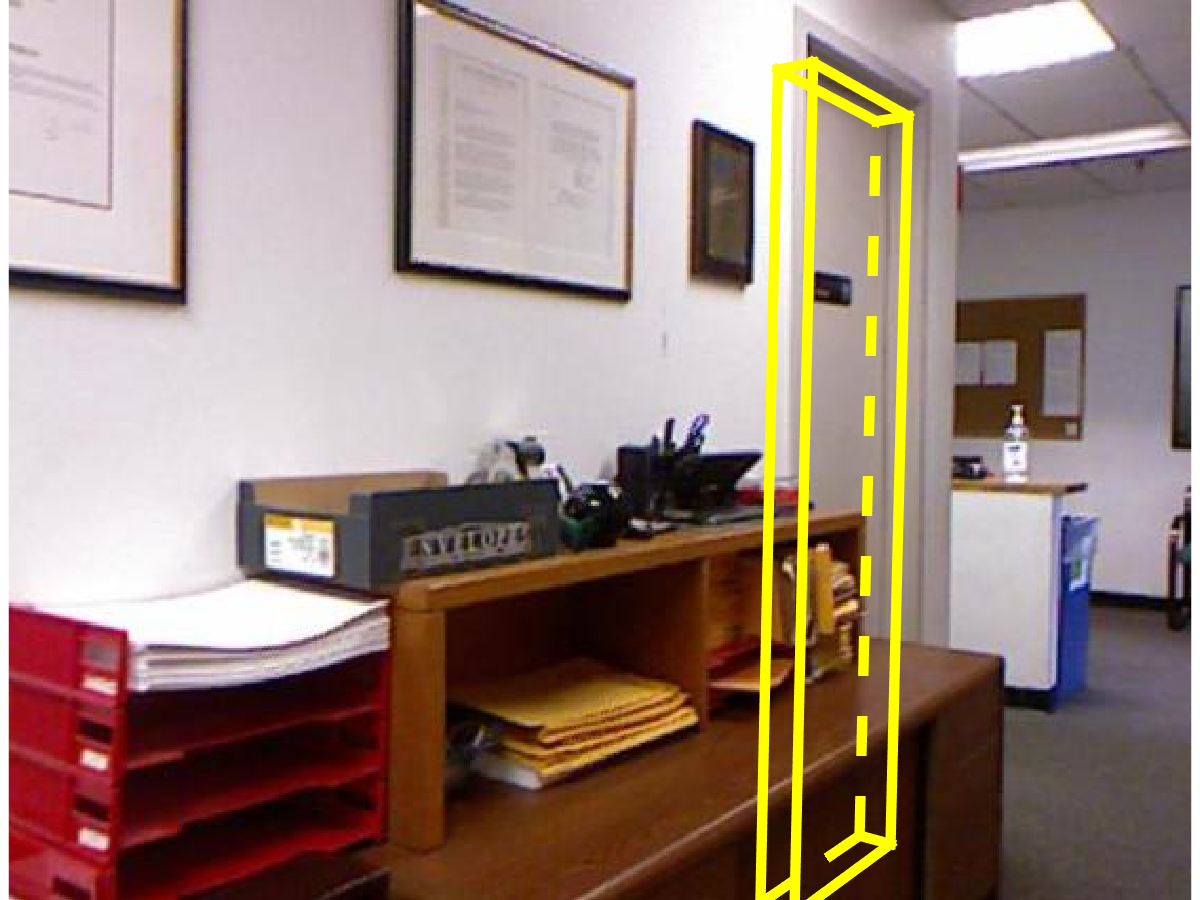}~%
\includegraphics[width=\mW\linewidth]{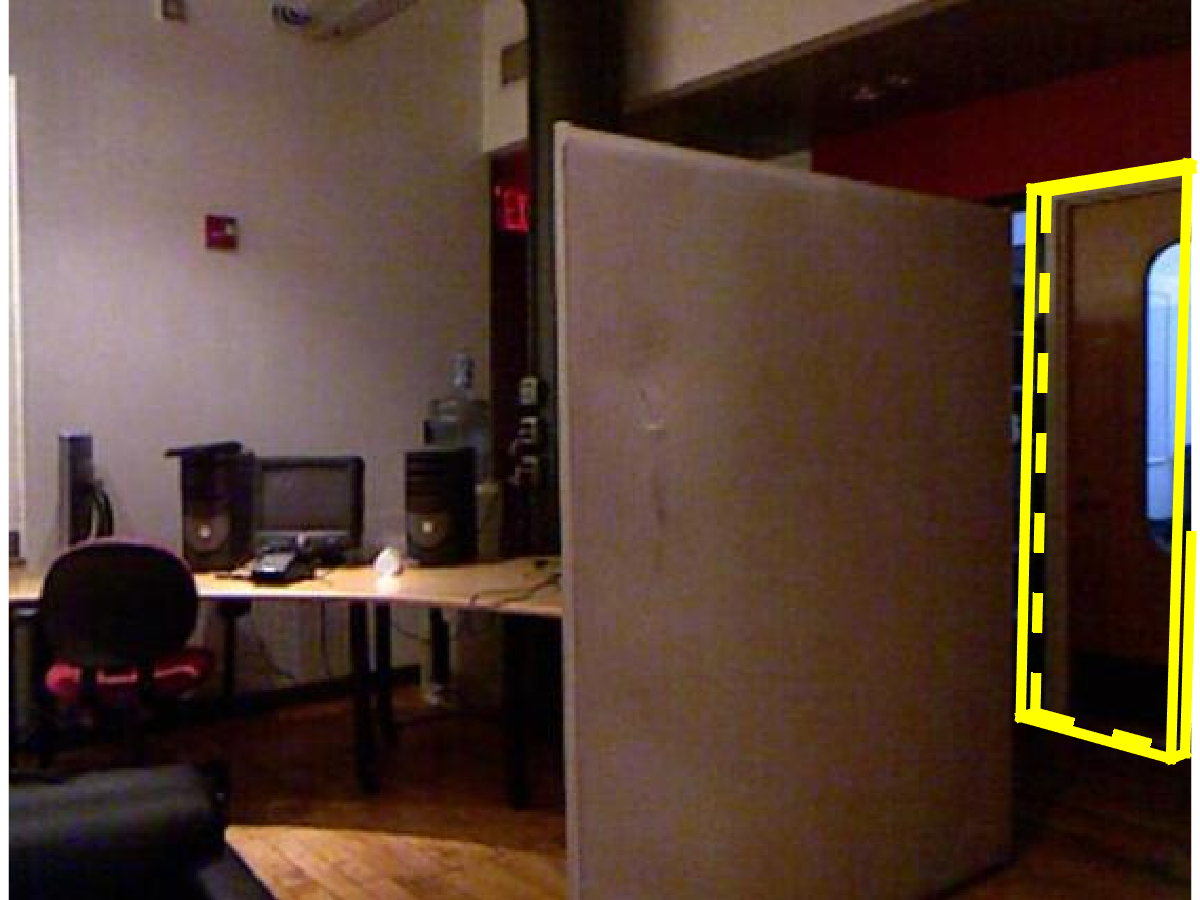}~%
\includegraphics[width=\mW\linewidth]{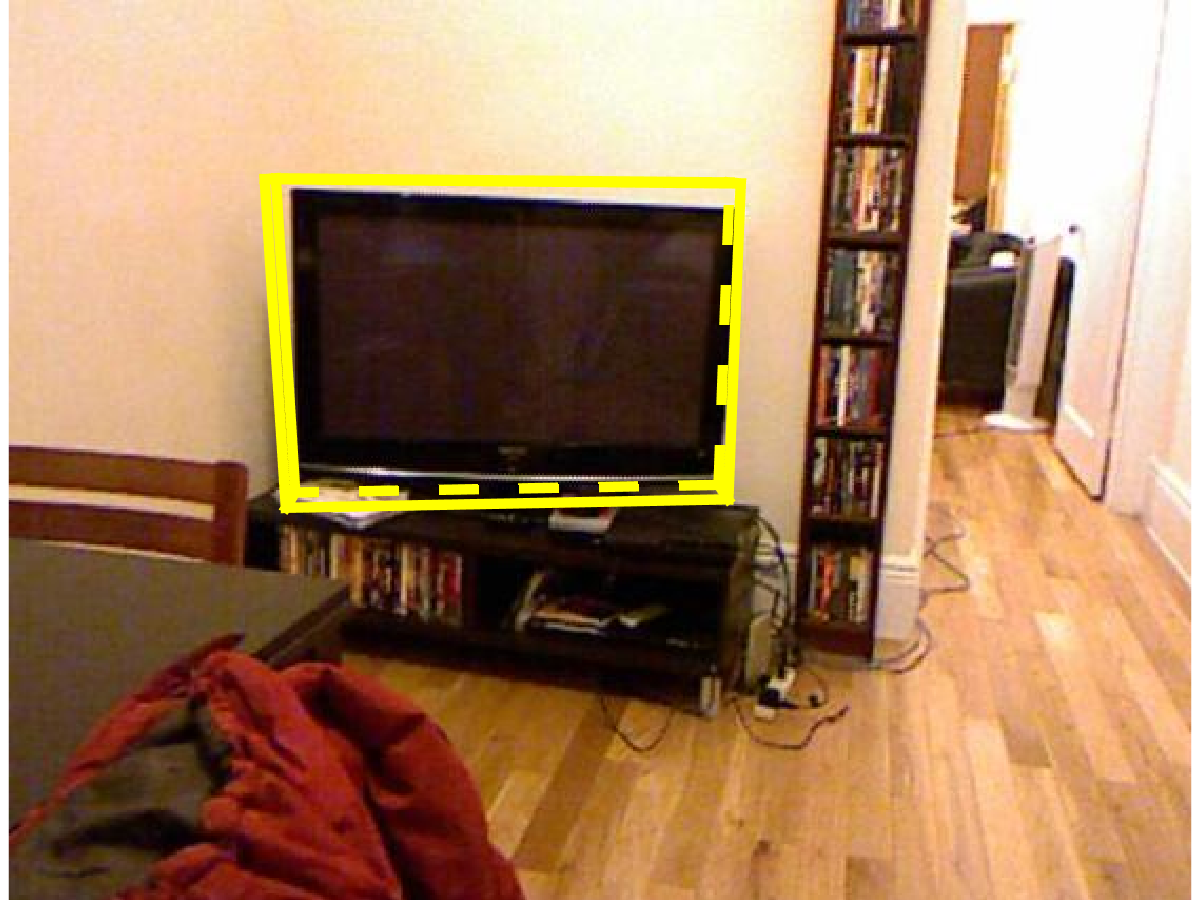}

\vspace{-1.2mm}

{\footnotesize
~~~~~bookshelf~~~~~~~~~~~~chair~~~~~~~~~~~~~~~dresser~~~~~~~~~~garbage bin~~~~~~~~~~~~sofa~~~~~~~~~~~~~~~~~~~box~~~~~~~~~~~~~~~~~~lamp~~~~~~~~~~~~~~~~~door~~~~~~~~~~~~~~~~~door~~~~~~~~~~~~~~~~~~tv}

\vspace{-0.3mm}

\caption{{\bf Misses.} Reasons: heavy occlusion, outside field of view, atypical size object, or missing depth.}
\label{fig:resultMiss}

\vspace{-3mm}
\end{figure*}

\section{Joint Amodal Object Recognition Network}
\label{sec:ObjectClassificationNetwork}
Given the 3D proposal boxes, we feed the 3D space within each box to the Object Recognition Network (ORN).
In this way, 
%although all receptive fields for the Region Proposal Network are cubes, because of the non-cubic anchors and box regression in Region Proposal Network,
the final proposal feed to ORN could be the actual bounding box for the object,
which allows the ORN to look at the full object to increase recognition performance,
while still being computationally efficient.
Furthermore,
because our proposals are amodal boxes containing the whole objects at their full extent,
the ORN can align objects in 3D meaningfully to be more invariant to occlusion or missing data for recognition.
Figure \ref{fig:amodalHist} shows statistics of object sizes between amodal full box \vs modal tight box.

\begin{figure}[t]
    \centering
    \includegraphics[width=\linewidth]{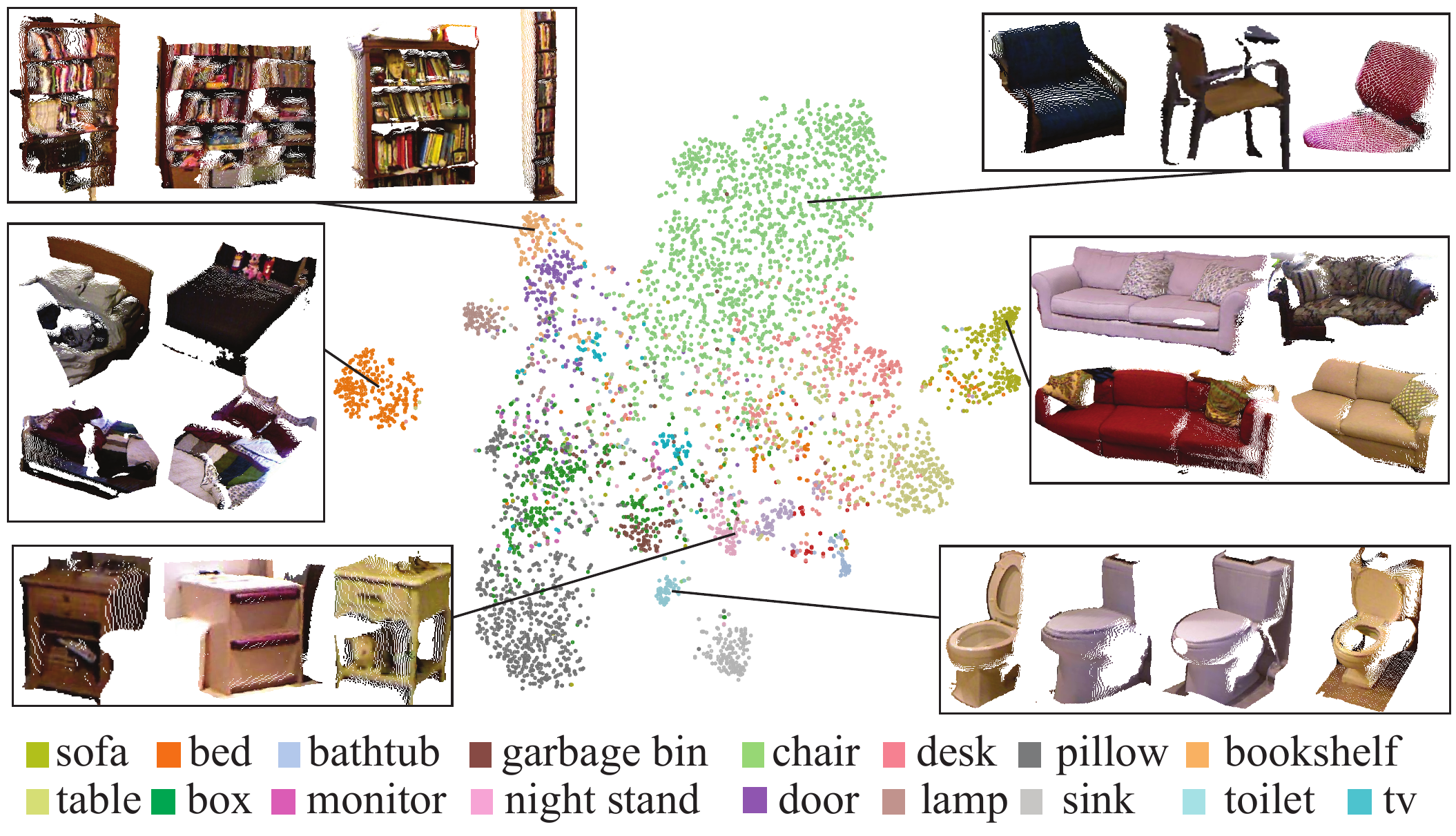}
    \caption{2D t-SNE embedding of the last layer features learned from the 3D ConvNet. Color encodes object category.}
    \label{fig:t-sne}
     \vspace{-3mm}
\end{figure}

\vspace{-4mm}\paragraph{3D object recognition network}
For each proposal box, 
we pad the proposal bounding box by $12.5\%$ of the sizes in each direction to encode some contextual information.
Then, we divide the space into a $30\times30\times30$ voxel grid and use TSDF (Section \ref{sec:Representation}) to encode the geometric shape of the object.
The network architecture is shown in Figure \ref{fig:ObjectRecogNet}.
All the max pooling layers are $2^3$ with stride $2$.
For the three convolution layers, 
the window sizes are $5^3$, $3^3$, and $3^3$, all with stride $1$.
Between the fully connected layers are ReLU and dropout layers (dropout ratio 0.5).  Figure \ref{fig:t-sne} visualizes the 2D t-SNE embedding of 5,000 foreground volumes using their the last layer features learned from the 3D ConvNet. Color encodes object category.

\vspace{-4mm}\paragraph{2D object recognition network}

The 3D network only makes use of the depth map, but not the color.
For certain object categories, color is a very discriminative feature,
and existing ConvNets provide very powerful features for image-based recognition that could be useful to us. % increase our recognition performance. 
%Therefore, we propose to use existing powerful color features for image recognition.
For each of the 3D proposal box,
we project the 3D points inside the proposal box to 2D image plane, and find the tightest 2D box that contains all these 2D point projections.
We use the state-of-the-art VGGnet \cite{vggnet} pre-trained on ImageNet \cite{imagenet} (without fine-tuning) to extract color features from the image.
We use a Region-of-Interest Pooling Layer from Fast RCNN \cite{FastRCNN} to uniformly sample $7\times7$ points from conv5\_3 layer 
using the 2D window with one more fully connected layer to generate 4096-dimensional features as the feature from 2D images.

We also tried the alternative to encode color on 3D voxels, but it performs much worse than the pre-trained VGGnet (Table \ref{fig:detection} [dxdydz+rgb] \vs [dxdydz+img]).
This might be because encoding color in 3D voxel grid significantly lowers the resolution compared to the original image, and hence high frequency signal in the image get lost. 
In addition, by using the pre-trained model of VGG, 
we are able to leverage the large amount of training data from ImageNet, and the well engineered network architecture.

\vspace{-4mm}\paragraph{2D and 3D joint recognition}
%To combine 2D and 3D, 
We construct a joint 2D and 3D network to make use of both color and depth.
The feature from both 2D VGG Net and our 3D ORN (each has 4096 dimensions) are concatenated into one feature vector, and fed into a fully connected layer 
%take this feature vector as input and
, which reduces the dimension to 1000.
Another two fully connected layer take this feature as input and predict the object label and 3D box.

\begin{table*}[t]
\vspace{-3mm}

\includegraphics[width=0.202\linewidth]{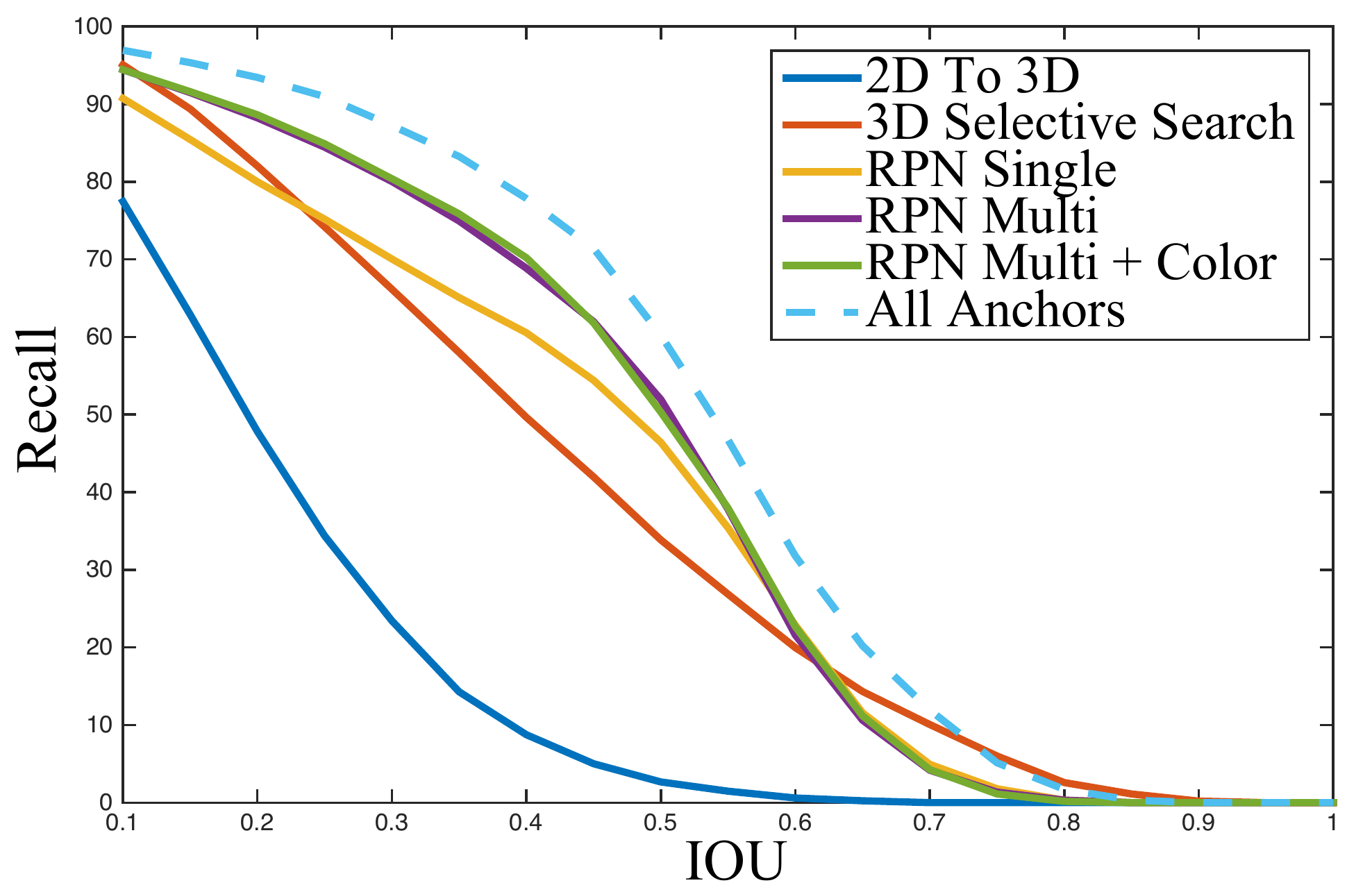}~%
\includegraphics[width=0.8\linewidth]{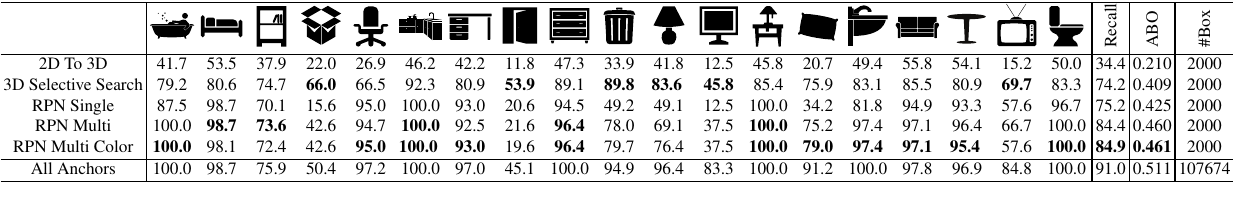}

\vspace{-2mm}
\caption{{\bf Evaluation for Amodal 3D Object Proposal.} [All Anchors] shows the performance upper bound when using all anchors.}
\label{fig:RPNeval}
\vspace{-3mm}
\end{table*}

\vspace{-4mm}\paragraph{Multi-task loss}
Similarly to RPN, 
the loss function consists of a classification loss and a 3D box regression loss: 
\vspace{-1mm}
\begin{equation}
\vspace{-1mm}
L(p,p^*,\mathbf{t},\mathbf{t}^*) = L_{\textrm{cls}}(p,p^*) + \lambda'[p^*>0]L_{\textrm{reg}}(\mathbf{t},\mathbf{t}^*),
\end{equation}
where the $p$ is the predicted probability over 20 object categories (negative non-objects is labeled as class 0).
%We consider the proposals that have 3D IOU with ground truth greater than 0.3 as positive examples, and those with IOU smaller than 0.1 as negative examples. 
%Different from RPN, for the ORN 
For each mini-batch, 
we sample 384 examples from different images, with a positive to negative ratio of 1:3.
For the box regression,
each target offset $\mathbf{t}^*$ is normalized element-wise with the object category specific mean and standard deviation. 

\vspace{-4mm}\paragraph{SVM and 3D NMS}
After training the network,
% jointly with all object categories, 
we extract the features from FC3 and train a linear Support Vector Machine (SVM) for each object category. 
During testing, we apply 3D NMS on the results with threshold 0.1, based on their SVM scores.
For box regressions, we directly use the results from the neural network.

\vspace{-4mm}\paragraph{Object size pruning}
As shown in Figure \ref{fig:amodalHist}, when we use amodal bounding boxes to represent objects, the bounding box sizes provide useful information about the object categories. 
To make use of this information, for each of the detected box, 
we check the box size in each direction, aspect ratio of each pair of box edge. We then compare these numbers with the distribution collected from training examples of the same category. 
If any of these values falls outside 1st to 99th percentile of the distribution, which indicates this box has a very different size,
we decrease its score by 2.

\vspace{-1mm}
\section{Experiments}
\vspace{-1mm}

%\section{Implementation}
The training of RPN and ORN takes around 10 and 17 hours respectively on a NVIDIA K40 GPU. 
During testing, RPN takes 5.62s and ORN takes 13.93s per image, % on average.
which is much faster than Depth RCNN (40s CPU + 30s GPU + expensive post alignment) and Sliding Shapes (25 mins $\times$ number of object categories).

For the VGG network \cite{vggnet}, we use the weights from \cite{gupta2015cross} without fine tuning.
To reduce GPU memory bandwidth, instead of storing as ``float'', we use ``half'' %with 2 bytes to represent a floating point number 
to store both the network parameters and activations. 
%We convert half to float for any computation and convert the result from float to half for storage.
We implement our own software framework from scratch using CUDA7.5 and CuDNN3 with FP16 support. 
The GPU memory usage is 5.41GB for RPN and 4.27GB for ORN (without VGGnet).

We evaluate our 3D region proposal and object detection algorithm on the standard NYUv2 dataset \cite{NYUdataset} and SUN RGB-D \cite{SUNRGBD} dataset. 
The amodal 3D bounding box are obtained from SUN RGB-D dataset. We modified the rotation matrix from SUN RGB-D dataset to 
eliminate the rotation on x,y plane and only contains camera tilt angle.
 Following the evaluation metric in \cite{SlidingShapes}, we assume the all predictions and ground truth boxes are aligned in the gravity direction. 
We use 3D volume intersection over union between ground truth and prediction boxes, and use 0.25 as the threshold to calculate the average recall for proposal generation and average precision for detection.

\subsection{Object Proposal Evaluation}
Evaluation of object proposal on NYU dataset is shown in Table \ref{fig:RPNeval}. On the left, we show the average recall over different IOUs.
On the right, we show the recall for each object category with IOU threshold 0.25, as well as the average best overlap ratio (ABO) across all ground truth boxes. 
Table \label{fig:resultRPN} shows the evaluation on SUNRGB-D dataset.
%We compare our method with the following baselines:

\vspace{-4mm}\paragraph{Na\"{\i}ve 2D To 3D}
%To see how well a na\"{\i}ve approach can work, our 
Our first baseline is to directly lift 2D object proposal to 3D.
We take the 2D object proposals from \cite{guptaCVPR15}.
For each of them, we get the 3D points inside the bounding box (without any background removal), remove those outside 2 percentiles along all three directions, and obtain a tight fitting box around these inlier points. 
%We use this as the most basic baseline. 
%This method is very sensitive to noise since we use the points in the bounding box without any figure/ground segmentation. 
%Besides, 
Obviously this method cannot predict amodal bounding box when the object is occluded or truncated, 
since 3D points only exist for the visible part of an object.

\def \mW {0.034}
\begin{table*}[t]
\vspace{-3mm}

\centering
\setlength{\tabcolsep}{1.7pt}
{
\centering
\footnotesize
\begin{tabular}{c|l|ccccccccccccccc|cccc|c}
\hline 
poposal & algorithm 
& \includegraphics[width=\mW\linewidth]{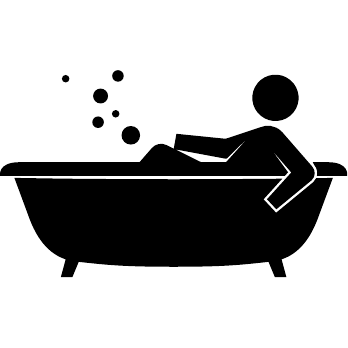} 
& \includegraphics[width=\mW\linewidth]{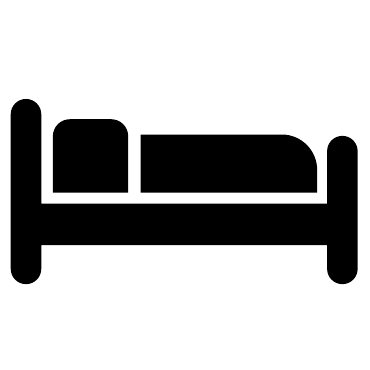} 
& \includegraphics[width=\mW\linewidth]{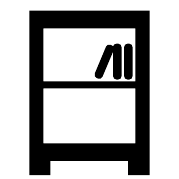} 
& \includegraphics[width=\mW\linewidth]{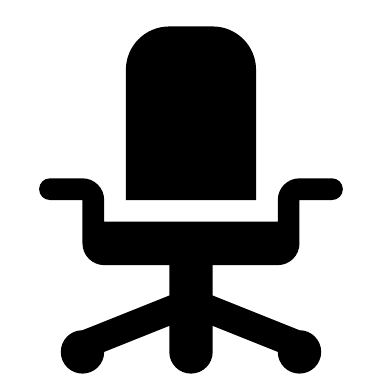}  
& \includegraphics[width=\mW\linewidth]{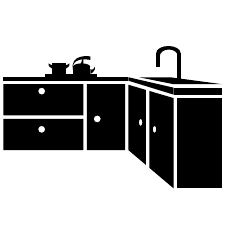}  
& \includegraphics[width=\mW\linewidth]{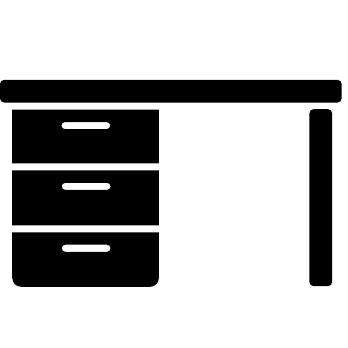}  
& \includegraphics[width=\mW\linewidth]{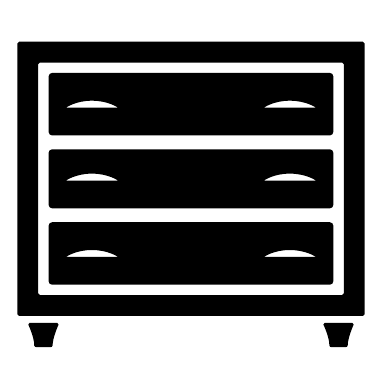}  
& \includegraphics[width=\mW\linewidth]{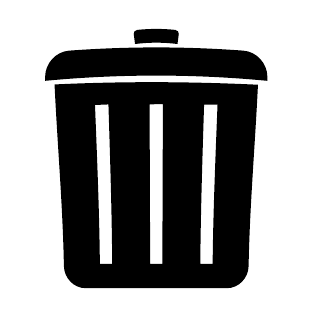}  
& \includegraphics[width=\mW\linewidth]{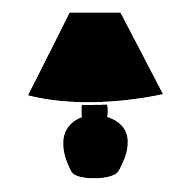}  
& \includegraphics[width=\mW\linewidth]{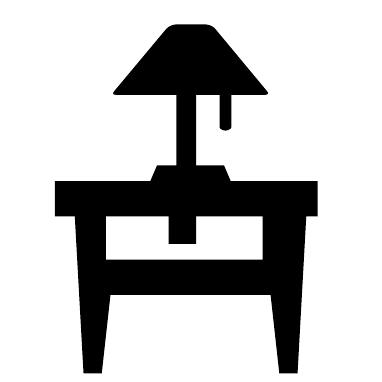}  
& \includegraphics[width=\mW\linewidth]{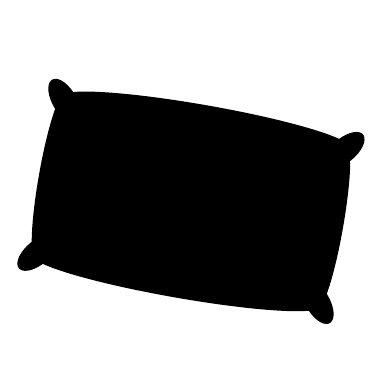}  
& \includegraphics[width=\mW\linewidth]{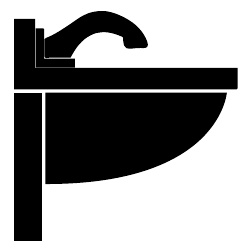}  
& \includegraphics[width=\mW\linewidth]{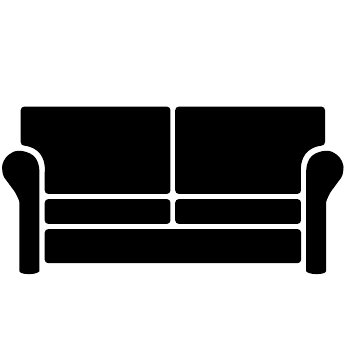}  
& \includegraphics[width=\mW\linewidth]{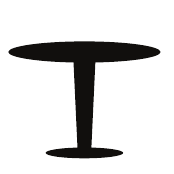}  
& \includegraphics[width=\mW\linewidth]{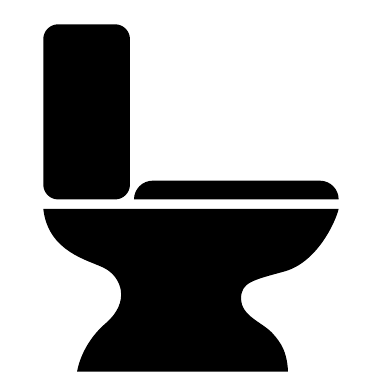}  
& \includegraphics[width=\mW\linewidth]{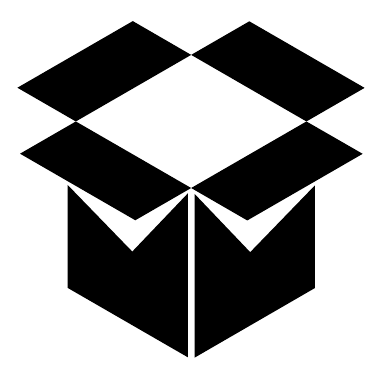} 
& \includegraphics[width=\mW\linewidth]{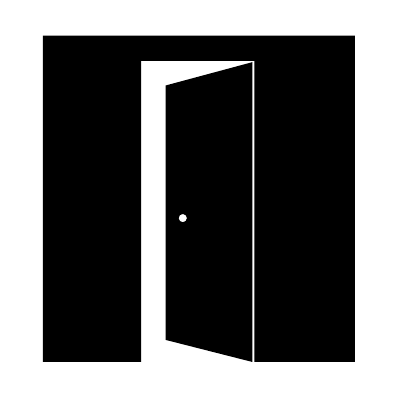}  
& \includegraphics[width=\mW\linewidth]{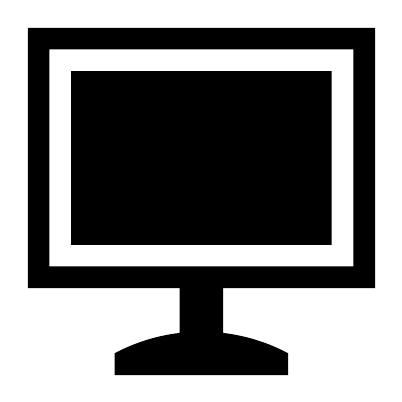}
& \includegraphics[width=\mW\linewidth]{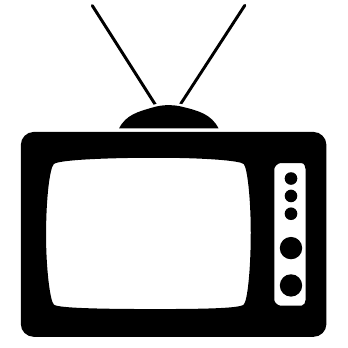} 
%& bathtub & bed & bookshelf & chair & counter & desk & dresser & garbage\_bin & lamp & night\_stand & pillow & sink & sofa & table & toilet & box & door & monitor & tv 
& mAP\tabularnewline

\hline 
\multirow{2}{*}{3D SS} & dxdydz no bbreg & 43.3 & 55.0 & 16.2 & 23.1 & 3.4 & 10.4 & 17.1 & 30.7 & 10.9 & 35.4 & 20.3 & 41.2 & 47.2 & 25.2 & 43.9 & 1.9 & 1.6 & 0.1 & 9.9 & 23.0\tabularnewline
 & dxdydz & 52.1 & 60.5 & 19.0 & 30.9 & 2.2 & 15.4 & 23.1 & 36.4 & 19.7 & 36.2 & 18.9 & 52.5 & 53.7 & 32.7 & 56.9 & 1.9 & 0.5 & 0.3 & 8.1 & 27.4\tabularnewline
\hline 
\multirow{10}{*}{RPN} & dxdydz no bbreg & 51.4 & 74.8 & 7.1 & 51.5 & 15.5 & 22.8 & 24.9 & 11.4 & 12.5 & 39.6 & 15.4 & 43.4 & 58.0 & 40.7 & 61.6 & 0.2 & 0.0 & 1.5 & 2.8 & 28.2\tabularnewline
 & dxdydz no svm & 58.9 & 79.8 & 15.7 & 56.3 & 11.3 & 20.3 & 18.8 & 16.5 & 18.2 & 38.1 & 15.1 & 54.0 & 57.7 & 47.2 & 66.8 & 1.3 & 0.0 & 0.6 & 8.3 & 30.8\tabularnewline
 & dxdydz no size & 59.9 & 78.9 & 12.0 & 51.5 & 15.6 & 24.6 & 27.7 & 12.5 & 18.6 & 42.3 & 15.1 & 59.4 & 59.6 & 44.7 & 62.5 & 0.3 & 0.0 & 1.1 & 12.9 & 31.5\tabularnewline
 & dxdydz & 59.0 & 80.7 & 12.0 & 59.3 & 15.7 & 25.5 & 28.6 & 12.6 & 18.6 & 42.5 & 15.3 & 59.5 & 59.9 & 45.3 & 64.8 & 0.3 & 0.0 & 1.4 & 13.0 & 32.3\tabularnewline
 & tsdf dis & 61.2 & 78.6 & 10.3 & 61.1 & 2.7 & 23.8 & 21.1 & 25.9 & 12.1 & 34.8 & 13.9 & 49.5 & 61.2 & 45.6 & 70.8 & 0.3 & 0.0 & 0.1 & 1.7 & 30.2\tabularnewline
 & dxdydz+rgb & 58.3 & 79.3 & 9.9 & 57.2 & 8.3 & 27.0 & 22.7 & 4.8 & 18.8 & 46.5 & 14.4 & 51.6 & 56.7 & 45.3 & 65.1 & 0.2 & 0.0 & 4.2 & 0.9 & 30.1\tabularnewline
 %& img+hha & 13.8 & 26.0 & 10.0 & 10.2 & 2.1 & 6.4 & 12.8 & 7.7 & 8.2 & 10.3 & 11.4 & 11.4 & 22.3 & 11.4 & 19.8 & 0.2 & 0.0 & 0.2 & 2.6 & 9.8\tabularnewline
 & proj dxdydz+img & 58.4 & 81.4 & 20.6 & 53.4 & 1.3 & 32.2 & 36.5 & 18.3 & 17.5 & 40.8 & 19.2 & 51.0 & 58.7 & 47.9 & 71.4 & 0.5 & 0.2 & 0.3 & 1.8 & 32.2\tabularnewline
 & dxdydz+img+hha & 55.9 & 83.0 & 18.8 & 63.0 & 17.0 & 33.4 & 43.0 & 33.8 & 16.5 & 54.7 & 22.6 & 53.5 & 58.0 & 49.7 & 75.0 & 2.6 & 0.0 & 1.6 & 6.2 & 36.2\tabularnewline
 & dxdydz+img & 62.8 & 82.5 & 20.1 & 60.1 & 11.9 & 29.2 & 38.6 & 31.4 & 23.7 & 49.6 & 21.9 & 58.5 & 60.3 & 49.7 & 76.1 & 4.2 & 0.0 & 0.5 & 9.7 & 36.4\tabularnewline
\hline 

\end{tabular}
}

\vspace{-1mm}
\caption{{\bf Control Experiments on NYUv2 Test Set.}
Not working: box (too much variance), door (planar), monitor and tv (no depth).
%We evaluate on 19 object categories using Average Precision (AP)
%bathtub, bed, bookshelf, box, chair, counter, desk, door, dresser, garbage bin, lamp, monitor, night stand, pillow, sink, sofa, table, tv and toilet.
}
\label{fig:detection}
\vspace{-3mm}
\end{table*}

\vspace{-4mm}\paragraph{3D Selective Search}

For 2D, Selective Search \cite{SelectiveSearch} is one of the most popular state-of-the-arts.
It starts with a 2D segmentation and uses hierarchical grouping to obtain the object proposals at different scales. 
We study how well a similar method based on bottom-up grouping can work in 3D (3D SS).
We first use plane fitting on the 3D point cloud to get an initial segmentation.
For each big plane that covers more than $10\%$ of the total image area,
we use the RGB-D UCM segmentation from \cite{depthRCNN} (with threshold 0.2) to further split it. 
Starting with on this over-segmentation, we hierarchically group \cite{SelectiveSearch} different segmentation regions,
with the following similarity measures: \\
{
\small
{$\cdot$ \small{$s_{\textrm{color}}(r_i ,r_j) $}} measures color similarity between region $r_t$ and $r_j$ using histogram intersection on RGB color histograms;\\
{$\cdot$ \small{$s_{\textrm{\#pixels}}(r_i ,r_j) = 1- \frac{\textrm{\#pixels}(r_i)+\textrm{\#pixels}(r_j)}{\textrm{\#pixels}(im)}$}}, where $\textrm{\#pixels}(\cdot)$ is number of pixels in this region;\\
%This term and the next encourage small regions to merge early;\\
{$\cdot$ \small{$s_{\textrm{volume}}(r_i ,r_j) = 1- \frac{\textrm{volume}(r_i)+\textrm{volume}(r_j)}{\textrm{volume}(room)}$}}, where $\textrm{volume}(\cdot) $ is the volume of 3D bounding boxes of the points in this region;\\
{$\cdot$ \small{$s_{\textrm{fill}}(r_i ,r_j) = 1- \frac{\textrm{volume}(r_i)+\textrm{volume}(r_j)}{\textrm{volume}(ri\cup rj)}$}} measures how well region $r_i$ and $r_j$ fit into each other to fill in gaps.\\
}
The final similarity measure is a weighted sum of these four terms.
%\begin{equation}
%s(r_i ,r_j) = w_1s_{\textrm{color}}+ w_2s_{\textrm{\#pixels}}+ w_3s_{\textrm{volume}}+ w_4s_{\textrm{fill}}
%\end{equation}
To diversify our strategies, we run the grouping 5 times with different weights: $[1,0,0,0],$ $[0,1,0,0],$ $[0,0,1,0],$ $[0,0,0,1],$ $[1,1,1,1]$.
For each of the grouped region, we will obtain two proposal boxes: one tight box and one box with height extended to the floor. 
%For all the proposal boxes, 
We also use the room orientation as the box orientation. 
The room orientation and floor are obtained under the Manhattan world assumption described in Section \ref{sec:RegionProposalNetwork}.
After that we will remove the redundant proposals with 3D IOU greater than 0.9 by arbitrary selection.
Using both 3D and color, this very strong baseline achieves an average recall 74.2\%. 
But it is slow because of its many steps, 
and the handcrafted segmentation and similarity might be difficult to tune optimally.
% Therefore, we propose to use a region proposal network to automatically learn the 3D objectness from end to end.

\vspace{-4mm}\paragraph{Our 3D RPN}

Row 3 to 5 in Table \ref{fig:RPNeval} shows the performance of our 3D region proposal network. 
Row 3 shows the performance of single-scale RPN. 
Note that the recalls for small objects like lamp, pillow, garbage bin are very low. 
When one more scale is added, the performance for those small objects boosts significantly.
Adding RGB color to the 3D TSDF encoding slightly improves the performance,
and we use this as our final region proposal result.
From the comparisons we can see that mostly planar objects (\eg door) are easier to locate using segmentation-based selective search.
Some categories (\eg lamp) have a lower recall mostly because of lack of training examples.
Table \ref{fig:detection} shows the detection AP when using the same ORN architecture but different proposals (Row [3D SS: dxdydz] and Row [RPN: dxdydz]). 
We can see that the proposals provided by RPN %indeed have higher quality, and 
helps to improve the detection performance by a large margin (mAP from 27.4 to 32.3). 
%As point out by many previous work \cite{FastRCNN, FasterRCNN} that the Recall-to-IOU

%\vspace{3mm}

%\begin{table}[t]
%\centering
%\setlength{\tabcolsep}{4pt}
%\footnotesize
%\begin{tabular}{c|c|c|c|c}
%\hline 
%& \includegraphics[width=0.08\linewidth]{figures/class/chair.pdf} 
%& \includegraphics[width=0.08\linewidth]{figures/class/bed2.pdf} 
%& \includegraphics[width=0.08\linewidth]{figures/class/toilet.pdf} 
%& \includegraphics[width=0.08\linewidth]{figures/class/sofa.pdf}  
%\tabularnewline
%% & floor & ceiling & chair & table & bed & nightstand & books & person & mean\tabularnewline
%\hline 
%Sliding Shapes \cite{SlidingShapes} & 33.42 & 25.78 & 42.09 & 61.86 \tabularnewline
%\hline 
%DSS (ours)  & & & & \tabularnewline
%\hline 
%\end{tabular}
%
%\vspace{1mm}
%\caption{{\bf Evaluation for 3D Object Orientation.}}
%\label{fig:3dorientation}
%\vspace{-3mm}
%\end{table}

\subsection{Object Detection Evaluation}

%Figure \ref{fig:resultTP}, \ref{fig:resultFP}, and \ref{fig:resultMiss}
%show some successful and failure cases. 
%Table \ref{fig:detection} and \ref{fig:3ddetection} include evaluations and comparisons.
%\vspace{-4mm}
\paragraph{Feature encoding}

From Row [RPN: dxdydz] to Row [RPN: dxdydz+img] in Table \ref{fig:detection}, we compare different feature encodings and reach the following conclusions.
%The conclusion is as follow: 
(1) TSDF with directions encoded is better than single TSDF distance ([dxdydz] \vs [tsdf dis]).
(2) Accurate TSDF is better than projective TSDF  ([dxdydz+img] \vs [proj dxdydz+img]). 
(3) Directly encoding color on 3D voxels is not as good as using 2D image VGGnet ([dxdydz+rgb] \vs [dxdydz+img]),
probably because the latter one can preserve high frequency signal from images.
(4) Adding HHA does not help, which indicates the depth information from HHA is already exploited by our 3D representation ([dxdydz+img+hha] \vs [dxdydz+img]).

\begin{table}[t]
\vspace{-1mm}
\centering
\setlength{\tabcolsep}{3pt}
\footnotesize
\begin{tabular}{l|c|c|c|c|c|c|c}
\hline 
Algorithm &  input
& \includegraphics[width=0.07\linewidth]{figures/class/bed2.pdf} 
& \includegraphics[width=0.07\linewidth]{figures/class/chair.pdf} 
& \includegraphics[width=0.07\linewidth]{figures/class/table.pdf} 
& \includegraphics[width=0.07\linewidth]{figures/class/sofa.pdf} 
& \includegraphics[width=0.07\linewidth]{figures/class/toilet.pdf} 

& mAP\tabularnewline
\hline 
Sliding Shapes \cite{SlidingShapes}  & d & 33.5 & 29 & 34.5 & 33.8 & 67.3 & 39.6\tabularnewline
\hline 
\cite{guptaCVPR15} on instance seg& d & 71 & 18.2 & 49.6 & 30.4 & 63.4 & 46.5\tabularnewline
\cite{guptaCVPR15} on instance seg & rgbd & 74.7 & 18.6 & 50.3 & 28.6 & 69.7 & 48.4\tabularnewline
\cite{guptaCVPR15} on estimated model & d & 72.7 & 47.5 & 54.6 & 40.6 & 72.7 & 57.6\tabularnewline
\cite{guptaCVPR15} on estimated model & rgbd & 73.4 & 44.2 & 57.2 & 33.4 & 84.5 & 58.5\tabularnewline
\hline 
ours  {[}depth only{]} & d & 83.0 & 58.8 & 68.6 & 49.5 & 79.2 & 67.8\tabularnewline
%ours {[}depth +img +hha{]} & rgbd & 82.7 & 51.9 & 69.7 & 49.1 & 83.1 & 67.3\tabularnewline
ours {[}depth + img{]} & rgbd & {\bf 84.7} & {\bf  61.1} & {\bf 70.5} & {\bf 55.4} & {\bf 89.9} & {\bf 72.3}\tabularnewline
\hline 
\end{tabular}
\vspace{-1mm}
\caption{{\bf Comparison on 3D Object Detection.}}
\label{fig:3ddetection}
\vspace{-5mm}
\end{table}

\begin{table*}[t]
\vspace{-3mm}
\setlength{\tabcolsep}{1.7pt}
{
\centering
\footnotesize
\begin{tabular}{c|ccccccccccccccccccc|c|c|c}
\hline
& \includegraphics[width=0.037\linewidth]{figures/class/bathtub2.pdf} 
& \includegraphics[width=0.037\linewidth]{figures/class/bed2.pdf} 
& \includegraphics[width=0.037\linewidth]{figures/class/bookshelf.pdf} 
& \includegraphics[width=0.037\linewidth]{figures/class/box2.pdf} 
& \includegraphics[width=0.037\linewidth]{figures/class/chair.pdf}  
& \includegraphics[width=0.037\linewidth]{figures/class/counter.pdf}  
& \includegraphics[width=0.037\linewidth]{figures/class/desk.pdf}  
& \includegraphics[width=0.037\linewidth]{figures/class/door.pdf}  
& \includegraphics[width=0.037\linewidth]{figures/class/dresser.pdf}  
& \includegraphics[width=0.037\linewidth]{figures/class/garbage_bin.pdf}  
& \includegraphics[width=0.037\linewidth]{figures/class/lamp.pdf}  
& \includegraphics[width=0.037\linewidth]{figures/class/monitor.pdf}  
& \includegraphics[width=0.037\linewidth]{figures/class/night_stand.pdf}  
& \includegraphics[width=0.037\linewidth]{figures/class/pillow.pdf}  
& \includegraphics[width=0.037\linewidth]{figures/class/sink2.pdf}  
& \includegraphics[width=0.037\linewidth]{figures/class/sofa.pdf}  
& \includegraphics[width=0.037\linewidth]{figures/class/table.pdf}  
& \includegraphics[width=0.037\linewidth]{figures/class/tv.pdf}  
& \includegraphics[width=0.037\linewidth]{figures/class/toilet.pdf}  
& \rotatebox[origin=lB]{90}{Recall} & \rotatebox[origin=lB]{90}{ABO} & \rotatebox[origin=lB]{90}{\#Box} 
\tabularnewline
\hline 
3D SS & 78.8 & 87.2 & 72.8 & 72.2 & 65.5 & 86.1 & 75.1 & 65.0 & 70.0 & 87.1 & 67.5 & 53.1 & 68.1 & 82.8 & 86.8 & 84.4 & 85.0 & 69.2 & 94.0 & 72.0 & 0.394 & 2000
\tabularnewline
\hline 
RPN & 98.1 & 99.1 & 79.5 & 51.5 & 93.3 & 89.2 & 94.9 & 24.0 & 87.0 & 79.6 & 62.0 & 41.2 & 96.2 & 77.9 & 96.7 & 97.3 & 96.7 & 63.3 & 100.0 & 88.7 & 0.485 & 2000
\tabularnewline
\hline 
\end{tabular}
}
\vspace{-1mm}
\caption{{\bf  Evaluation for regoin proposal generation on SUN RGB-D test set.}}
\label{fig:resultRPN}
\end{table*}

\def \mW {0.032}
\begin{table*}[t]
{
\setlength{\tabcolsep}{2.1pt}
\centering
\footnotesize
\begin{tabular}{l|ccccccccccccccccccccc|c}
\hline 
& \includegraphics[width=\mW\linewidth]{figures/class/bathtub2.pdf} 
& \includegraphics[width=\mW\linewidth]{figures/class/bed2.pdf} 
& \includegraphics[width=\mW\linewidth]{figures/class/bookshelf.pdf} 
& \includegraphics[width=\mW\linewidth]{figures/class/box2.pdf} 
& \includegraphics[width=\mW\linewidth]{figures/class/chair.pdf}  
& \includegraphics[width=\mW\linewidth]{figures/class/counter.pdf}  
& \includegraphics[width=\mW\linewidth]{figures/class/desk.pdf}  
& \includegraphics[width=\mW\linewidth]{figures/class/door.pdf}  
& \includegraphics[width=\mW\linewidth]{figures/class/dresser.pdf}  
& \includegraphics[width=\mW\linewidth]{figures/class/garbage_bin.pdf}  
& \includegraphics[width=\mW\linewidth]{figures/class/lamp.pdf}  
& \includegraphics[width=\mW\linewidth]{figures/class/monitor.pdf}
& \includegraphics[width=\mW\linewidth]{figures/class/night_stand.pdf}  
& \includegraphics[width=\mW\linewidth]{figures/class/pillow.pdf}  
& \includegraphics[width=\mW\linewidth]{figures/class/sink2.pdf}  
& \includegraphics[width=\mW\linewidth]{figures/class/sofa.pdf}  
& \includegraphics[width=\mW\linewidth]{figures/class/table.pdf}  
& \includegraphics[width=\mW\linewidth]{figures/class/tv.pdf} 
& \includegraphics[width=\mW\linewidth]{figures/class/toilet.pdf}  

& mAP\tabularnewline
\hline  
Sliding Shapes \cite{SlidingShapes} & - & 42.09 & - &  & 33.42 & - & -&-  &- &-  & - &-  & - & - & - & 23.28  &25.78 & - & 61.86 &- \tabularnewline
%\hline  
%COG wo. Context \cite{Zhile2016COG} &47.6&52.9&12.8&-&45.1&-&28.1&-&7.9&-&-&-&14.2&-&-&-&28.6&-&43.0&-\tabularnewline
%\hline  
%COG w. Context \cite{Zhile2016COG}  &58.3&63.6&31.8&-&62.2&-&45.2&-&15.47&-&-&-&27.3&-&-&-&51.2&-&70.0&-\tabularnewline
\hline  
Deep Sliding Shapes &44.2&78.8&11.9&1.5&61.2&4.1&20.5&0.0&6.4&20.4&18.4&0.2&15.4&13.3&32.3&53.5&50.3&0.5&78.9&26.9\tabularnewline
\hline 
\end{tabular}
}
\vspace{1mm}
\caption{{\bf Evaluation for 3D amodal object detection on SUN RGB-D test set.}}
\label{fig:result}
\end{table*}

\vspace{-4mm}\paragraph{Design justification}
We conducted several control experiments to understand the importance of each component. % to justify our design decisions. 

{\bf Does bounding box regression help?}
Previous works have shown that box regression can significantly improve 2D object detection \cite{FastRCNN}. 
For our task, although we have depth,
%But at the same time, 
there is more freedom on 3D localization, which makes regression harder.
%To see the benefits of 3D bounding box regression, 
We turn the 3D box regression on ([3DSS dxdydz], [RPN dxdydz])
and off ([3DSS dxdydz no bbreg], [RPN dxdydz no bbreg]).
%We compare without bounding box regression 
%and with bounding box regression ([3DSS dxdydz], [RPN dxdydz]).
Whether we use 3D Selective Search or RPN for proposal generation,
the 3D box regression always helps significantly (mAPs improve +4.4 and +4.1 respectively).
%, indicates the 3D bounding box regression is very critical for better object detection performance. 

{\bf Does SVM outperform softmax?}
Instead of training one-vs-rest linear SVMs, we can also directly use the softmax score as detection confidence. Experiments in \cite{FastRCNN} shows that softmax slightly outperforms SVM. 
But in our case, using SVM improves the mAP by 0.5.
This may be because SVM can better handle the unbalanced numbers of training examples among different categories in NYU.

{\bf Does size pruning help?}
Compared with and without the post-processing ([dxdydz] \vs [dxdydz no size]),
we observe that for most categories, size pruning reduces false positives and improves the AP by the amount from 0.1 to 7.8, showing a consistent positive effect.

\begin{figure}[t]

\vspace{-2mm}
{\footnotesize
~~~~~~~~~~Depth~~~~~~~~~~~~~~~~~~~~~~~~~~Sliding Shapes \cite{SlidingShapes} ~~~~~~~~~~~~~~~~~~~~~~Ours~~

}

\vspace{-4mm}
\begin{center}
\includegraphics[width=0.227\linewidth]{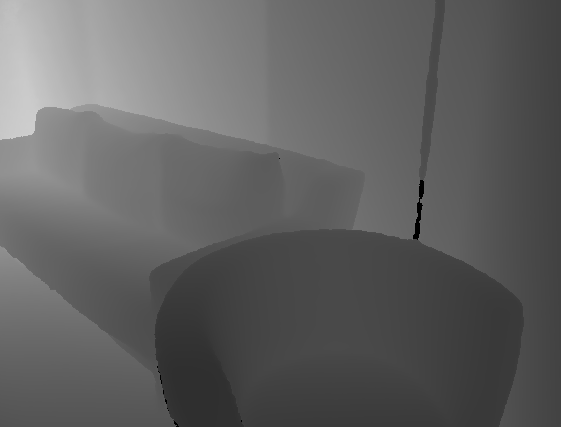}~~~%
\includegraphics[width=0.227\linewidth]{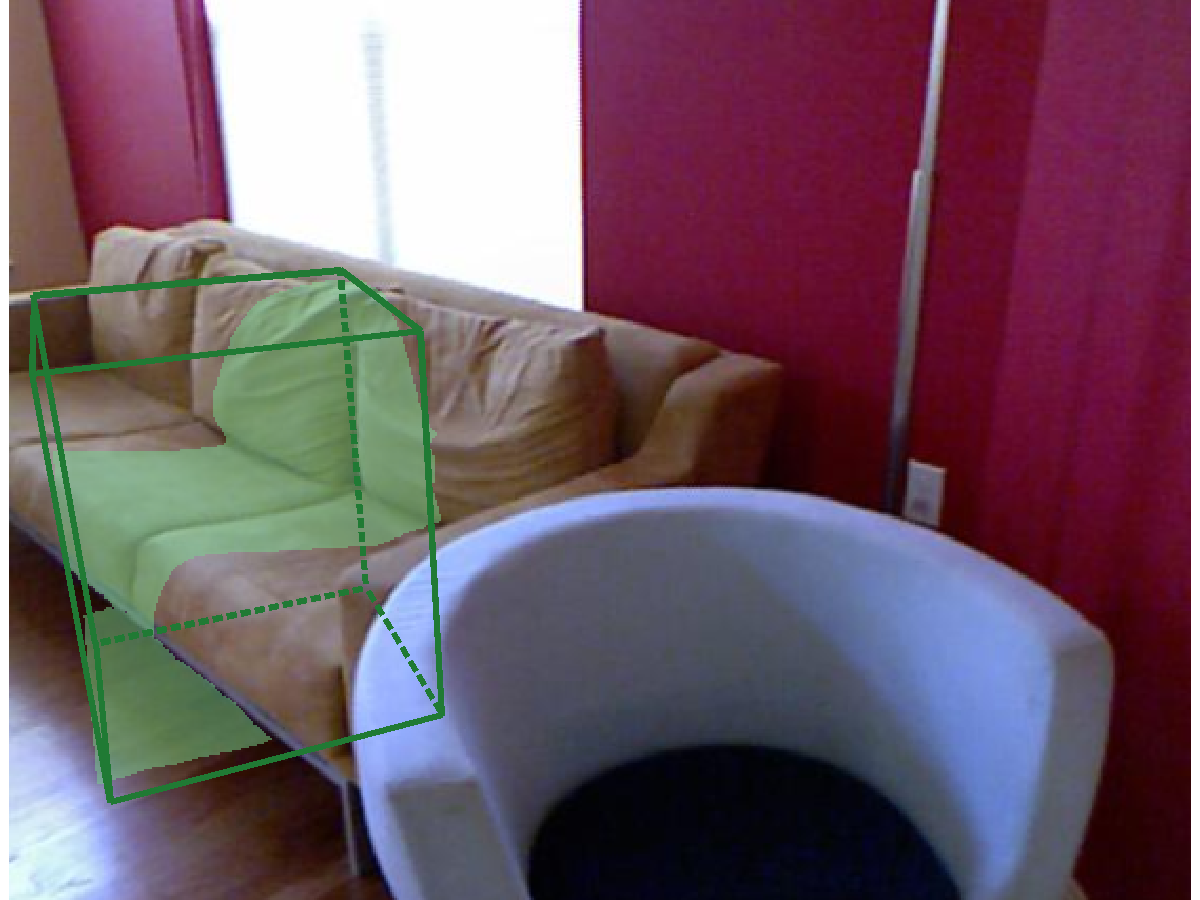}%
\includegraphics[width=0.227\linewidth]{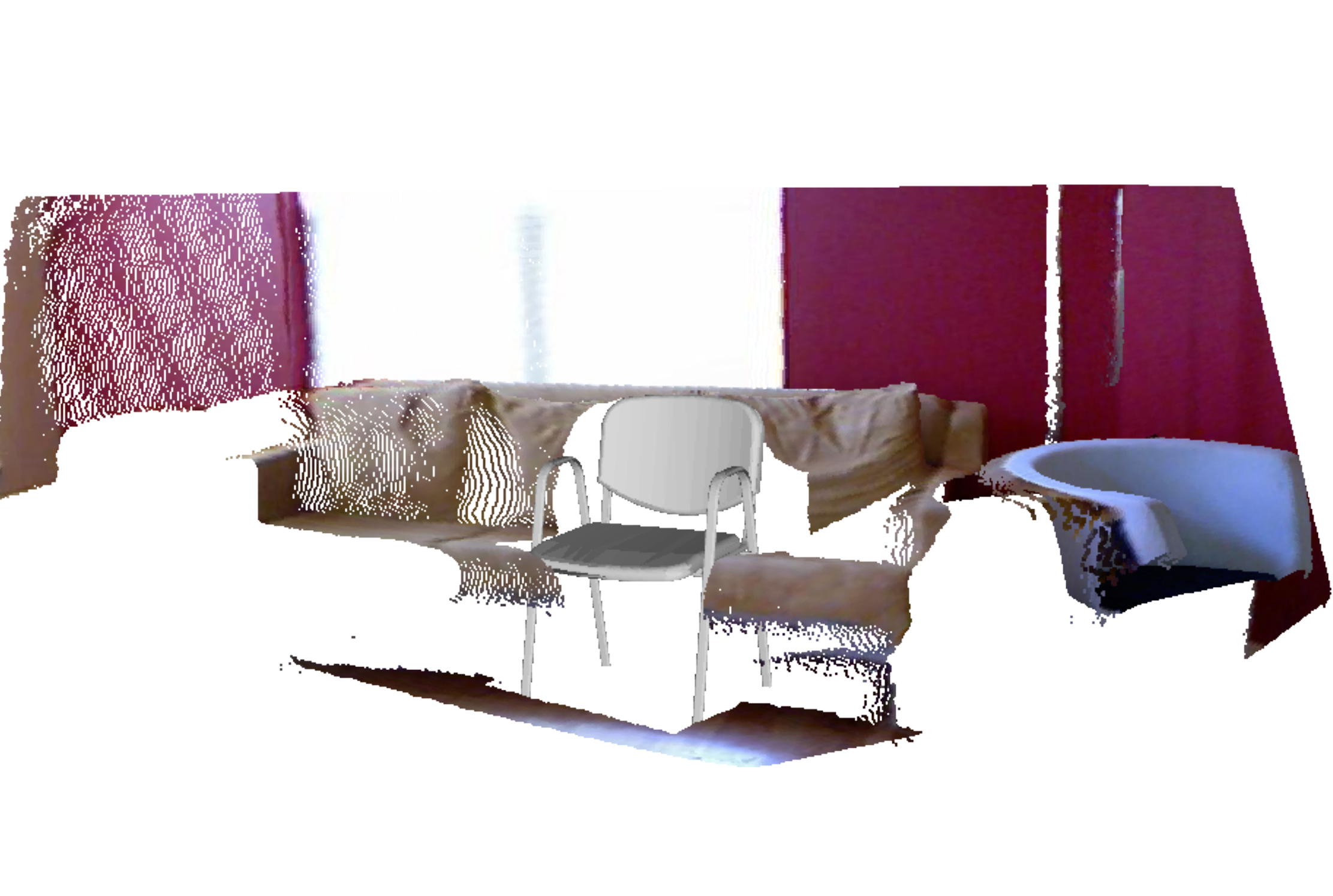}~~~~%
\includegraphics[width=0.227\linewidth]{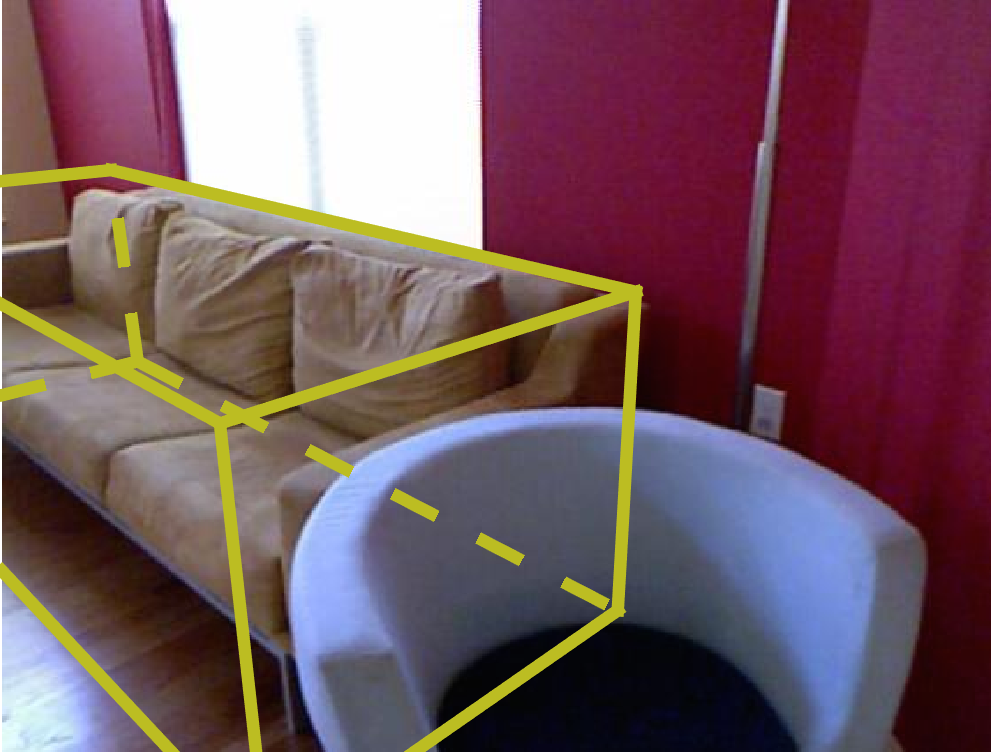}

\includegraphics[width=0.227\linewidth]{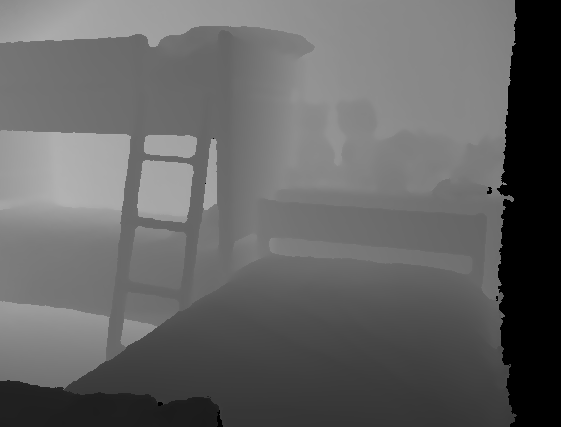}~~~%
\includegraphics[width=0.227\linewidth]{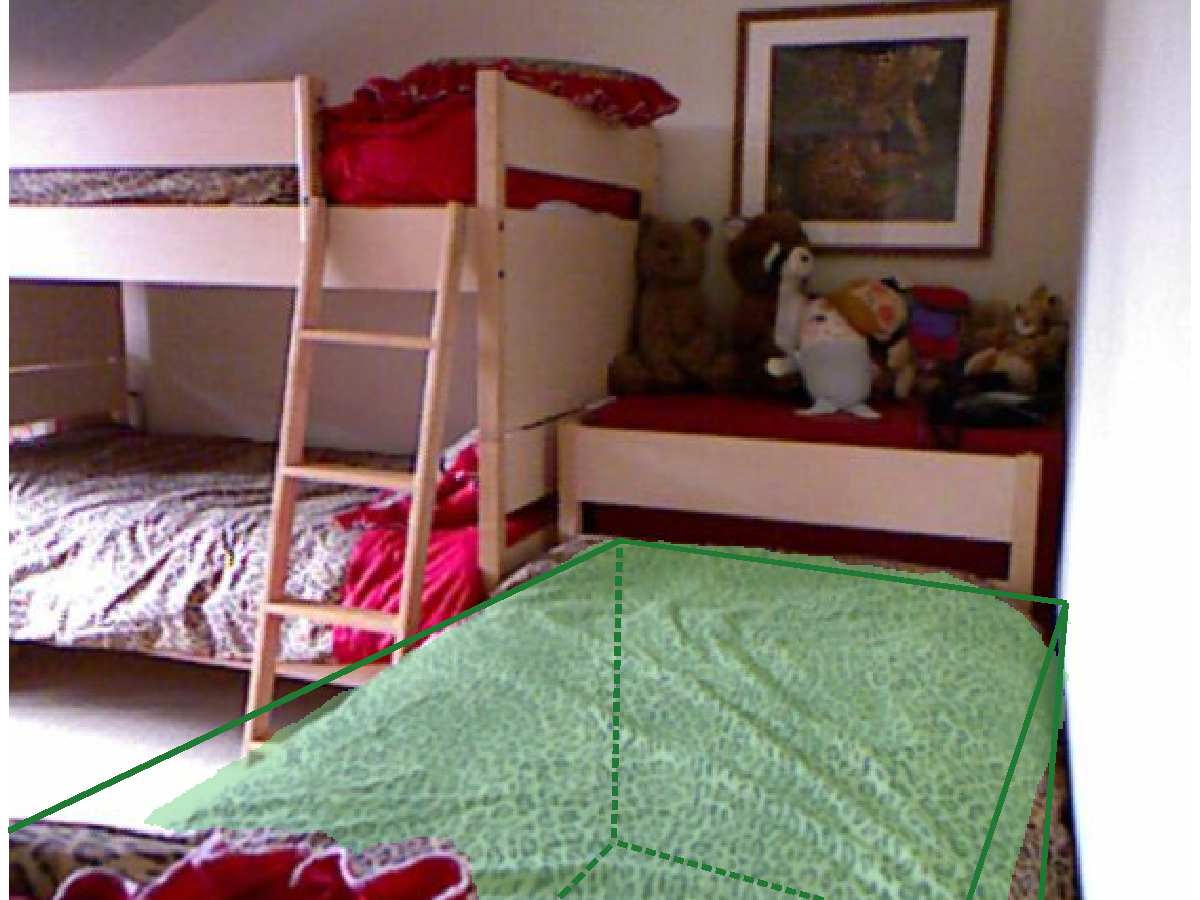}%
\includegraphics[width=0.227\linewidth]{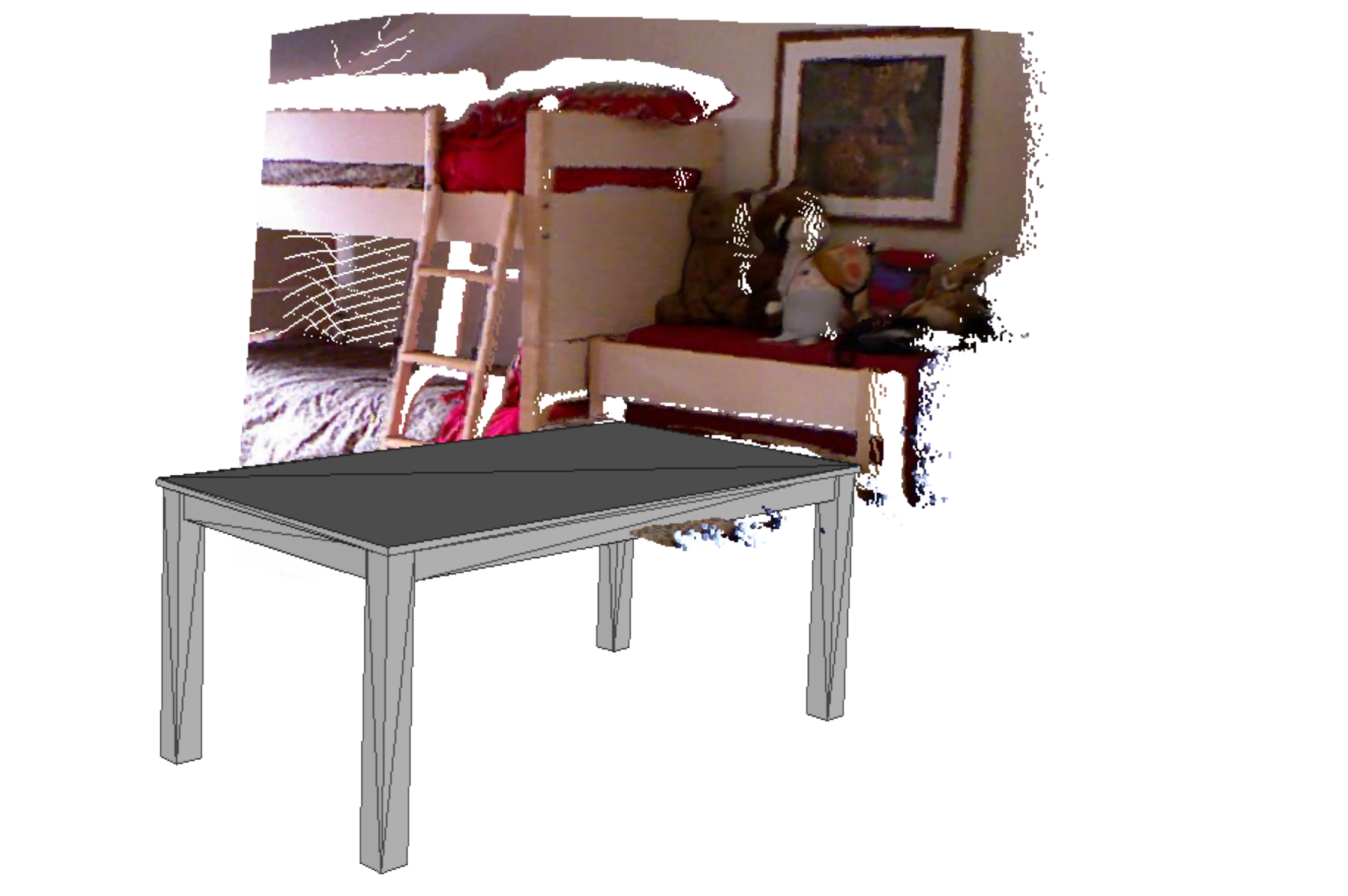}~~~~%
\includegraphics[width=0.227\linewidth]{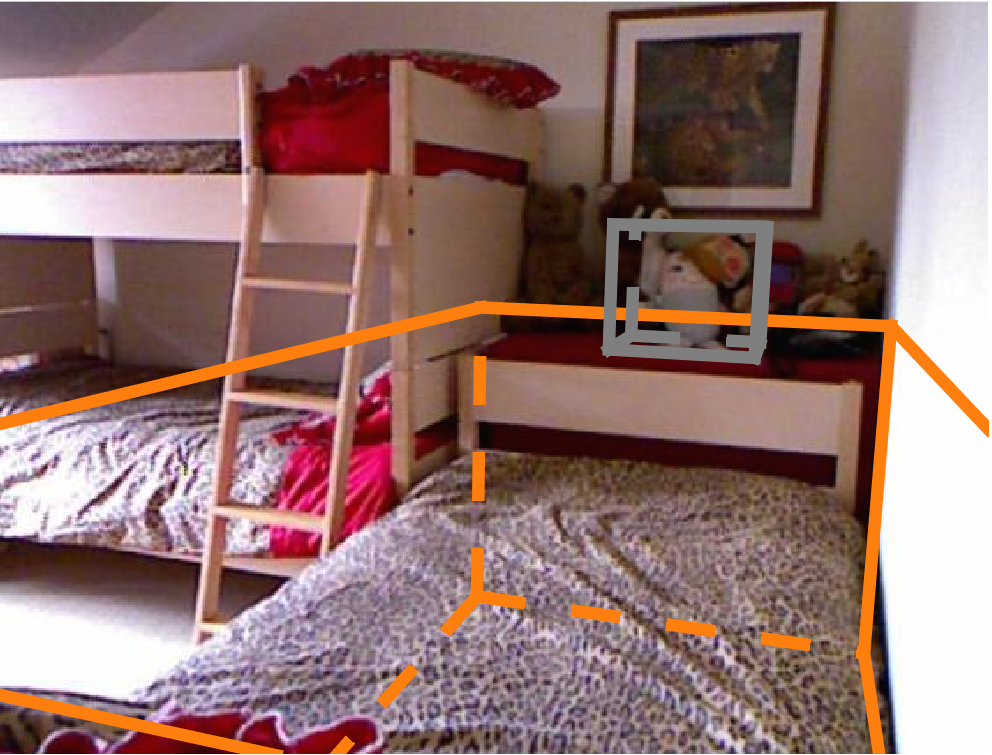}

\includegraphics[width=0.227\linewidth]{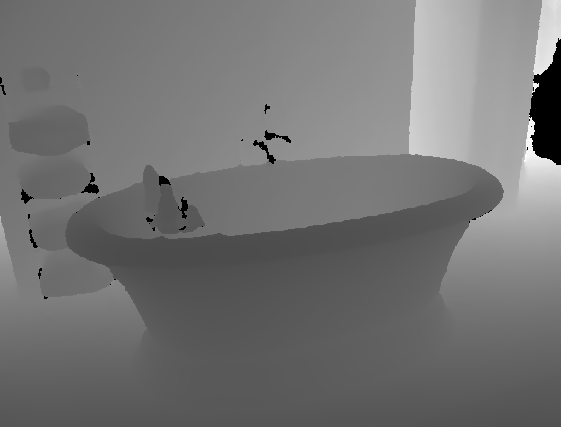}~~~%
\includegraphics[width=0.227\linewidth]{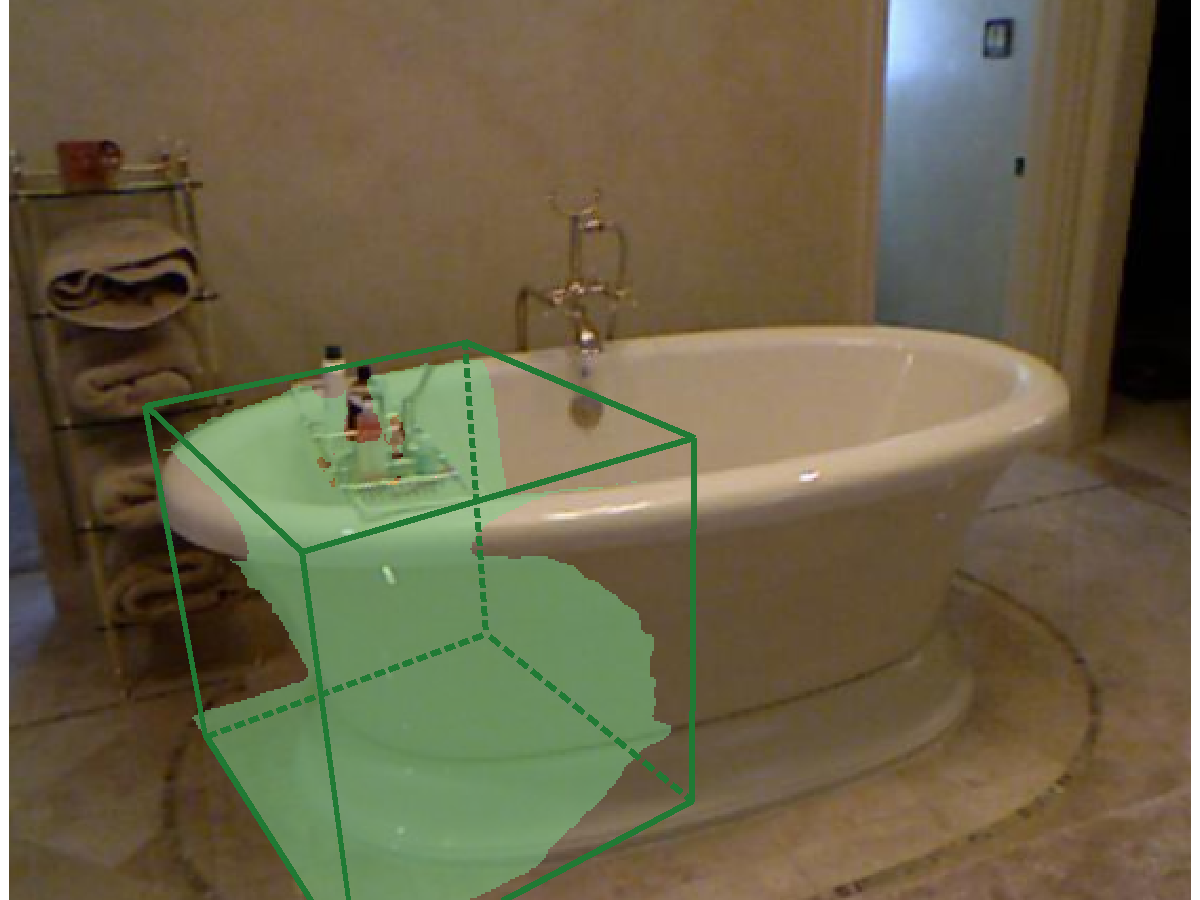}%
\includegraphics[width=0.227\linewidth]{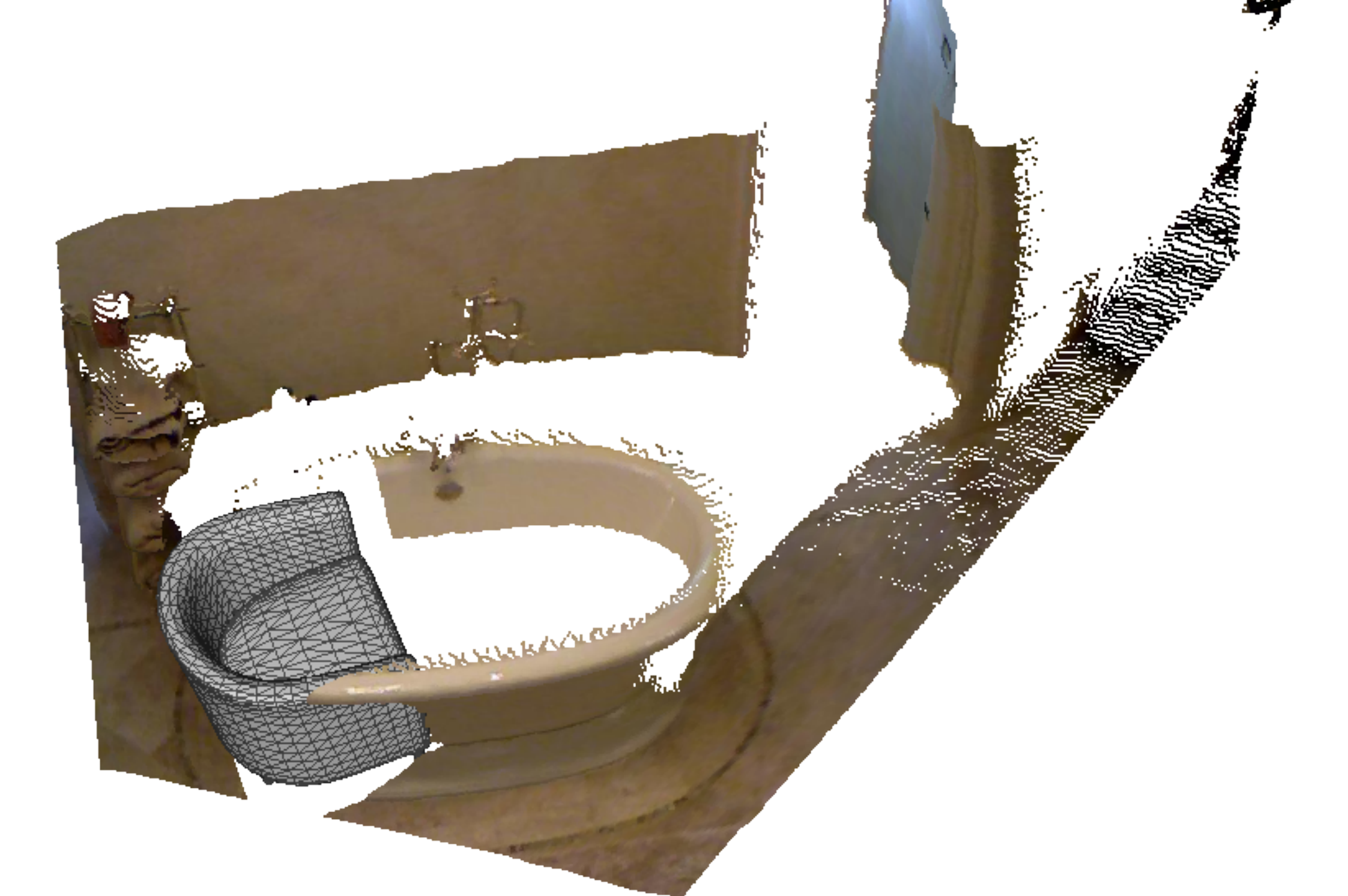}~~~~%
\includegraphics[width=0.227\linewidth]{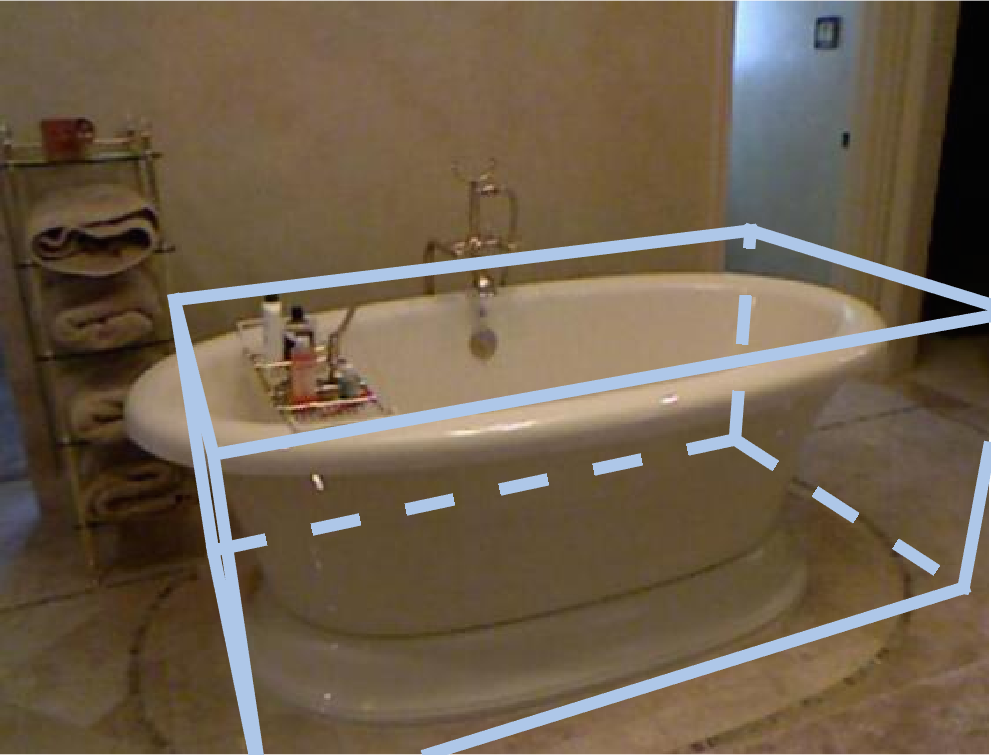}

\includegraphics[width=0.227\linewidth]{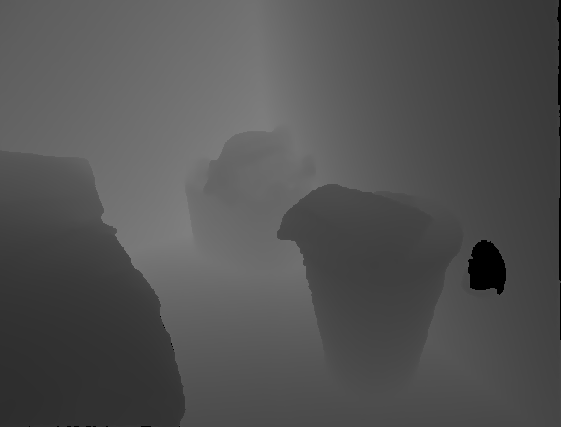}~~~%
\includegraphics[width=0.227\linewidth]{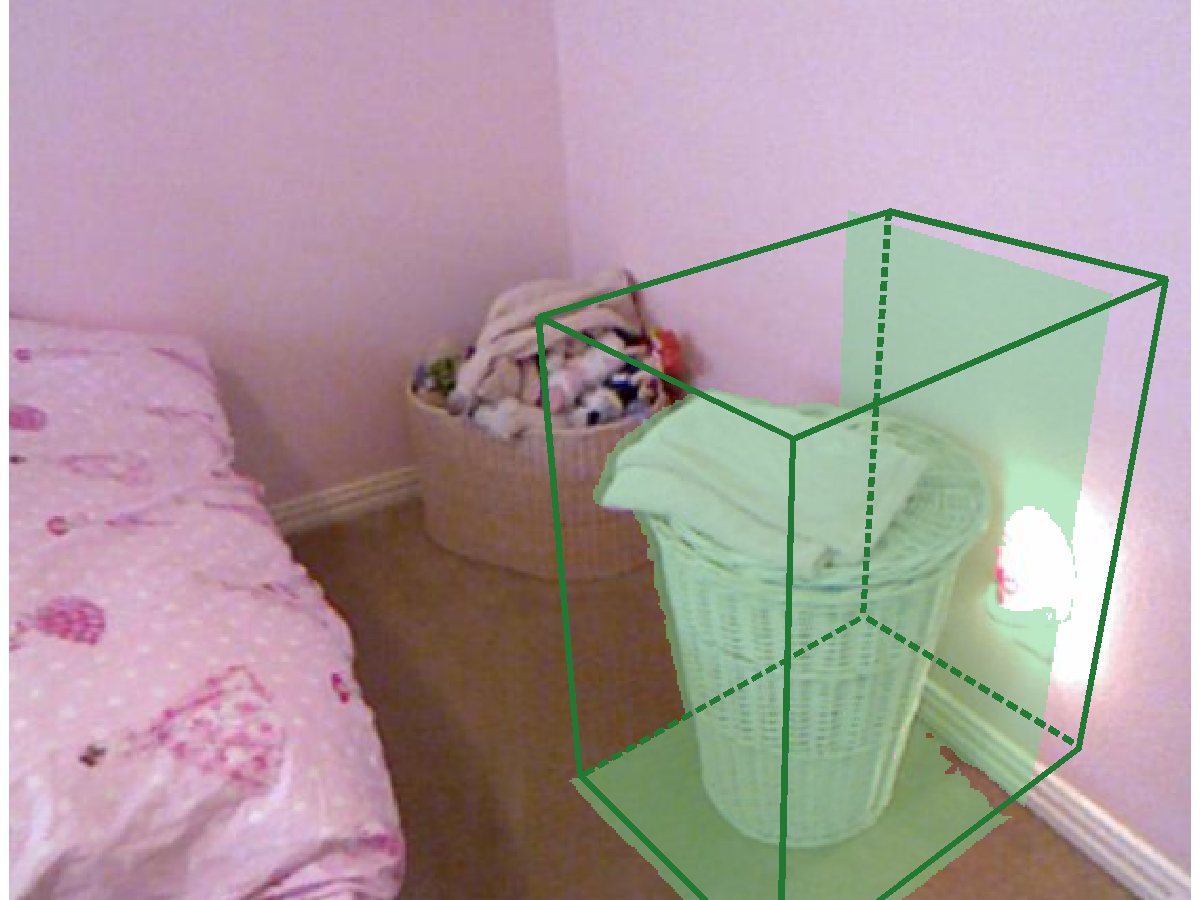}%
\includegraphics[width=0.227\linewidth]{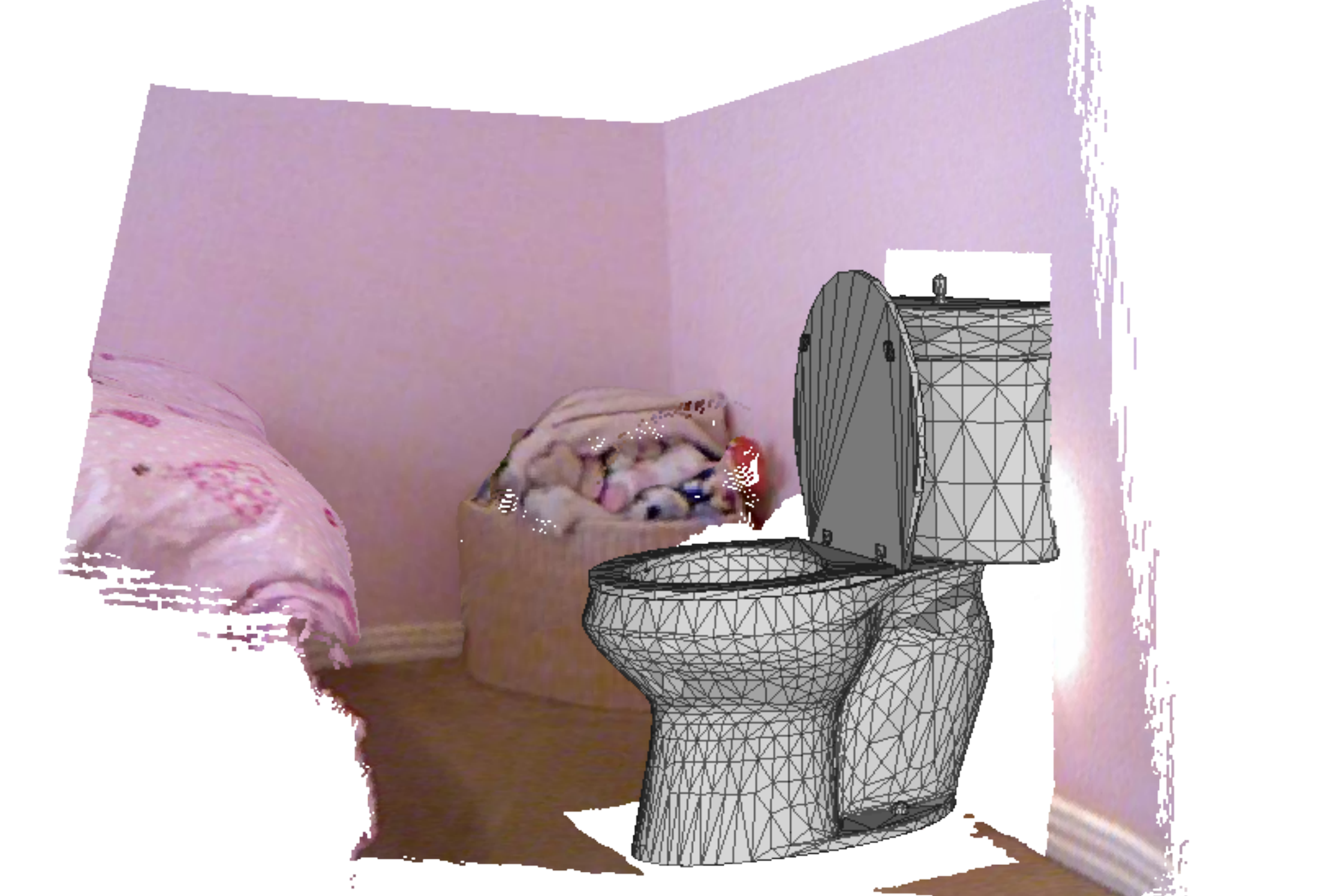}~~~~%
\includegraphics[width=0.227\linewidth]{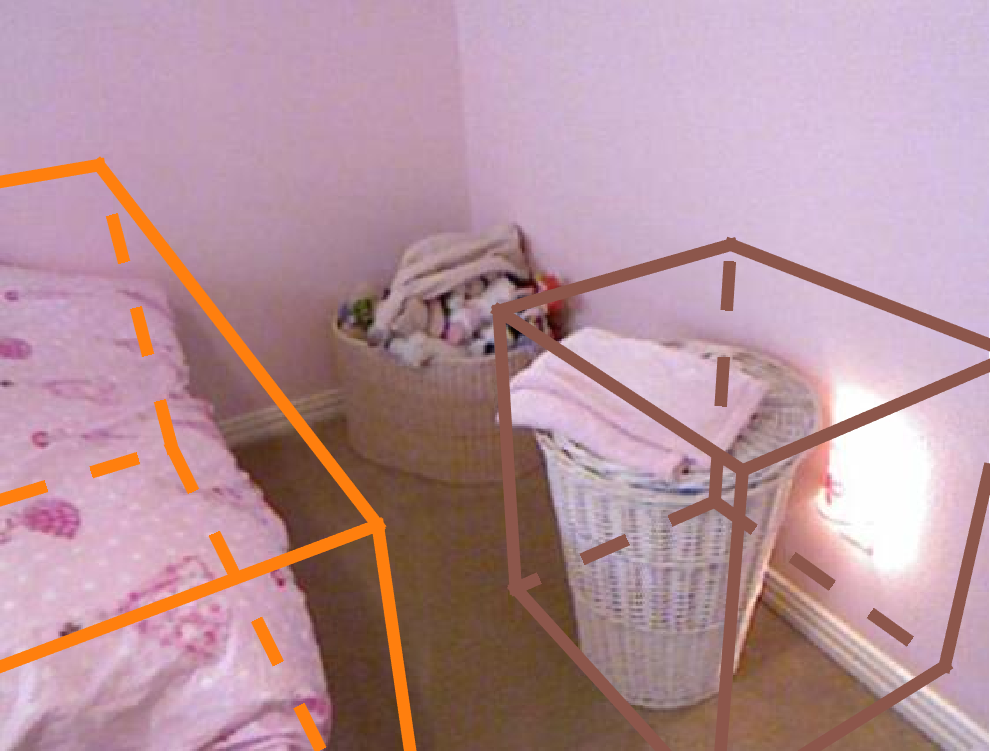}

\includegraphics[width=0.227\linewidth]{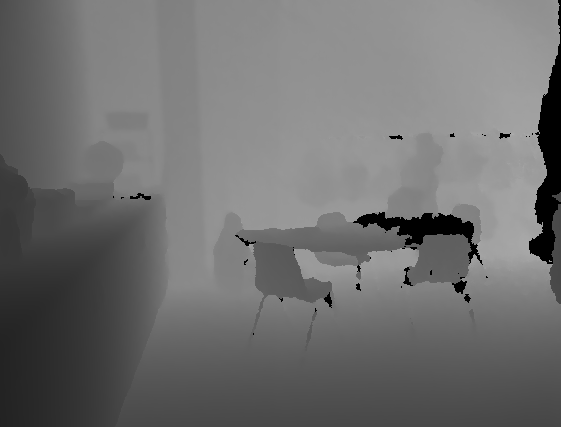}~~~%
\includegraphics[width=0.227\linewidth]{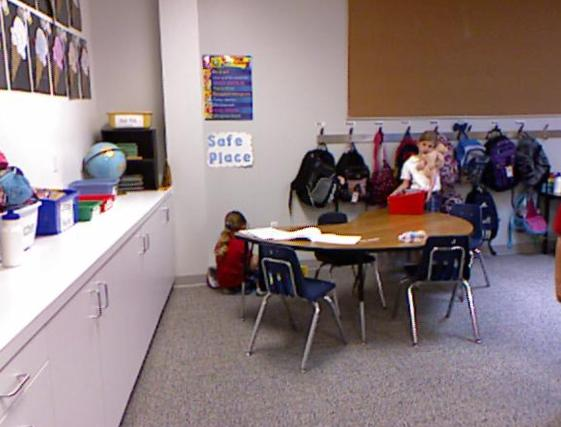}%
\includegraphics[width=0.227\linewidth]{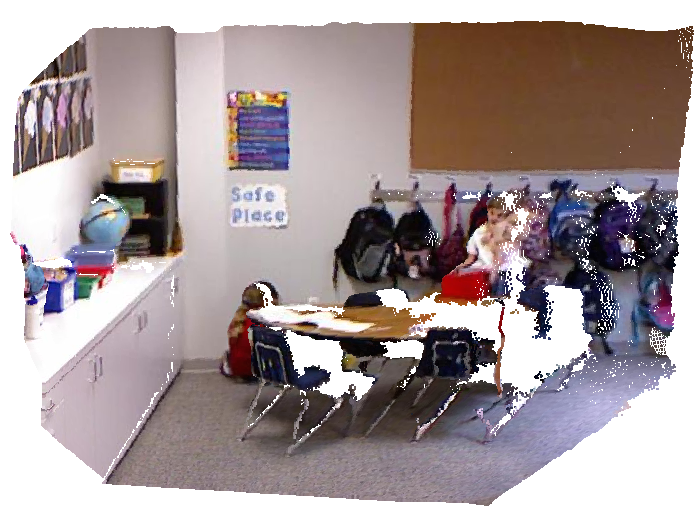}~~~~%
\includegraphics[width=0.227\linewidth]{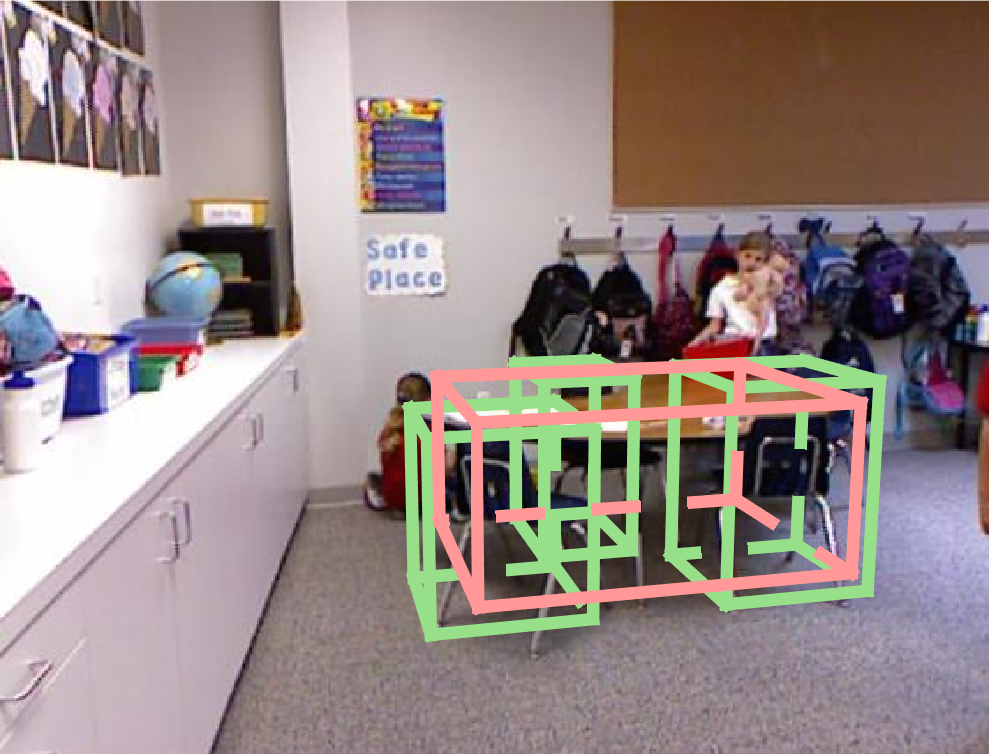}

\includegraphics[width=0.95\linewidth]{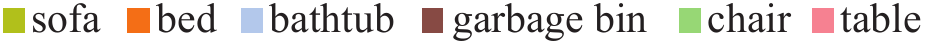}
\end{center}

\vspace{-4mm}

\caption{{\bf Comparision with Sliding Shapes \cite{SlidingShapes}.} Our algorithm is able to better use shape, color and contextual information to handle more object categories, resolve the ambiguous cases, and detect objects with atypical size (\eg smaller than regular).
}
\label{fig:compareSS}
\vspace{-5mm}
\end{figure}

{\bf Is external training data necessary?}
Comparing to Sliding Shapes that uses extra CAD models,
and Depth-RCNN that uses Image-Net for pre-training and CAD models for 3D fitting,
our [depth only] 3D ConvNet does not require any external training data outside NYUv2 training set, and still outperforms the previous methods, which shows the power of 3D deep representation. 

\vspace{-4mm}\paragraph{Comparison to the state-of-the-arts}
We evaluate our algorithm on the same test set as \cite{guptaCVPR15}
(The intersection of the NYUv2 test set and Sliding Shapes test set for the five categories being studied under ``3D all'' setting).
Table \ref{fig:3ddetection} shows the comparison with the two state-of-the-arts for amodal 3D detection:
3D Sliding Shapes \cite{SlidingShapes} with hand-crafted features,
and 2D Depth-RCNN \cite{guptaCVPR15} with ConvNets features.
Our algorithm outperforms by large margins with or without colors.
Different from Depth-RCNN that requires fitting a 3D CAD model as post-processing, 
our method outputs the 3D amodal bounding box directly,
and it is much faster.
Table \ref{fig:result} shows the amodal 3D object detection results on SUN RGB-D dataset compared with Sliding Shapes \cite{SlidingShapes}. 
%Therefore, it is also much faster than those methods.

Figure \ref{fig:compareSS} shows side-by-side comparisons to Sliding Shapes.
First, the object proposal network and box regression provide more flexibility to detect objects with atypical sizes. 
For example, the small child's chairs and table in the last row are missed by Sliding Shapes but detected by Deep Sliding Shape.
Second, color helps to distinguish objects with similar shapes (\eg bed \vs table).
Third, the proposed algorithm can extend to many more object categories easily.
%\eg bathtub and garbage bin.
%The proposed algorithm can easily handle more object categories, which helps to resolve ambiguous cases. 
%For example, the bathtub is now correctly detected, instead of being mis-classified as sofa. 

\vspace{-2mm}
\section{Conclusion}
\vspace{-2mm}

We present a 3D ConvNet pipeline for amodal 3D object detection, including a Region Proposal Network and a joint 2D+3D Object Recognition Network.
Experiments show our algorithm significantly outperforms the state-of-the-art approaches,
and is much faster than the original Sliding Shapes, which demonstrates the power of 3D deep learning for 3D shape representation.

\vspace{-2mm}
\paragraph{Acknowledgment.}
This work is supported by NSF/Intel VEC program. 
Shuran is supported by a Facebook fellowship.
We thank NVIDIA and Intel for hardware donation.
We thank Jitendra Malik and Thomas Funkhouser for valuable discussion.

\vspace{-2mm}
{\small
\bibliographystyle{ieee}
\bibliography{RGBDdetect,pvg}
}

\end{document}